\pdfoutput=1
\documentclass[11pt]{article}

\usepackage[final]{acl}

\usepackage{times}
\usepackage{latexsym}
\usepackage{csquotes}
\usepackage{amsmath}

\usepackage[T1]{fontenc}
\usepackage[utf8]{inputenc}
\usepackage{csquotes} % for \enquote{...}

\usepackage{microtype}
\usepackage{comment}
\usepackage{inconsolata}

\usepackage{graphicx}
\usepackage{subcaption}
\usepackage{enumitem}

\usepackage{array}
\newcolumntype{P}[1]{>{\raggedright\arraybackslash}p{#1}}

\title{Word-Centered Semantic Graphs for Interpretable Diachronic Sense Tracking}

\author{
 \textbf{Imene Kolli\textsuperscript{1,2}},
 \textbf{Kai-Robin Lange\textsuperscript{2}},
 \textbf{Jonas Rieger\textsuperscript{2}},
 \textbf{Carsten Jentsch\textsuperscript{2}},
\\
 \textsuperscript{1}University of Zurich,
 \textsuperscript{2}TU Dortmund University,
\\
\texttt{imene.kolli@df.uzh.ch, \{kalange, rieger, jentsch\} @statistik.tu-dortmund.de}\\
}
\begin{document}
\maketitle
\begin{abstract}
We propose an interpretable, graph-based framework for analyzing semantic shift in diachronic corpora. For each target word and time slice, we induce a word-centered semantic network that integrates distributional similarity from diachronic Skip-gram embeddings with lexical substitutability from time-specific masked language models. We identify sense-related structure by clustering the peripheral graph, align clusters across time via node overlap, and track change through cluster composition and normalized cluster mass. In an application study on a corpus of \textit{New York Times Magazine} articles (1980--2017), we show that graph connectivity reflects polysemy dynamics and that the induced communities capture contrasting trajectories: event-driven sense replacement (\textit{trump}), semantic stability with cluster over-segmentation effects (\textit{god}), and gradual association shifts tied to digital communication (\textit{post}). Overall, word-centered semantic graphs offer a compact and transparent representation for exploring sense evolution without relying on predefined sense inventories.
\end{abstract}

\section{Introduction}

Languages evolve with their speaker communities, leading to changes in word meanings over time. One prominent form of language change is \emph{semantic shift}, where a word's meaning transforms due to factors such as cultural developments, technological innovations, and contact with other languages \citep[p, 1--2]{tahmasebi_computational_2021}.\footnote{A classic example is the English word \textit{boy}, which originally meant \enquote{male servant} in the 14th century and later came to denote \enquote{any male child} \citep{giulianelli_analysing_2020}.} Understanding how such semantic changes occur is central to historical linguistics and cultural studies, and has become increasingly important for language technologies operating on long-spanning textual data \citep{vydaichuk_semantic_2024}.

In semantic linguistics, word meaning refers to the conceptual content conveyed by a word \citep{sep-word-meaning}. A key property of word meaning is \emph{polysemy}, where a single word expresses multiple senses depending on context. Polysemy plays a crucial role in semantic change, as different senses often coexist for extended periods before one sense becomes dominant, leading to gradual shifts in meaning \citep{article_im}. Semantic shift can take several forms, including meaning expansion, contraction, and the emergence of new senses \citep{daiu_semantic_2015}.

Identifying semantic shift requires encoding a word's senses into a prototypical representation through high-level semantic properties such as the dominant sense and the degree of polysemy, or low-level properties such as the sense distribution \citep{10.1145/3672393}. In this work, we operationalize these properties through graph structure and community evolution: dense cohesive neighborhoods suggest stable meanings, while the emergence of disconnected regions indicates increasing polysemy and sense differentiation. Tracking these changes can reveal the emergence of new meanings, as well as the broadening or narrowing of existing ones \citep{10.1145/3672393}. It also highlights ephemeral changes, where external events temporarily influence usage before a word reverts to earlier meanings \citep{shoemark-etal-2019-room}.

Recent advances in natural language processing have enabled computational approaches to semantic shift analysis, reducing reliance on time-consuming manual annotation \citep{survey}. Among these, diachronic word embedding methods \citep[e.g.,][]{rieger2022, lange2025a, lange2022, schmidt2025} have become the dominant paradigm, extending both static \citep[e.g.,][]{hamilton-etal-2016-diachronic, dubossarsky-etal-2017-outta} and contextual \citep[e.g.,][]{hu-etal-2019-diachronic, kanjirangat-etal-2020-sst} embedding models to temporally ordered corpora. While effective at capturing large-scale semantic trends, both approaches face inherent limitations when representing polysemous words. Static embeddings compress multiple senses into a single vector, making it difficult to disentangle sense-specific change \citep{giulianelli_analysing_2020}. Contextual embeddings can represent multiple senses, but typically require aggregation into discrete sense inventories or clusters, introducing challenges related to comparability and interpretability across time \citep{10.1145/3672393}.

In this work, we address these limitations by combining static and contextual diachronic word embeddings into a unified graph-based representation. For each target word and time period, we construct a semantic network that integrates global distributional similarity, captured by static embeddings, with context-sensitive lexical substitutability \citep{zhou-etal-2019-bert}, captured by contextual embeddings. Concretely, we (i) induce word-centered graphs from static similarity and contextual substitutability, (ii) extract sense communities via peripheral connectivity, and (iii) track sense evolution with overlap-based alignment and normalized community mass.

Semantic networks provide an interpretable framework for modeling meaning through relational structure \citep{sowa1992semantic}, and their topological organization has been shown to encode important semantic properties \citep{budel2023topological}. Unlike manually curated resources such as WordNet \citep{10.1145/219717.219748}, the networks we construct are induced directly from corpus evidence and reflect usage-driven semantic organization. By representing a word's multiple senses as spatial structures within these graphs and tracking their evolution over time, our approach enables graph-based analysis of semantic change that supports sense identification, sense distribution estimation, and temporal shift analysis while maintaining interpretability at both the sense and word levels.

\section{Related Work}

Computational approaches to semantic shift are commonly grouped into \emph{form-based} and \emph{sense-based} methods \citep{giulianelli_analysing_2020}. Form-based approaches rely on static embeddings, assigning a single vector per word (and per time period), with semantic change approximated via vector drift across aligned spaces. A standard pipeline trains separate embedding models on time-sliced corpora, aligns the resulting spaces, and uses measures such as cosine distance to rank words by change magnitude \citep[e.g.,][]{hamilton-etal-2016-diachronic, dubossarsky-etal-2017-outta}. While effective at capturing shifts in a word's \emph{dominant} usage, these methods are inherently limited for polysemous words because they compress multiple senses into one representation \citep{giulianelli_analysing_2020}. They are also sensitive to modeling choices, including context window size and the selection of a reference period \citep{levy-etal-2015-improving, shoemark-etal-2019-room}.\footnote{With small windows, embeddings tend to reflect local interchangeability (synonymy/collocation), whereas larger windows can emphasize broader topical/contextual similarity rather than core meaning \citep{levy-etal-2015-improving}.}

Sense-based approaches aim to explicitly model multiple meanings using contextual embeddings, where each occurrence of a word is represented in context \citep{giulianelli_analysing_2020}. To enable diachronic comparison, token representations are typically aggregated into sense-level units via supervised classification or unsupervised clustering \citep[e.g.,][]{hu-etal-2019-diachronic, kanjirangat-etal-2020-sst}. While these methods enable finer-grained analysis of sense usage, they introduce challenges related to aggregation, comparability across time, and cluster alignment \citep{10.1145/3672393}. In particular, supervised approaches constrain senses to predefined inventories, while clustering-based methods face sensitivity to cluster number, frequency effects, and the conflation of usage similarity with semantic relatedness \citep{10.1145/3672393,i_dont_belive_in_word_senses}.

More recently, semantic change has been studied through relational representations, using graphs to model meaning via structured connections rather than isolated vectors or fixed sense inventories. \citet{ma-etal-2024-graph} construct semantic trees by clustering contextual embeddings into a predefined set of sense centroids and tracking these structures over time. While this enables a relational view of meaning, it fixes sense granularity a priori and relies solely on contextual similarity, limiting its ability to capture emergent or transient senses. In contrast, our approach induces word-centered semantic graphs directly from corpus evidence by integrating static and contextual signals, allowing sense structure and semantic change to emerge dynamically over time.

Beyond representation, prior work has also proposed various metrics for quantifying semantic change, including divergence between sense distributions \citep{schlechtweg-etal-2020-semeval}, distances between sense representations \citep{giulianelli_analysing_2020}, and novelty-based scores \citep{cook-etal-2014-novel}. Interpreting these signals typically still requires manual inspection of clusters or representative terms \citep{giulianelli_analysing_2020,kellert-mahmud-uz-zaman-2022-using}.

\section{Methodology}

We model word meaning through word-centered semantic graphs, adopting a relational and holistic perspective in which semantic properties are expressed through patterns of relations among words rather than isolated representations \citep{budel2023topological,musil-2021-representations}. In this view, a word's meaning is characterized by the structure of its semantic neighborhood and the interactions among its neighbors. Differences in connectivity patterns, neighborhood composition, and local density provide the basis for identifying semantic shift and sense evolution, rather than by tracking the movement of a single embedding vector. 

\subsection{Word neighborhood network}
\label{sec:word_neighborhood}

Let $G_w^t = (N_w^t, E_w^t)$ denote the word neighborhood network of target word $w$ in time slice $t$. The node set $N_w^t$ contains $w$ and a set of neighboring words, while edges $E_w^t$ encode semantic relatedness derived from distributional and contextual evidence.

To populate the network, we rely on two complementary types of semantic relations: (i) \emph{distributional similarity}, capturing global lexical associations rather than strict synonymy, and (ii) \emph{lexical substitutability}, capturing usage-conditioned meanings grounded in concrete contexts of occurrence.

\paragraph{Distributional neighbors (static).}
We use Word2Vec with a Skip-gram architecture \citep{mikolov2013a} to capture distributional similarity. Skip-gram is particularly suited to neighborhood-based representations because it emphasizes word-context associations \citep{musil-2021-representations}, yielding sharper nearest-neighbor relations aligned with lexical relatedness \citep{levy-etal-2015-improving}.

We train a separate Skip-gram model on the corpus of each time slice $t$. For a target word $w$, we select the $k_i$ words with highest cosine similarity in the embedding space, forming the set of distributional neighbors:
\[
N_{\text{dist}}^t(w) = \text{TopK}_{k_i}\big(\cos(\mathbf{e}_w^t, \mathbf{e}_u^t)\big).
\]

\paragraph{Substitution neighbors (contextual).}
To capture lexical substitutability, we fine-tune a RoBERTa masked language model \citep{liu2019roberta} on the same time-slice corpus. For each occurrence of $w$ in the corpus, we mask the token and obtain a ranked list of predicted substitutes. Candidates are ranked by prediction frequency and the top $k_c$ are selected to form the substitution neighbor set:
\[
N_{\text{sub}}^t(w) = \text{TopK}_{k_c}\big(\text{freq}(\hat{u} \mid \text{mask}(w))\big).
\]

\paragraph{Graph initialization and expansion.}
The first-layer neighborhood of $w$ is defined as
\[
N_1^t(w) = N_{\text{dist}}^t(w) \cup N_{\text{sub}}^t(w).
\]
We initialize the graph by adding edges between $w$ and each word in $N_1^t(w)$.

To capture indirect semantic associations, we expand the network in a layered fashion up to depth $L$. For each node $u$ introduced at layer $l-1$, we compute its own distributional and substitution neighbors, selecting $k_i^{(l)}$ and $k_c^{(l)}$ neighbors respectively, which are added as nodes at layer $l$ for $l \geq 2$. The resulting graph captures both direct and indirect relations while remaining anchored in the usage of the target word within time slice $t$.

The connectivity within layers is an important aspect of the graph structure. As the graph expands to include more layers, the relationships between the nodes may become more interconnected. Two types of connectivity phenomena are expected to emerge in the graph:
\begin{itemize}
	\item \textbf{Local connectivity}: edges among nodes introduced at the same depth, reflecting overlap in distributional similarity or substitution behavior.
	\item \textbf{Global connectivity}: edges between nodes at different depths created during expansion, which arise when different first-layer neighbors introduce shared second-layer neighbors.
\end{itemize}

\subsection{Sense communities via peripheral connectivity}
\label{sec:graph_clustering}

To identify sense-related structure within the word neighborhood network, we analyze the graph's connectivity patterns rather than clustering embedding vectors directly. Since the central node $w$ trivially connects all first-layer neighbors, we perform clustering on a \emph{peripheral graph} that excludes the central node. We define the peripheral graph
\[
G_{w,\text{per}}^t = G_w^t \setminus \{w\},
\]
obtained by removing $w$ and all incident edges. The remaining graph retains only relations among neighboring words and higher-layer nodes.

We define sense communities as the connected components of $G_{w,\text{per}}^t$:
\[
\mathcal{C}^t(w) = \{ C_1^t, \dots, C_m^t \},
\]

This yields a variable number of communities per $t$, reflecting changes in neighborhood structure. The communities are intra-connected through paths of distributional and substitution edges and inter-connected only via the central target node $w$, leading to the formation of cohesive regions that are interpreted as a sense-related semantic community, following prior work that associates clusters of semantically related neighbors with word senses \citep{kanjirangat-etal-2020-sst,10.1145/3672393}.

\subsection{Cluster alignment and refinement}
\label{sec:alignment}

To track senses across time periods, we align clusters using node overlap. For a cluster $C_i^t \in \mathcal{C}^t(w)$, we identify the corresponding cluster in the previous time slice by maximizing set intersection:
\[
\text{Align}(C_i^t) = \arg\max_{C_j^{t-1} \in \mathcal{C}^{t-1}} |C_i^t \cap C_j^{t-1}|.
\]
To account for re-emerging senses that may be absent in intermediate periods, we also allow alignment to any earlier time slice:
\[
\text{AlignHist}(C_i^t) = \arg\max_{k < t} \max_{C_j^{k} \in \mathcal{C}^{k}} |C_i^t \cap C_j^{k}|.
\]

If no overlapping nodes are found, the cluster is considered a new sense. Both methods track sense continuity via node overlap without requiring embedding-space alignment.

After alignment, we refine the cluster assignments by ignoring clusters that do not persist for at least two time periods, merging their nodes into a \emph{residual cluster} that captures infrequent or ephemeral senses.

\subsection{Sense usage distribution}
\label{sec:sense_distribution}

We estimate the usage distribution of senses for target word $w$ at time $t$ based on cluster sizes. Let $|C_i^t|$ denote the number of nodes in cluster $C_i^t$. We define the sense usage distribution as
\[
P^t(C_i \mid w) = \frac{|C_i^t|}{\sum_{j=1}^{m} |C_j^t|}.
\]
This normalization yields a probability distribution over sense communities that is comparable across time slices despite variation in graph size.

\section{Application results}
\label{sec:application_results}

We evaluate the proposed framework on a diachronic corpus of \textit{New York Times Magazine} articles spanning 1980–2017. Our analysis focuses on three target words: \textit{trump}, \textit{god}, and \textit{post}, selected to exhibit contrasting semantic profiles in terms of corpus relevance, frequency dynamics, and potential for semantic change. Details on corpus construction, preprocessing, target word selection, and modeling setup are provided in Appendix~\ref{app:data}--\ref{app:setup}.

The evaluation proceeds in three stages. First, we examine the structure of word-centered neighborhood graphs to assess how semantic properties such as polysemy are reflected in connectivity patterns. Second, we apply peripheral connectivity clustering to identify sense-related communities and trace their temporal evolution. Finally, we analyze sense usage distributions derived from cluster sizes to characterize different types of semantic shift.

\subsection{Target word neighborhood structure}
\label{sec:target_network}

Graphs are constructed using a fixed configuration (Appendix~\ref{app:setup}), yielding a theoretical maximum of 9 first-layer neighbors and 27 second-layer neighbors, for a total of at most 37 nodes per graph including the target word. In practice, observed graph sizes are often smaller due to several structural factors.

\begin{figure}[t]
	\centering
	\begin{subfigure}[b]{\linewidth}
	   \centering
	   \includegraphics[width=\linewidth]{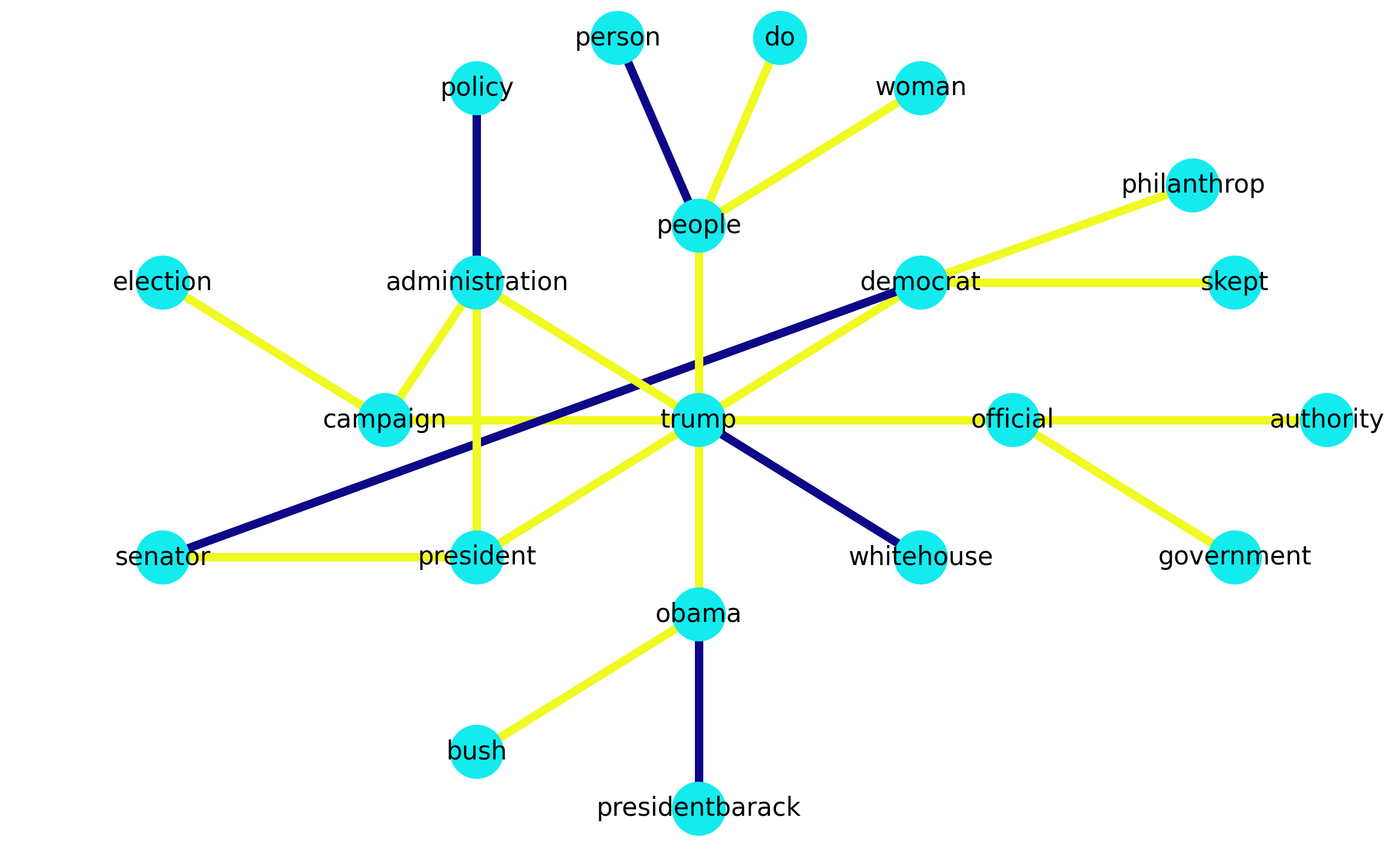}
		\caption{}
		\label{fig:trump_enet_2017}
    \end{subfigure}
	\vfill
	\begin{subfigure}[b]{\linewidth}
		\centering
		\includegraphics[width=\linewidth]{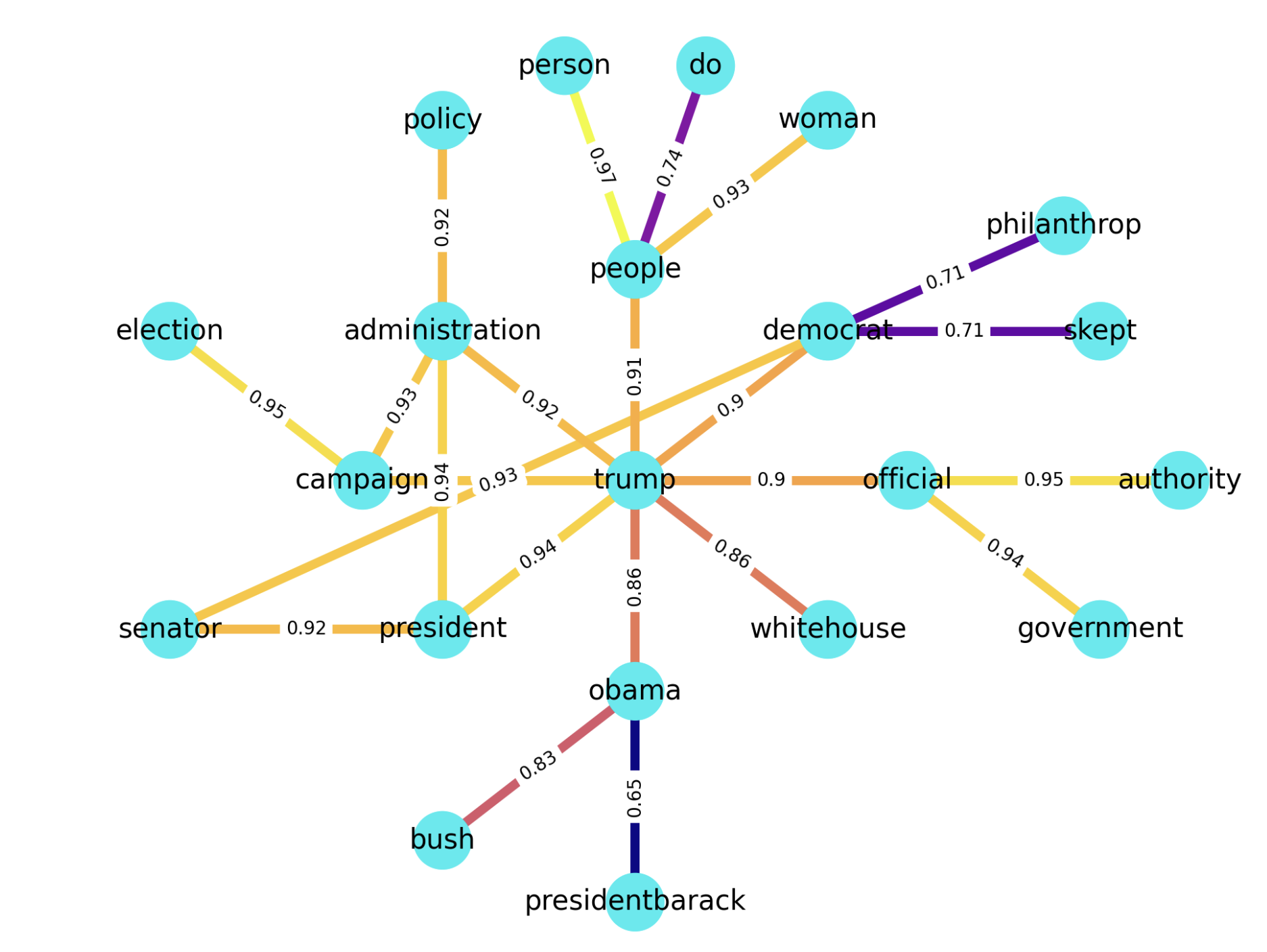}
		\caption{}
		\label{fig:trump_net_2017}
	\end{subfigure}
    \caption{Word-centered semantic networks of the word \textit{trump} in 2017. (Top) Edge: blue edges denote distributional similarity (Word2Vec), yellow edges denote substitution-based similarity (RoBERTa); if both relations hold, the edge is shown in yellow. (Bottom) Edge: cosine similarity between contextual (RoBERTa) embeddings of the connected nodes. Darker edges denote lower similarity. The network reflects a politically grounded semantic neighborhood dominated by institutional and partisan associations.}
    \label{fig:trump_2017_networks}
\end{figure}

First, the symmetric nature of cosine similarity leads distributional relations to be frequently reciprocal. As a result, certain nodes cannot be expanded further if their most similar neighbor is the target word itself. For example, in the 2017 \textit{trump} graph (Figure~\ref{fig:trump_enet_2017}), the node \textit{whitehouse} has no outgoing distributional expansions because its closest SGNS neighbor is \textit{trump}. 

Second, local and global connectivity among neighbors induces overlap across layers, reducing the number of unique nodes. This is illustrated by nodes such as \textit{obama}, whose contextual neighborhood is dominated by \textit{trump} and \textit{bush}, limiting further expansion. 

Third, vocabulary mismatches between static and contextual models can prevent certain nodes from being expanded; for instance, \textit{whitehouse} does not appear in the vocabulary of the fine-tuned RoBERTa model and therefore cannot generate substitution-based neighbors.

Beyond topology, edge similarity values provide additional interpretive structure. In the 2017 \textit{trump} network (Figure~\ref{fig:trump_net_2017}), lower cosine similarities are more prevalent among distributional edges, consistent with the fact that these neighbors are selected in Word2Vec's vector space and need not be close in contextual embedding space. Nevertheless, similarity remains interpretable: e.g., \textit{democrat} is closer to \textit{senator} than to \textit{philanthrop}, reflecting the political framing of the neighborhood during Trump's presidency.

The \textit{god} network (Figures~\ref{fig:god_networks}) exhibits a comparatively dense and stable node composition, with consistently high embedding similarities (often exceeding 90\%), suggesting limited semantic diversification over time. The target \textit{post} shows a large variation in neighbor composition (Figure~\ref{fig:post_networks}), where nodes such as \textit{share} and \textit{socialmedium} replace earlier neighbors like \textit{position} and \textit{paper}, indicating a semantic association shift toward digital communication.

\subsection{Connectivity and polysemy}

\begin{figure*}
    \centering
    % Left image
    \begin{subfigure}{0.495\textwidth}
        \centering
        \includegraphics[width=\linewidth]{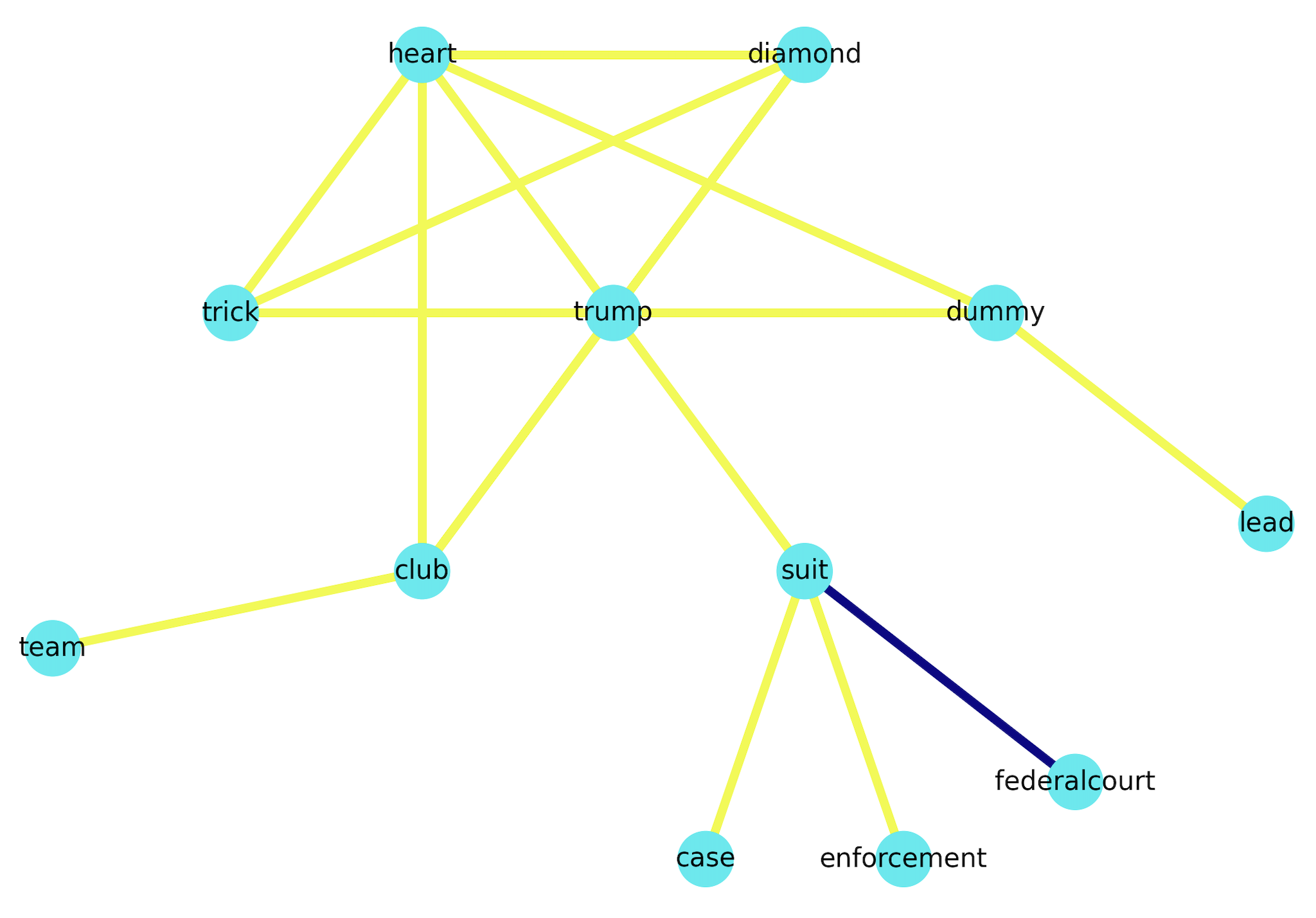}
        \caption{Year 1980}
	    \label{fig:trump_net_1980}
    \end{subfigure}
    \hfill
    % Right image
    \begin{subfigure}{0.495\textwidth}
        \centering
        \includegraphics[width=\linewidth]{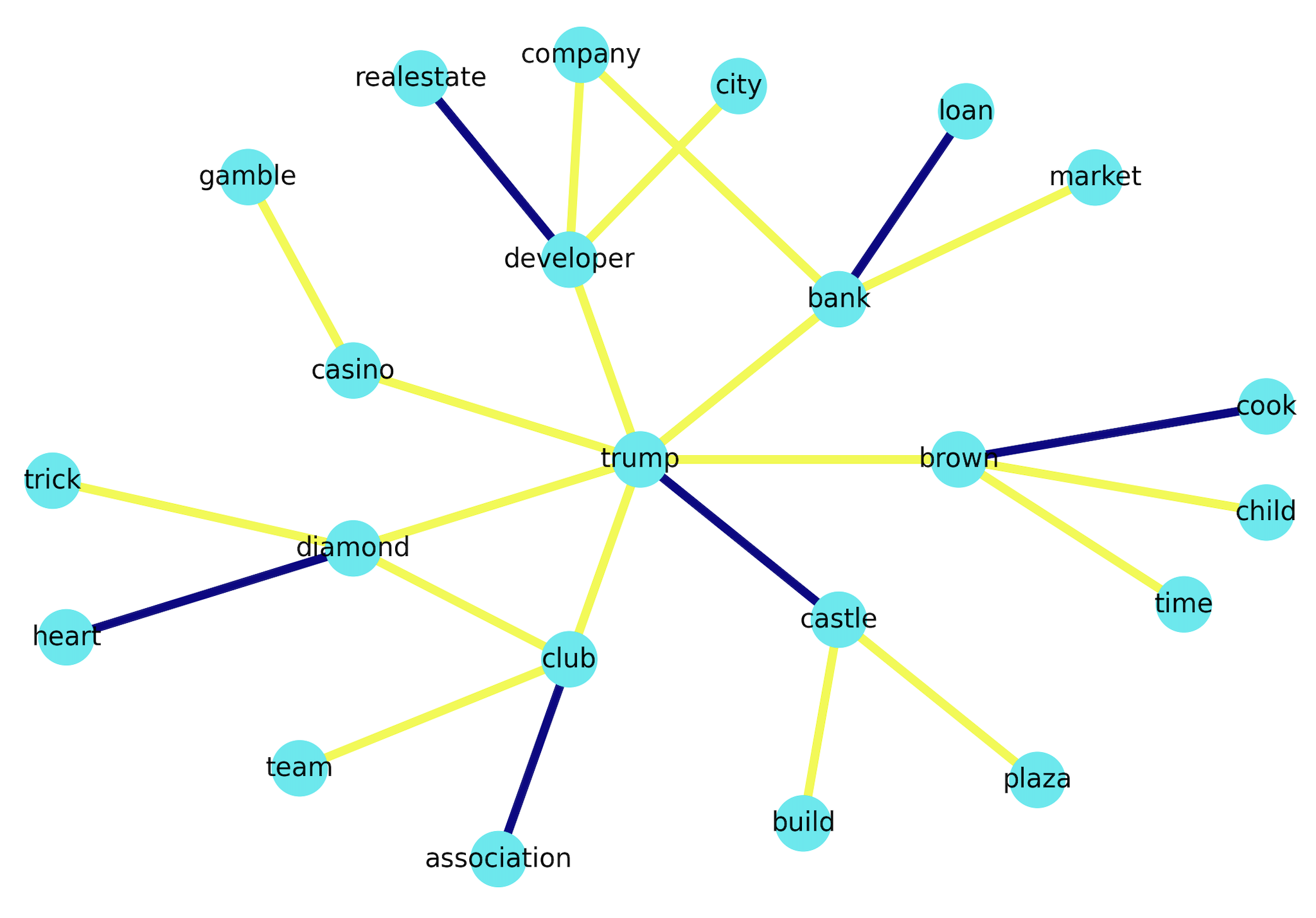}
        \caption{Year 1990}
	    \label{fig:trump_net_1990}
    \end{subfigure}
	\caption{Word-centered semantic networks of \textit{trump} in 1980 (left) and 1990 (right). Blue edges indicate distributional similarity; yellow edges indicate substitution-based similarity. The 1980 network forms a dense, cohesive structure corresponding to the literal card-game sense (low polysemy), whereas the 1990 network exhibits disconnected communities associated with business-related meanings, indicating increased polysemy.}
    \label{fig:trump_enet}
\end{figure*}

Figure~\ref{fig:trump_enet} contrasts the \textit{trump} networks in 1980 and 1990. In 1980 (Figure~\ref{fig:trump_net_1980}), most neighbors are related to card games (e.g., \textit{diamond}, \textit{heart}). The graph is densely interconnected, exhibiting both local and global connectivity: nodes such as \textit{trick} and \textit{diamond} do not introduce many new second-layer nodes because they are mutual contextual substitutes and share overlapping neighbors already present in the first layer. This dense, cohesive structure indicates a low degree of polysemy, with a single dominant sense.

By 1990 (Figure~\ref{fig:trump_net_1990}), the neighborhood composition changes substantially: new nodes such as \textit{developer} and \textit{casino} appear and form communities disconnected from the card-game sense. The emergence of these disconnected communities indicates a polysemous stage in which the literal sense coexists with a culturally grounded business-related sense that is not attested in classical dictionaries but is associated with Donald Trump's rise as a real-estate developer and casino owner.

\begin{figure}[t]
    \centering
	\includegraphics[width=\linewidth]{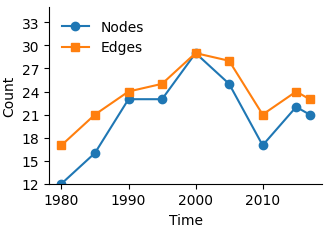}
     \caption{Graph properties of the \textit{trump} neighborhood over time. Blue line shows the number of nodes; orange line shows the number of edges. The peak around 2000 corresponds to a transitional polysemous phase, followed by a decline reflecting sense contraction and replacement.}
    \label{fig:trump_graph_ts}
\end{figure}

Graph size over time further supports this interpretation. For \textit{trump}, the number of nodes and edges increases steadily and peaks around 2000 (Figure~\ref{fig:trump_graph_ts}). This peak corresponds to a transitional polysemous phase \citep{giulianelli_analysing_2020}. After 2000, node and edge counts decline, suggesting reduced polysemy and the disappearance of previously observed senses. Notably, the graph never reaches its theoretical maximum size, reflecting the structural constraints induced by neighbor overlap and reciprocal relations.

In contrast, \textit{god} shows relatively stable node and edge counts over time (Figure~\ref{fig:god_ts}), consistent with a unified semantic interpretation and limited semantic change. The \textit{post} networks exhibit a noticeable increase in connectivity around 2005--2010 (Figure~\ref{fig:post_ts}), aligning with periods where new associations related to digital communication become more prominent.

\subsection{Semantic properties encoded in the network}
\label{sec:semantic_properties}

We now examine how the network representation captures fine-grained semantic properties beyond global polysemy indicators, by tracking how specific neighbor communities emerge, recede, and stabilize across time.

\paragraph{\textit{trump}.}
Figure~\ref{fig:trump_other_nets} shows the evolution of the \textit{trump} network in 1985, 2005, 2010, and 2015. In 1985 (Figure~\ref{fig:trump_net_1985}), the network continues to reflect the literal card-game sense observed in 1980 (Figure~\ref{fig:trump_net_1980}): neighbors form a cohesive and highly interconnected semantic category. Compared to 1980, node frequency is higher, indicating an increased degree of polysemy despite continued dominance of the literal sense.

By 2005 (Figure~\ref{fig:trump_net_2005}), the literal sense remains present (e.g., \textit{diamond}, \textit{heart}) but is less prominent than in 1985. At the same time, new nodes such as \textit{condominium} appear, forming communities that are disconnected from the literal sense and co-occurring with business-related nodes such as \textit{casino}. These nodes reflect a culturally grounded business sense associated with Donald Trump's real-estate activity, a sense that is not captured by classical dictionaries \citep{i_dont_belive_in_word_senses}.

In 2010 (Figure~\ref{fig:trump_net_2010}), the neighborhood reverts to one dominated by the literal card-game sense (e.g., \textit{diamond}, \textit{heart}, \textit{trick}). This indicates that while business-related senses emerged in earlier periods, they did not persist in the network representation. 

In 2015 (Figure~\ref{fig:trump_net_2015}), a new semantic category centered on political terms (e.g., \textit{candidate}, \textit{campaign}) emerges. This community is culturally grounded and not derivable from classical dictionary senses, reflecting Donald Trump's presidential campaign. Crucially, the literal card-game sense is absent from the neighborhood in this period, indicating a more substantive semantic reconfiguration. Together with the political framing visible in Figures~\ref{fig:trump_net_2017}, this suggests a shift in dominant usage from the literal sense to the proper-noun political sense in the corpus.

\paragraph{\textit{god}.}
Across time, the neighborhood networks of the word \textit{god} remain structurally consistent (Figures~\ref{fig:god_other_nets}), with similar node sets and connectivity patterns recurring across periods. Inter-node similarities remain uniformly high, and no new disconnected communities emerge. This stability is consistent with the observed constancy in graph size (Figure~\ref{fig:god_ts}), with no evidence of sustained sense emergence or replacement.

\paragraph{\textit{post}.}
The target word reflects stable neighborhoods before 2010 (Figure~\ref{fig:post_other_nets}), when the network undergoes a noticeable change: similarity decreases and new nodes such as \textit{twitter} and \textit{video} appear. This shift is consistent with the graph size increase observed in Figure~\ref{fig:post_ts}. Unlike for the \textit{trump} network, this change is not characterized by abrupt disappearance of earlier senses, but rather by an expansion of associations as technological discourse becomes more central in the corpus.

\subsection{Sense identification via peripheral connectivity}
\label{sec:clustering}

To identify sense-level structure within the word neighborhood networks, we apply \textbf{peripheral connectivity clustering} (Section~\ref{sec:graph_clustering}), which groups nodes based on their mutual connectivity after removing the target word.

\paragraph{Peripheral clusters as sense proxies.}
For the target word \textit{trump}, peripheral clustering consistently separates the literal card-game sense from culturally grounded senses tied to Donald Trump's public career (Figure~\ref{fig:trump_clusters}). In early periods, the dominant cluster contains nodes such as \textit{diamond}, \textit{heart}, and \textit{trick}, forming a cohesive semantic category aligned with the dictionary sense. Minor clusters, when present, are small and weakly connected, indicating limited polysemy.

From the late 1980s onward, additional clusters emerge that are disconnected from the literal sense, including business-related communities (e.g., \textit{casino}, \textit{developer}) and later political communities (e.g., \textit{campaign}, \textit{candidate}). The presence of multiple disconnected components indicates periods of heightened polysemy, while their disappearance or dominance signals sense contraction or replacement.

\paragraph{Contrastive targets: stability and over-segmentation.}
Applying the same procedure to \textit{god} yields a markedly different outcome. Peripheral clustering produces 22 clusters across all periods, but these clusters remain semantically homogeneous and strongly interconnected. Nodes may transition between clusters over time, and no cluster persists as an isolated, competing sense. This behavior suggests that the detected clusters reflect \emph{facets of a single underlying sense} rather than distinct meanings, consistent with linguistic arguments against overly discrete sense inventories \citep{i_dont_belive_in_word_senses}.

\paragraph{Cluster alignment across time.}
Raw peripheral clustering can yield a large number of small or transient clusters. To track senses across time, we therefore align clusters using node overlap (Section~\ref{sec:alignment}). We consider two alignment strategies for the target word \textit{trump}: alignment restricted to the immediately preceding period (Figure~\ref{fig:trump_clusters_refined}), and alignment across all historical periods (Figure~\ref{fig:trump_clusters_refined2}).

Alignment to the previous period preserves fine-grained, temporally localized senses and is sensitive to short-lived semantic changes. However, it can lead to over-segmentation (7 clusters) when clusters fluctuate due to corpus-specific events. It also leads to a larger residual cluster, as many ephemeral clusters fail to persist for two periods.

In contrast, alignment to all historical periods produces fewer (4 clusters), more stable clusters but may suppress genuine minor senses by enforcing long-term consistency, causing newly emerging meanings to be lumped into earlier general clusters.

Empirically, the former strategy better captures transient semantic phenomena, while the latter is more appropriate for semantically stable senses. These results highlight a trade-off between temporal sensitivity and long-term coherence when aligning sense clusters diachronically.

\paragraph{Sense distributions and semantic shift.}
We normalize cluster sizes within each period, derived from both alignment methods, treating cluster mass as a proxy for sense frequency (Section~\ref{sec:sense_distribution}). 

\begin{figure}[t]
	\centering
	\begin{subfigure}[b]{\linewidth}
	   \centering
	   \includegraphics[width=\linewidth]{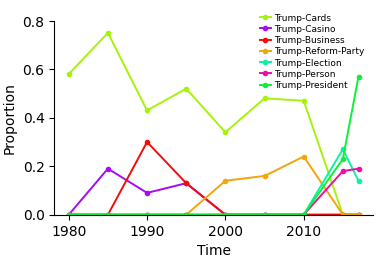}
		\caption{Previous period alignment.}
		\label{fig:trump_sd}
    \end{subfigure}
	\vfill
	\begin{subfigure}[b]{\linewidth}
		\centering
		\includegraphics[width=\linewidth]{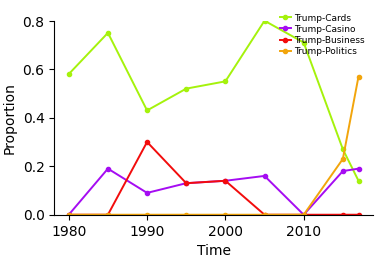}
		\caption{All historical periods alignment.}
		\label{fig:trump_sd_op}
	\end{subfigure}
    \caption{Sense usage distributions for \textit{trump} derived from normalized cluster mass over time.
(Top) Clusters aligned to the immediately preceding period, highlighting short-lived (event-driven) senses but yielding more fragmentation and a larger residual cluster.
(Bottom) Clusters aligned across all historical periods, producing fewer, more stable senses while potentially absorbing newly emerging meanings into earlier clusters.
Legend labels are assigned by manual inspection of cluster nodes.}

    \label{fig:trump_sds}
\end{figure}

For the target word \textit{trump} (Figure~\ref{fig:trump_sds}), this analysis reveals phases of sense broadening, narrowing, and emergence. Business- and casino-related senses rise and fall in the late 20th century, while political senses become dominant after 2015, coinciding with Donald Trump's presidential campaign and election. Earlier literal senses diminish, indicating a substantive shift in dominant usage.

For the target word \textit{god} (Figure~\ref{fig:god_clusters_sd}), sense communities converge over time, with initially fragmented clusters merging into a single dominant cluster (God-General). This convergence reflects refinement rather than loss of meaning and underscores the importance of temporal context when interpreting clustering outputs.

For the target word \textit{post} (Figure~\ref{fig:post_clusters_refined2}), sense distributions evolve gradually, with new clusters related to digital communication (e.g., \textit{instagram}, \textit{email}) gaining prominence over time. This pattern reflects a broadening of associations rather than abrupt sense shifts, consistent with the gradual integration of new technologies into everyday language.

\section{Summary}

This paper introduced a graph-based framework for analyzing semantic shift in diachronic corpora, representing word meaning as a word-centered semantic network induced from both static and contextual embeddings. By modeling meaning relationally rather than as a single vector or a fixed sense inventory, the approach enables interpretable analysis of polysemy, sense structure, and their evolution over time. Analysis of neighborhood graph structure showed that dense and cohesive connectivity corresponds to stable or low-polysemy meanings, while the emergence of disconnected communities reflects polysemous stages and sense differentiation.

Using peripheral connectivity clustering, we identified sense-related communities directly from graph structure and tracked their evolution via cluster composition and size. This enabled estimation of sense usage distributions and revealed multiple types of semantic change, including sense emergence, broadening, narrowing, and transient shifts driven by historically salient events. The results highlight clear contrasts between pronounced, event-driven semantic reconfiguration (\textit{trump}), relative semantic stability (\textit{god}), and gradual association shifts linked to technological change (\textit{post}).

Overall, the findings demonstrate that word-centered semantic graphs provide a flexible and interpretable representation for studying semantic change, capturing both global semantic properties and fine-grained sense dynamics without relying on predefined sense inventories or embedding-space clustering.

\section*{Limitations}
\label{sec:limitations}

\paragraph{Application study rather than benchmark evaluation.}
We present an application-oriented analysis on three target words in a corpus of \textit{New York Times Magazine} articles. We did not evaluate our method on benchmarks corpora yet. Consequently, the paper emphasizes interpretability and sense tracking rather than competitive detection performance.

\paragraph{Sensitivity to hyperparameters and preprocessing.}
The neighborhood graphs depend on fixed hyperparameters ($k_i$, $k_c$, depth $L$). Different settings may change the number of clusters, the degree of fragmentation, and the apparent stability of senses. In particular, rare or newly emerging senses can be suppressed by frequency cutoffs.

\paragraph{Corpus and domain bias.}
The corpus is a single source with editorial and topical biases. Some observed changes are transient and event-driven, reflecting shifts in coverage rather than persistent changes in general language.

\bibliography{custom}

\appendix

\section{Data and pre-processing}
\label{app:data}

Our experiments are conducted on a diachronic corpus of articles from the \textit{New York Times Magazine}, covering the period from 1980 to 2017. To balance computational efficiency with temporal resolution, the corpus is partitioned into nine sub-corpora using a five-year sampling interval. For each period, a random sample of 20,000 articles is extracted, resulting in nine temporally ordered sub-corpora.

Each sub-corpus is pre-processed independently using the same pipeline. Text is lowercased and cleaned by removing punctuation, numerals, and stop words. All tokens are then lemmatized using the NLTK library \citep{Bird2009}. This normalization step reduces lexical sparsity and ensures that semantic comparisons across time are not confounded by surface-level morphological variation.

\section{Target word selection}
\label{app:targets}

To evaluate the proposed framework under different semantic conditions, we select a small set of target words based on three criteria: relative frequency, corpus relevance, and potential for semantic shift. Relative frequency captures how often a word appears across time and helps avoid spurious signals caused by rare or highly transient usage \citep{shoemark-etal-2019-room}. Corpus relevance reflects the extent to which a word is central to the themes and discourse of the \textit{New York Times Magazine}. Potential shift refers to the likelihood that a word exhibits multiple senses or meaning change over time \citep{giulianelli_analysing_2020}.

Based on these criteria, we analyze the target words \textit{trump}, \textit{god}, and \textit{post}. These words were chosen to represent contrasting semantic profiles: pronounced semantic shift driven by socio-political events (\textit{trump}), relative semantic stability (\textit{god}), and gradual sense evolution associated with technological change (\textit{post}).

Table~\ref{tab:target_words} summarizes the target words, including their dictionary definitions (Oxford English Dictionary), their relevance within the corpus, and their hypothesized potential for semantic shift.

\section{Relative frequency trends}
\label{app:frequency}

\begin{figure}
	\centering
	\includegraphics[width=0.47\textwidth]{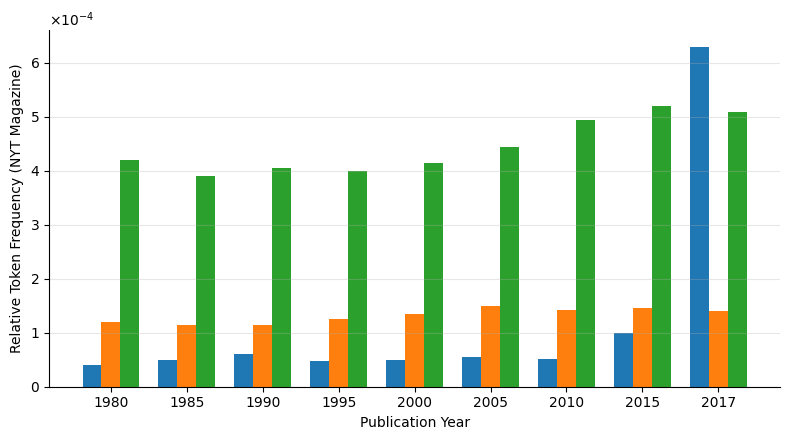}
	\caption{Relative frequency of target words \textit{trump} (Blue), \textit{god} (Orange), and \textit{post} (Green) across nine time periods (1980-2017) in the New York Times Magazine corpus. The targets exhibit distinct frequency trajectories, supporting their use as contrasting probes for semantic change analysis.}
	\label{fig:data}
\end{figure}

Figure~\ref{fig:data} shows the relative frequency of the target words \textit{trump}, \textit{god}, and \textit{post} across the nine sampled time periods. The three targets exhibit markedly different frequency trajectories, reflecting distinct patterns of semantic stability and change. These contrasting trends make them suitable probes for evaluating the robustness of the proposed framework under different diachronic conditions.

\section{Modeling setup and hyperparameters}
\label{app:setup}

All experiments are conducted using a fixed modeling configuration across target words and time periods to ensure comparability.

\paragraph{Static embeddings.}
Distributional representations are trained using the Skip-gram Word2Vec (SGNS) model, following the diachronic setup of \citet{hamilton-etal-2016-diachronic}. We use a context window size of 4, an embedding dimensionality of 300, and a minimum word frequency threshold of 50. Due to this frequency cutoff, some low-frequency words may not have distributional neighbors; in such cases, only contextually derived neighbors are included in the word neighborhood graph.

\paragraph{Contextual embeddings.}
Contextual representations are obtained by fine-tuning a \textit{RoBERTa-base} masked language model independently for each time slice. Fine-tuning follows the configuration of \citet{liu2019roberta}, using 30 epochs, a batch size of 32, a learning rate of $6\times10^{-4}$, and a masking probability of 0.15.

\paragraph{Graph construction.}
Word neighborhood graphs are constructed with a fixed depth of $L=2$. In the first layer, we select $k_i^{(1)}=3$ distributional neighbors and $k_c^{(1)}=6$ substitution neighbors of the target word. In the second layer, $k_i^{(2)}=1$ and $k_c^{(2)}=2$ neighbors are added for each node in the first layer.

\begin{table*}
	\small
    \centering
    \begin{tabular}{l P{0.28\linewidth} P{0.26\linewidth} P{0.28\linewidth}}
        \hline
        \textbf{Target} & \textbf{Definition} & \textbf{Relevance} & \textbf{Potential Shift} \\ 
        \hline
        \textit{trump} & A playing-card of that suit which for the time being ranks above the other three, or a proper noun. & Low relevance until the rise of Donald Trump as a prominent businessman and politician. & High potential (e.g. the 2016 U.S. presidential election). \\ 
        \hline
        \textit{god} & An entity regarded as having power over nature and human fortunes. & Consistent low relevance, used in discussions related to ethics, society, and religion. & Low potential as the definition for the term is set. \\ 
        \hline
        \textit{post} & An upright supporting structure, the system or act of mailing, a specific job position, or the act of publishing content online. & High relevance, used in technology, communication, and literal postal services. & Moderate potential, with the evolution of digital communication overtaking traditional postal methods. \\ 
        \hline
    \end{tabular}
	\caption{Target words selected for the application study, together with Oxford English Dictionary definitions, corpus-specific relevance, and potential for semantic shift. Words were chosen based on (i) relative frequency variation across time to avoid purely transient change \citep{shoemark-etal-2019-room}, (ii) relevance to the themes and topics of the New York Times Magazine corpus, and (iii) potential for semantic shift driven by polysemy and historically conditioned usage \citep{giulianelli_analysing_2020}.}
    \label{tab:target_words}
\end{table*}

\begin{figure*}
	\centering
	\begin{subfigure}[b]{0.48\textwidth}
		\centering
		\includegraphics[width=\textwidth]{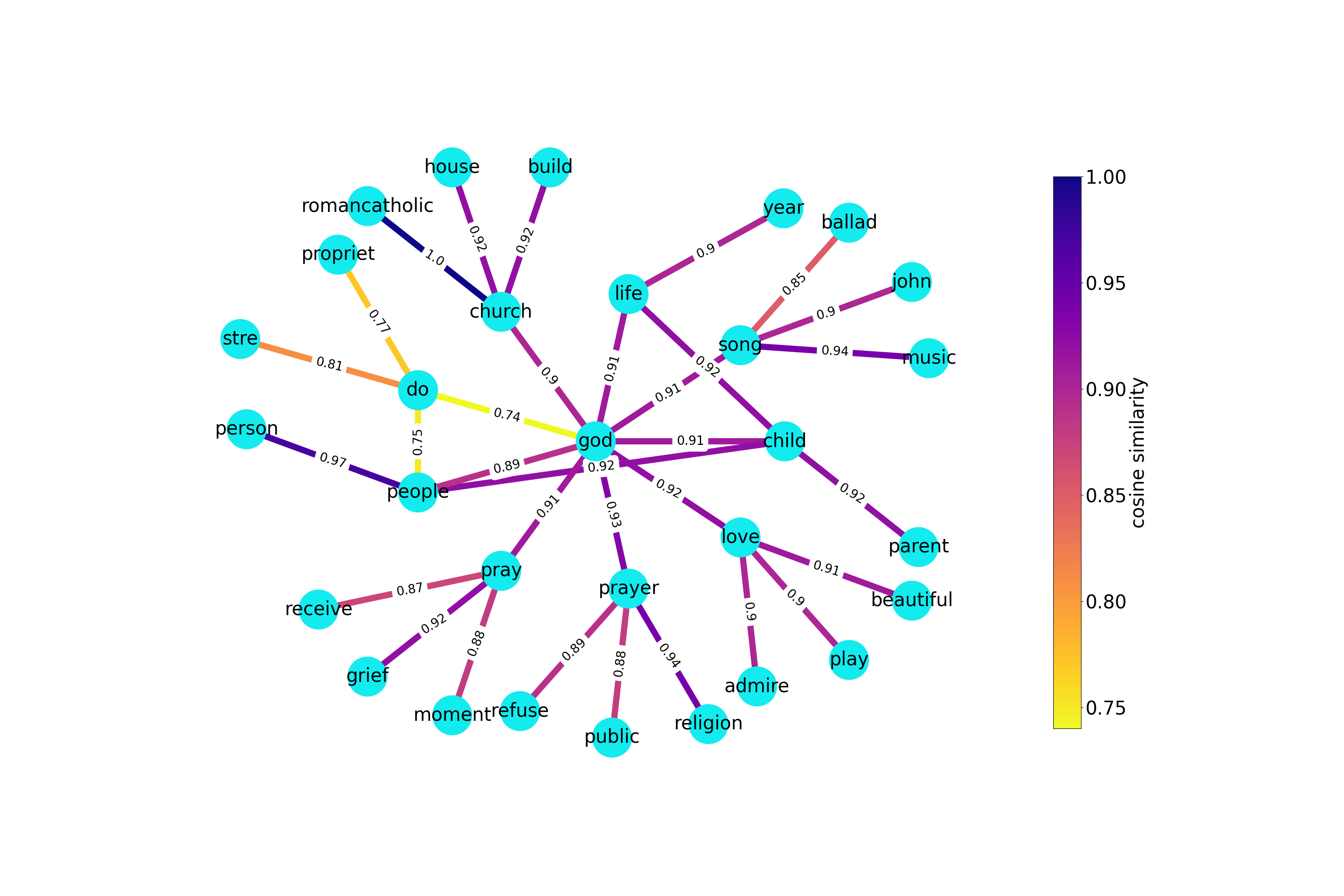}
		\caption{Year 1980}
		\label{fig:god_enet_2017}
	\end{subfigure}
	\hfill
	\begin{subfigure}[b]{0.48\textwidth}
	   \centering
	   \includegraphics[width=\textwidth]{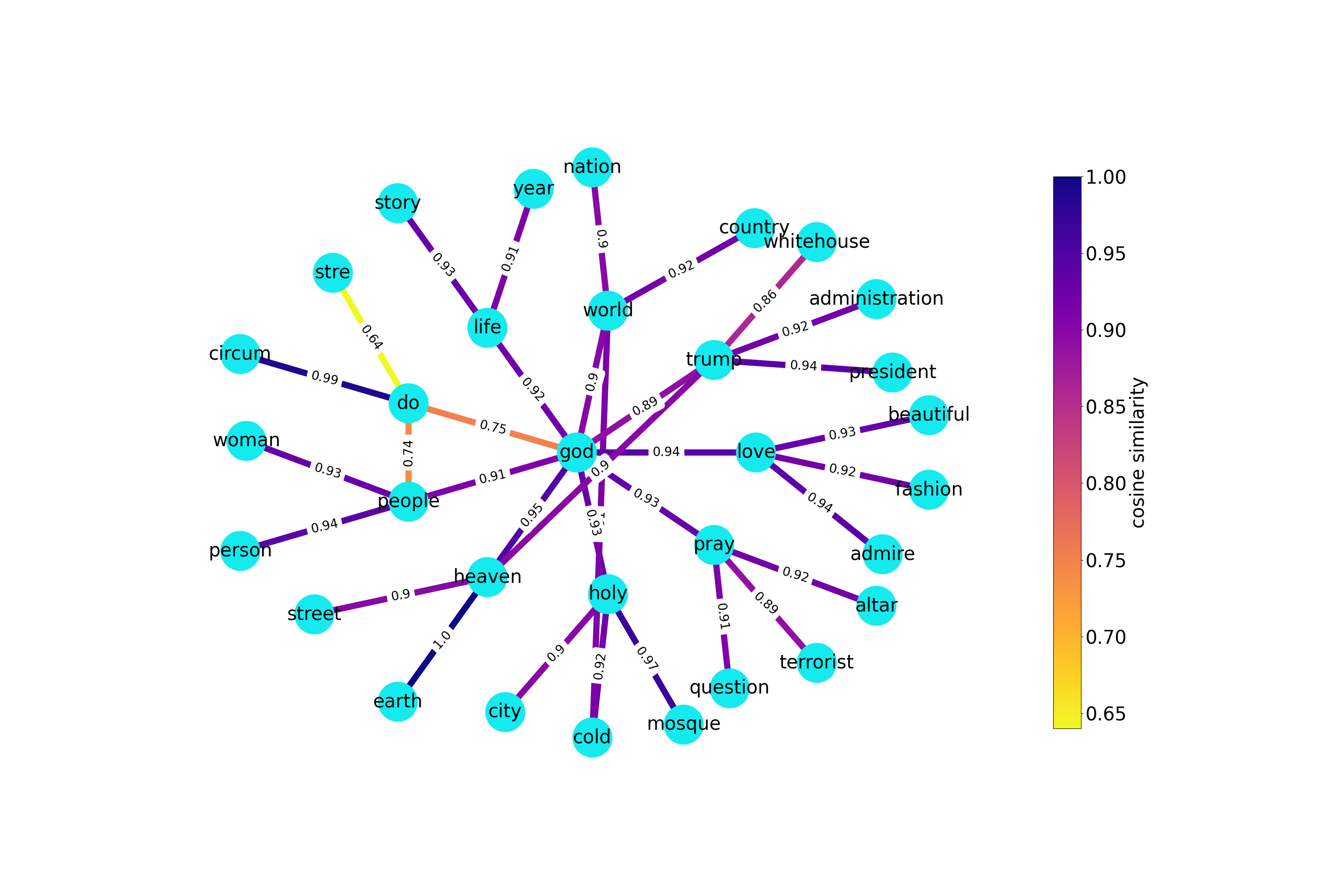}
	   \caption{Year 2017}
		\label{fig:god_net_2017}
    \end{subfigure}
	\caption{Word-centered semantic networks of \textit{god} in 1980 and 2017. Edge color encodes cosine similarity between contextual node embeddings. The networks exhibit stable structure and uniformly high similarity across time, consistent with a unified semantic interpretation and limited polysemy.}
	\label{fig:god_networks}
\end{figure*}

\begin{figure*}
	\centering
	\begin{subfigure}[b]{0.48\textwidth}
		\centering
		\includegraphics[width=\textwidth]{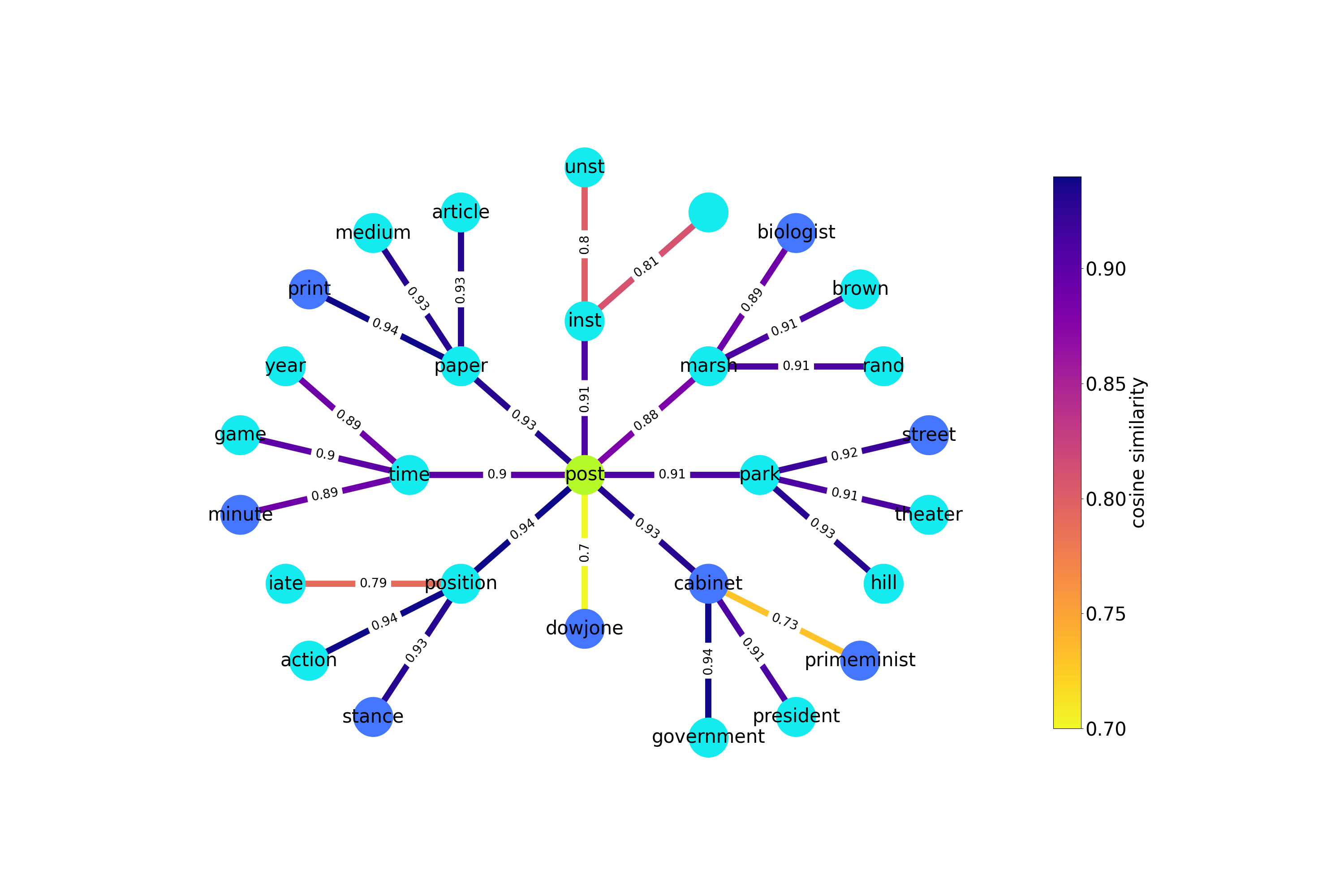}
		\caption{Year 1980}
		\label{fig:post_enet_2017}
	\end{subfigure}
	\hfill
	\begin{subfigure}[b]{0.48\textwidth}
	   \centering
	   \includegraphics[width=\textwidth]{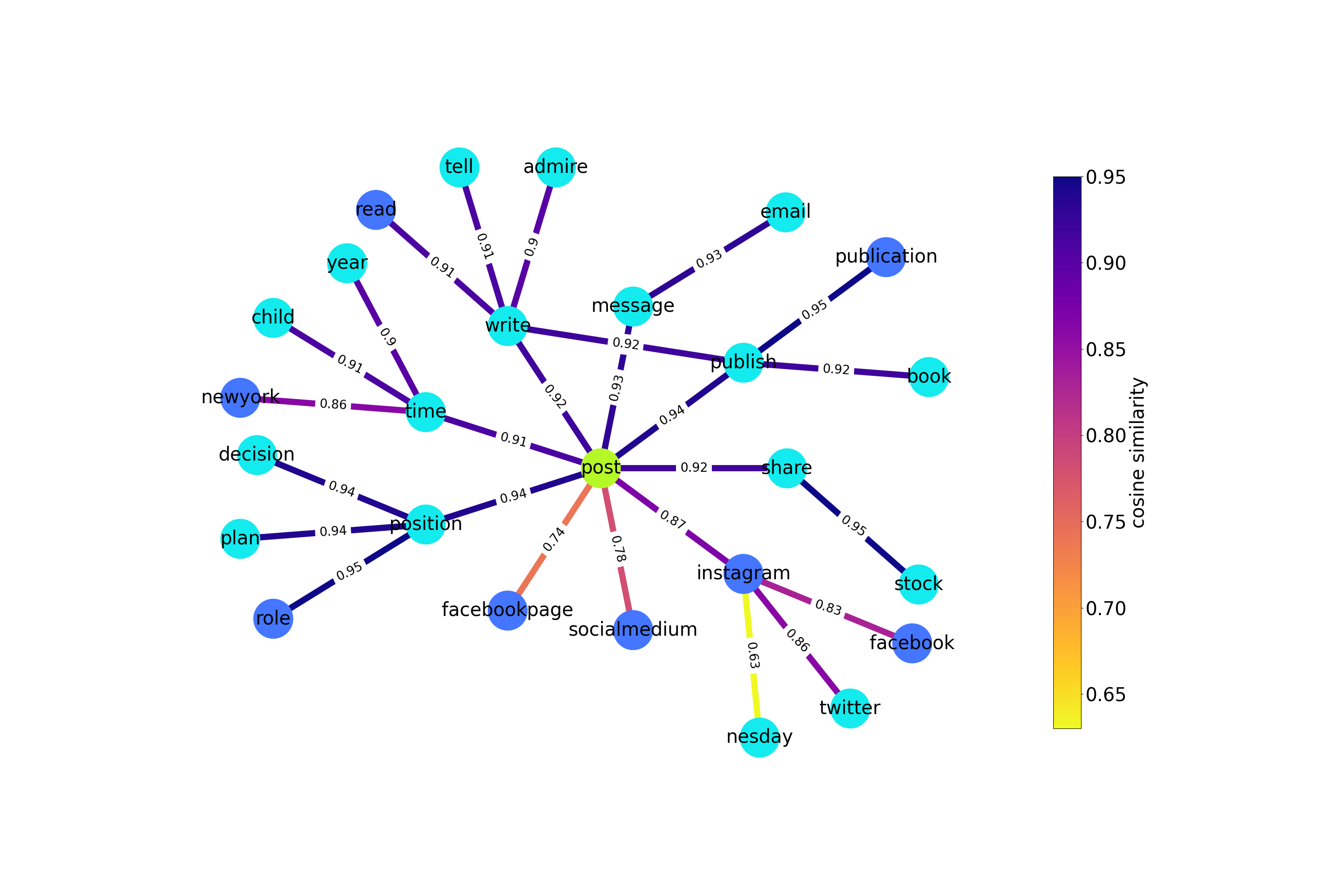}
	   \caption{Year 2017}
		\label{fig:post_net_2017}
    \end{subfigure}
	\caption{Word-centered semantic networks of \textit{post} in 1980 and 2017. Edge color encodes cosine similarity between contextual node embeddings. Later networks introduce technology-related neighbors (e.g., social media platforms), indicating a gradual shift in associative structure rather than abrupt sense replacement.}

	\label{fig:post_networks}
\end{figure*}

\begin{figure*}
	\centering
	\begin{subfigure}[b]{0.48\textwidth}
		\centering
		\includegraphics[width=\textwidth]{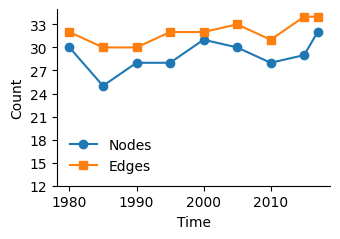}
		\caption{\textit{god}}
		\label{fig:god_ts}
	\end{subfigure}
	\hfill
	\begin{subfigure}[b]{0.48\textwidth}
	   \centering
	   \includegraphics[width=\textwidth]{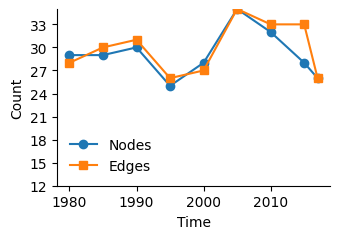}
	   \caption{\textit{post}}
		\label{fig:post_ts}
    \end{subfigure}
	\caption{Graph properties over time for \textit{god} (left) and \textit{post} (right). Blue lines denote number of nodes; orange lines denote number of edges. The stable profile of \textit{god} contrasts with increased connectivity for \textit{post} during periods of technological change.}

	\label{fig:tss}
\end{figure*}

\begin{figure*}
	\centering
	\begin{subfigure}[b]{0.48\textwidth}
		\centering
		\includegraphics[width=\textwidth]{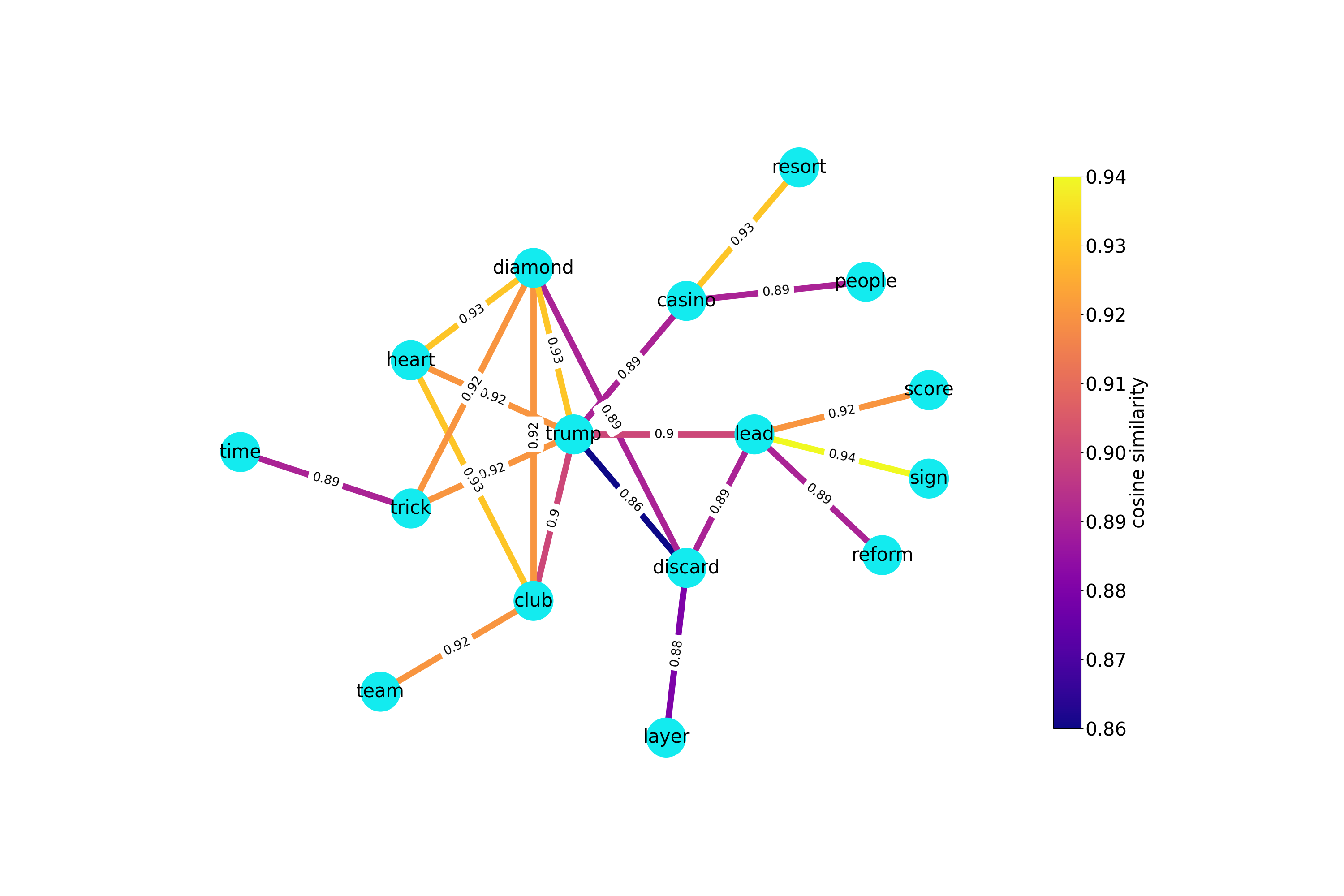}
		\caption{Year 1985}
		\label{fig:trump_net_1985}
	\end{subfigure}
	\hfill
	\begin{subfigure}[b]{0.48\textwidth}
	   \centering
	   \includegraphics[width=\textwidth]{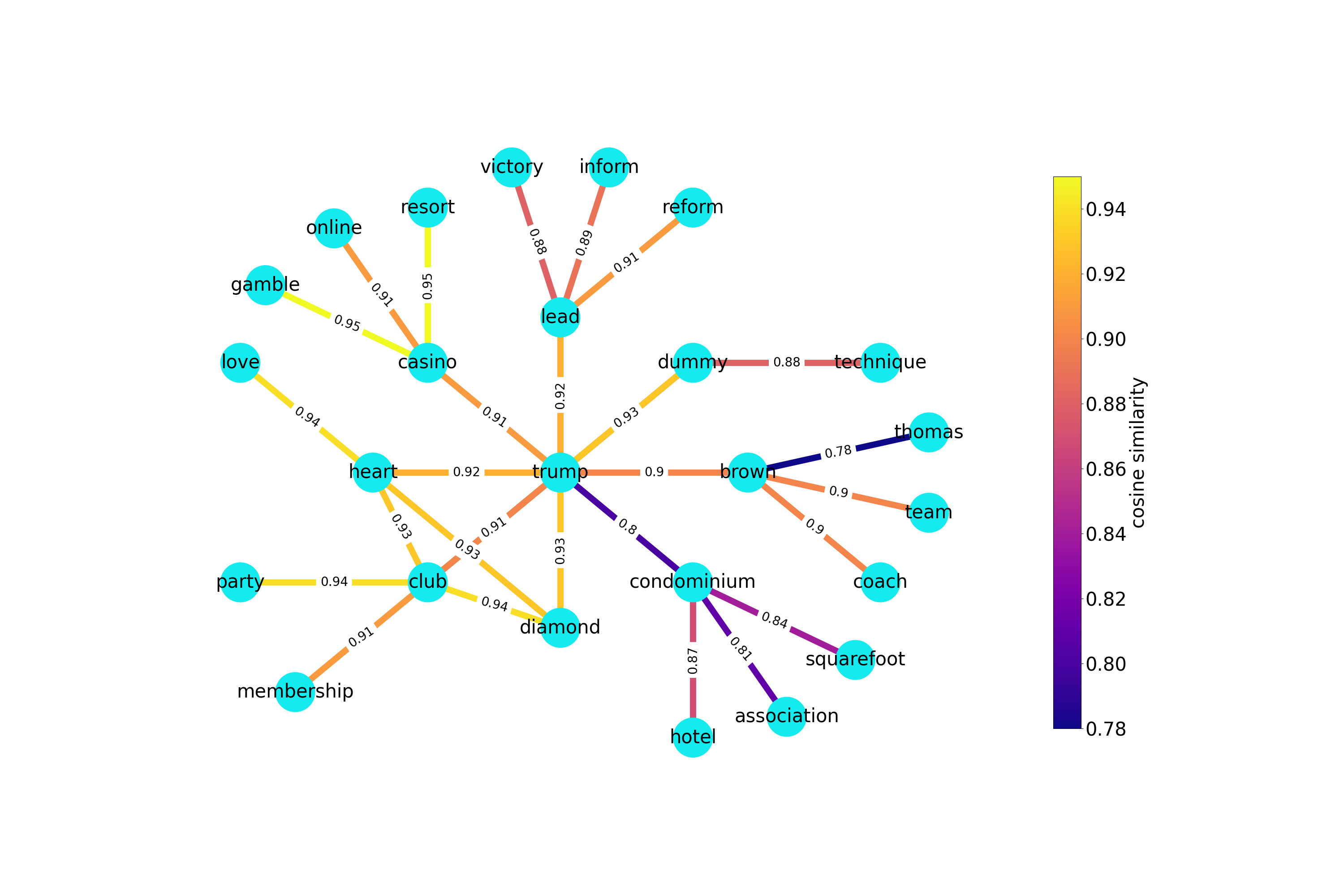}
	   \caption{Year 2005}
	   \label{fig:trump_net_2005}
    \end{subfigure}
    \vfill
	\begin{subfigure}[b]{0.48\textwidth}
	   \centering
	   \includegraphics[width=\textwidth]{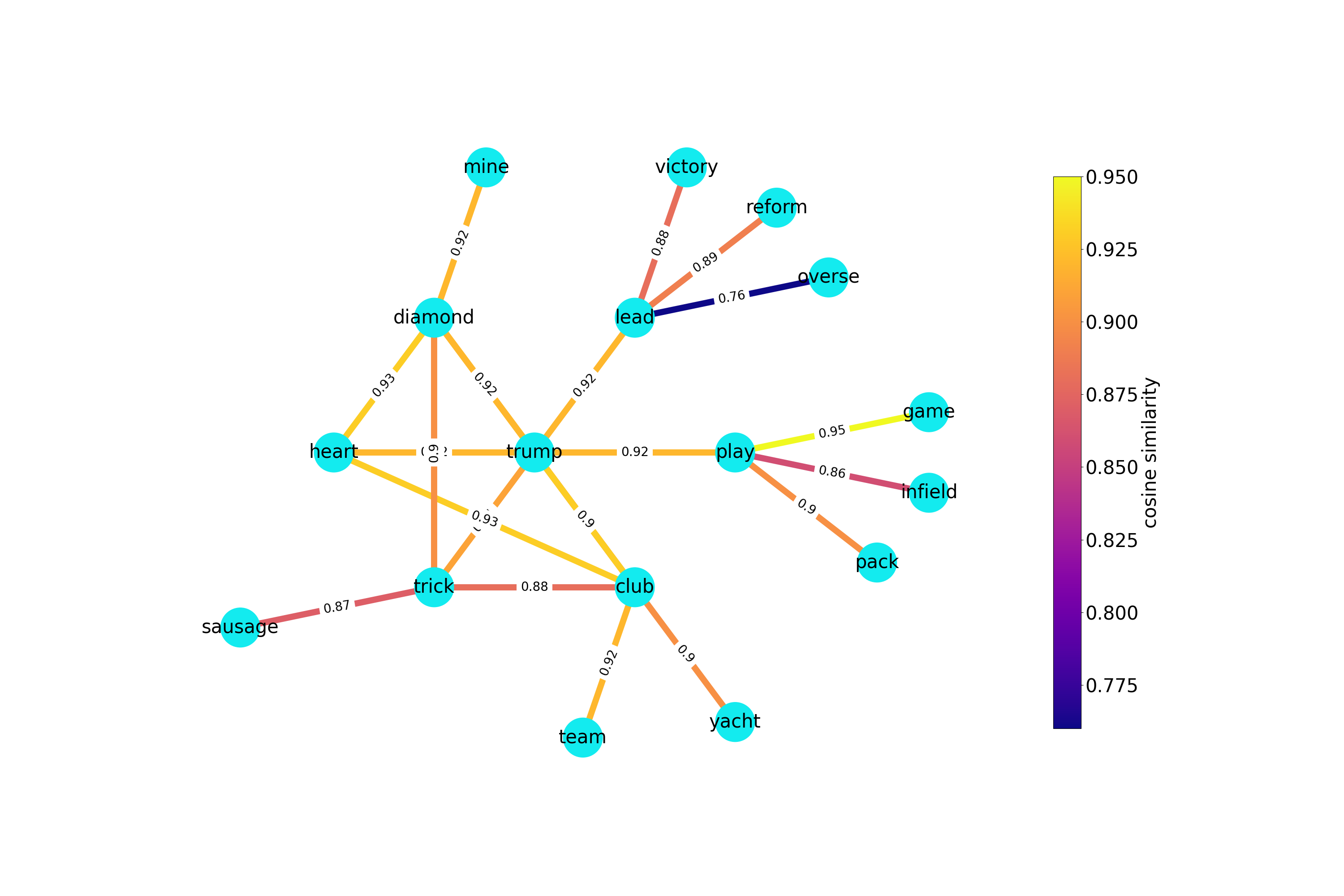}
	   \caption{Year 2010}
	   \label{fig:trump_net_2010}
    \end{subfigure}
	\hfill
	\begin{subfigure}[b]{0.48\textwidth}
	   \centering
	   \includegraphics[width=\textwidth]{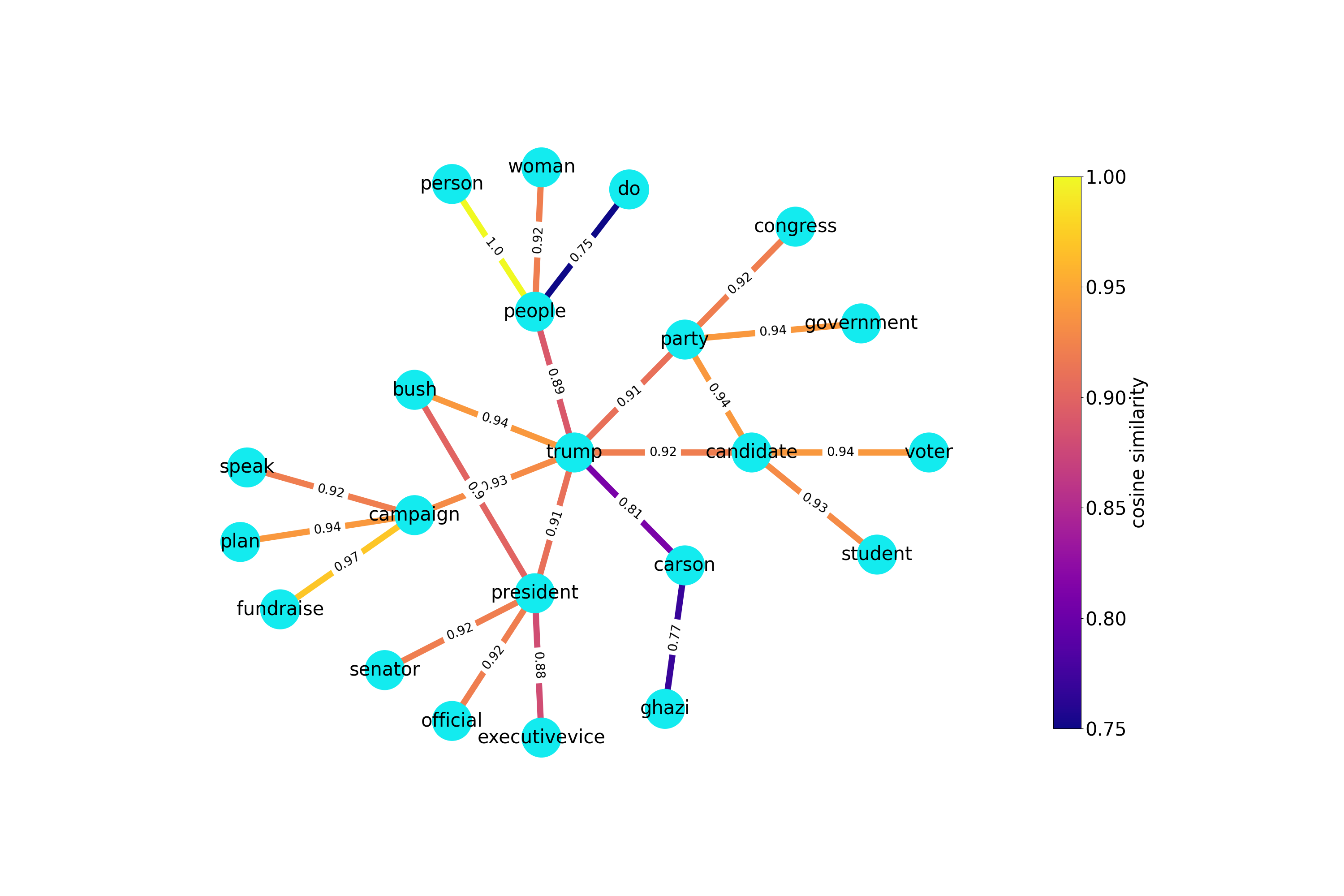}
	   \caption{Year 2015}
	   \label{fig:trump_net_2015}
    \end{subfigure}
    \caption{Evolution of the word-centered semantic network of \textit{trump} across four time periods. Early networks reflect the literal card-game sense, mid-period networks introduce business-related communities, and later networks are dominated by political associations, illustrating transient and substantive semantic shifts.}
    \label{fig:trump_other_nets}
\end{figure*}

\begin{figure*}
	\centering
	\begin{subfigure}[b]{0.48\textwidth}
		\centering
		\includegraphics[width=\textwidth]{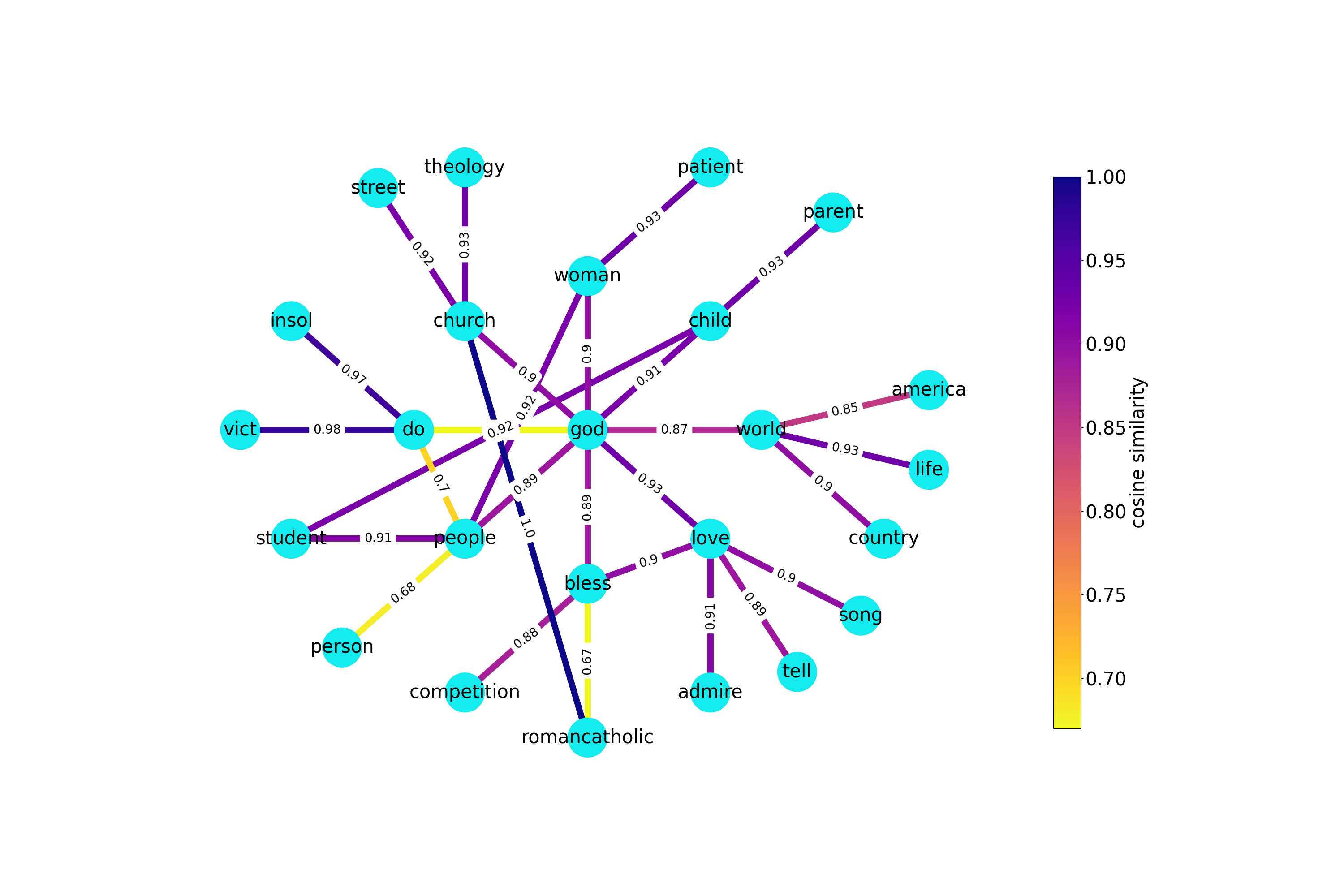}
		\caption{Year 1985}
		\label{fig:god_net_1985}
	\end{subfigure}
	\hfill
	\begin{subfigure}[b]{0.48\textwidth}
	   \centering
	   \includegraphics[width=\textwidth]{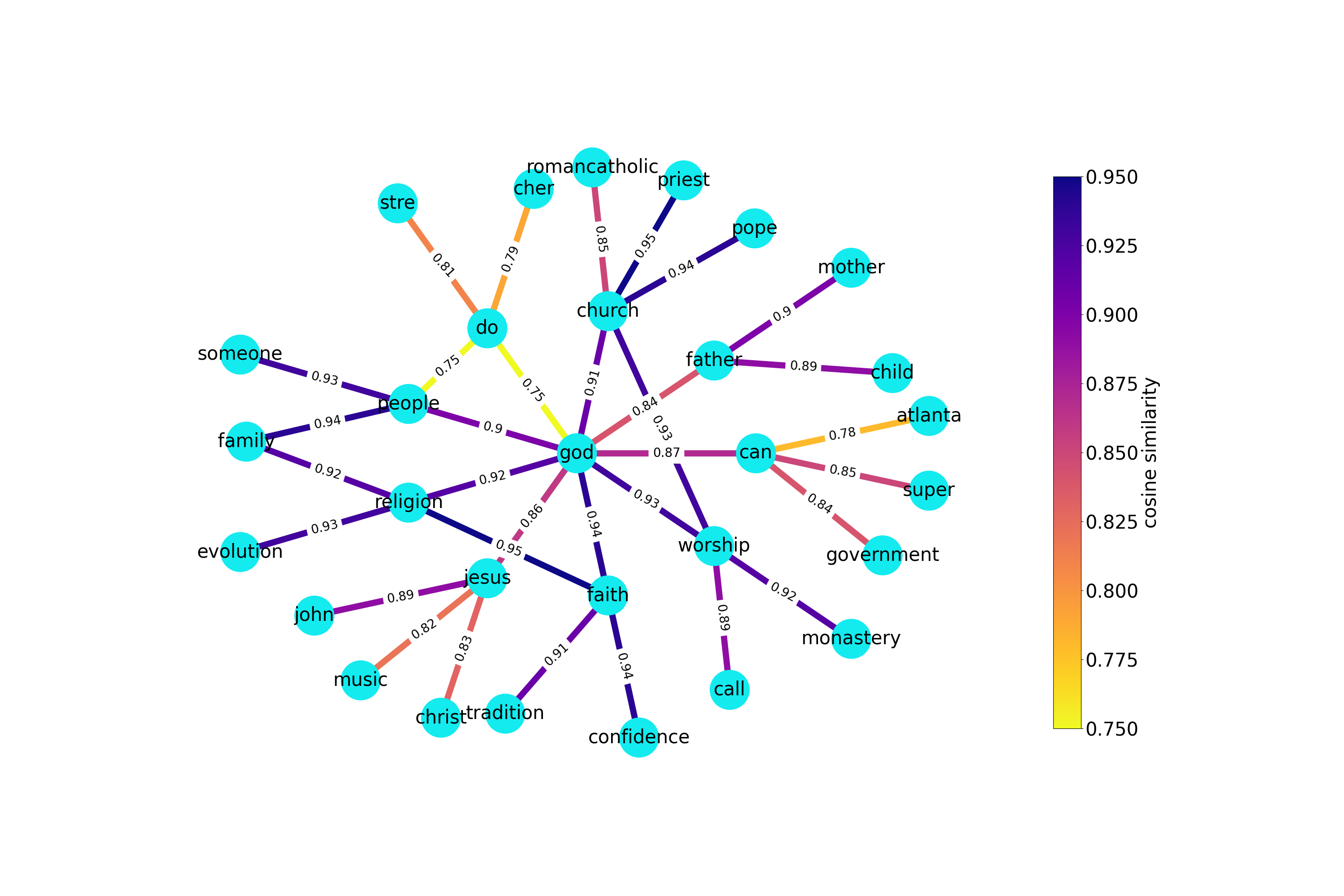}
	   \caption{Year 2005}
	   \label{fig:god_net_2005}
    \end{subfigure}
    \vfill
	\begin{subfigure}[b]{0.48\textwidth}
	   \centering
	   \includegraphics[width=\textwidth]{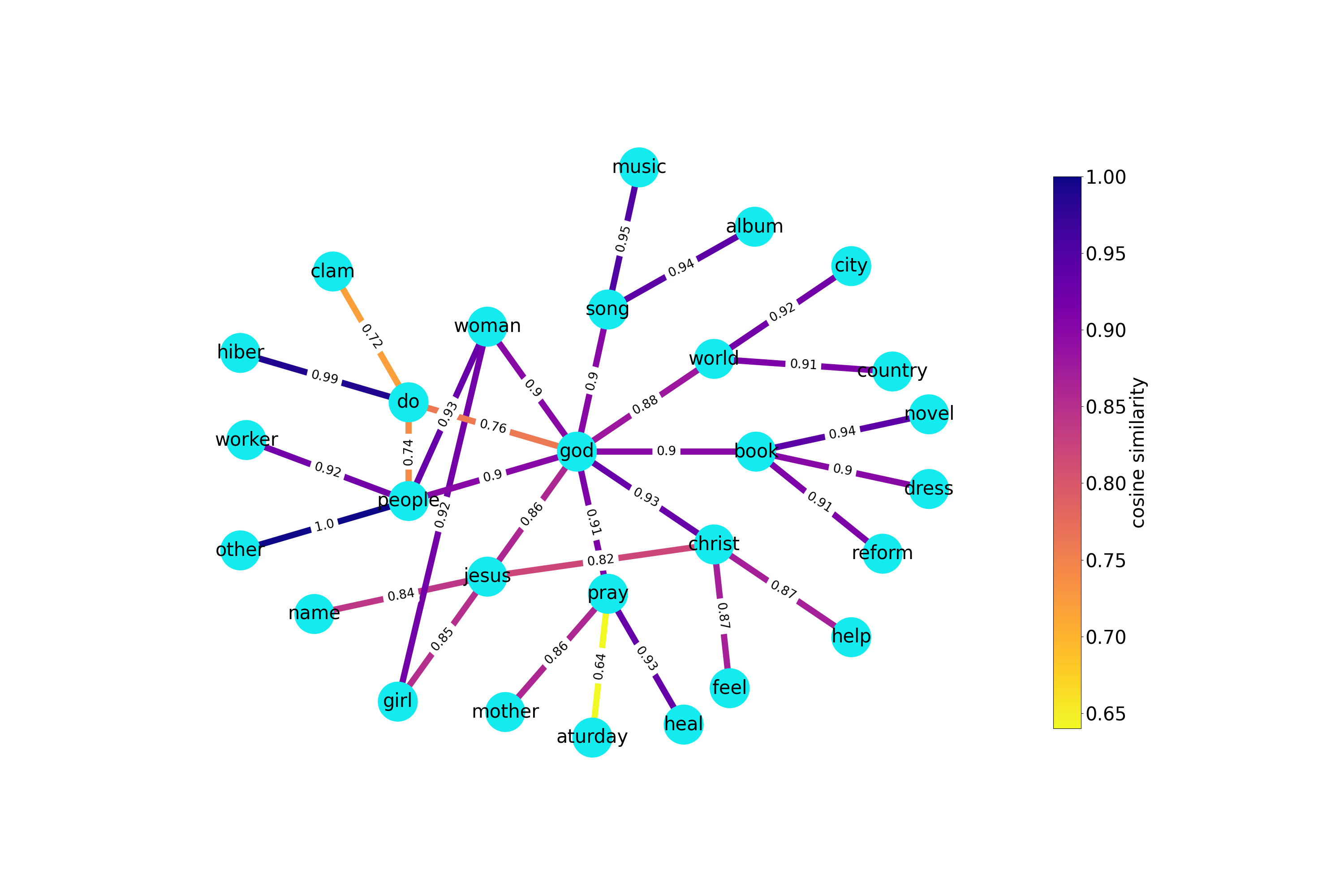}
	   \caption{Year 2010}
	   \label{fig:god_net_2010}
    \end{subfigure}
	\hfill
	\begin{subfigure}[b]{0.48\textwidth}
	   \centering
	   \includegraphics[width=\textwidth]{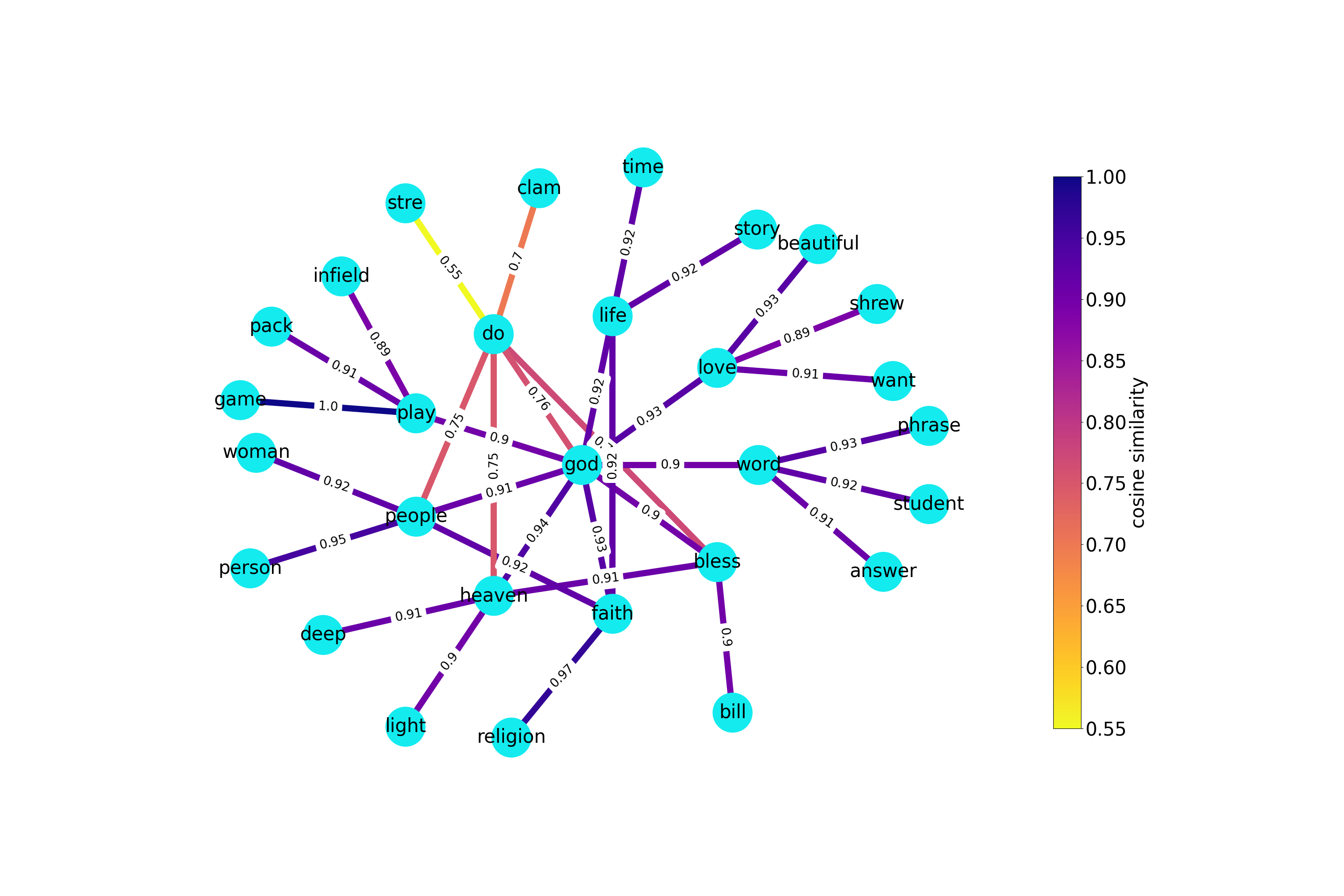}
	   \caption{Year 2015}
	   \label{fig:god_net_2015}
    \end{subfigure}
    \caption{Word-centered semantic networks of \textit{god} across multiple time periods. Despite minor fluctuations in connectivity, the networks maintain consistent node composition and structure, supporting semantic stability over time.}
    \label{fig:god_other_nets}
\end{figure*}

\begin{figure*}
	\centering
	\begin{subfigure}[b]{0.48\textwidth}
		\centering
		\includegraphics[width=\textwidth]{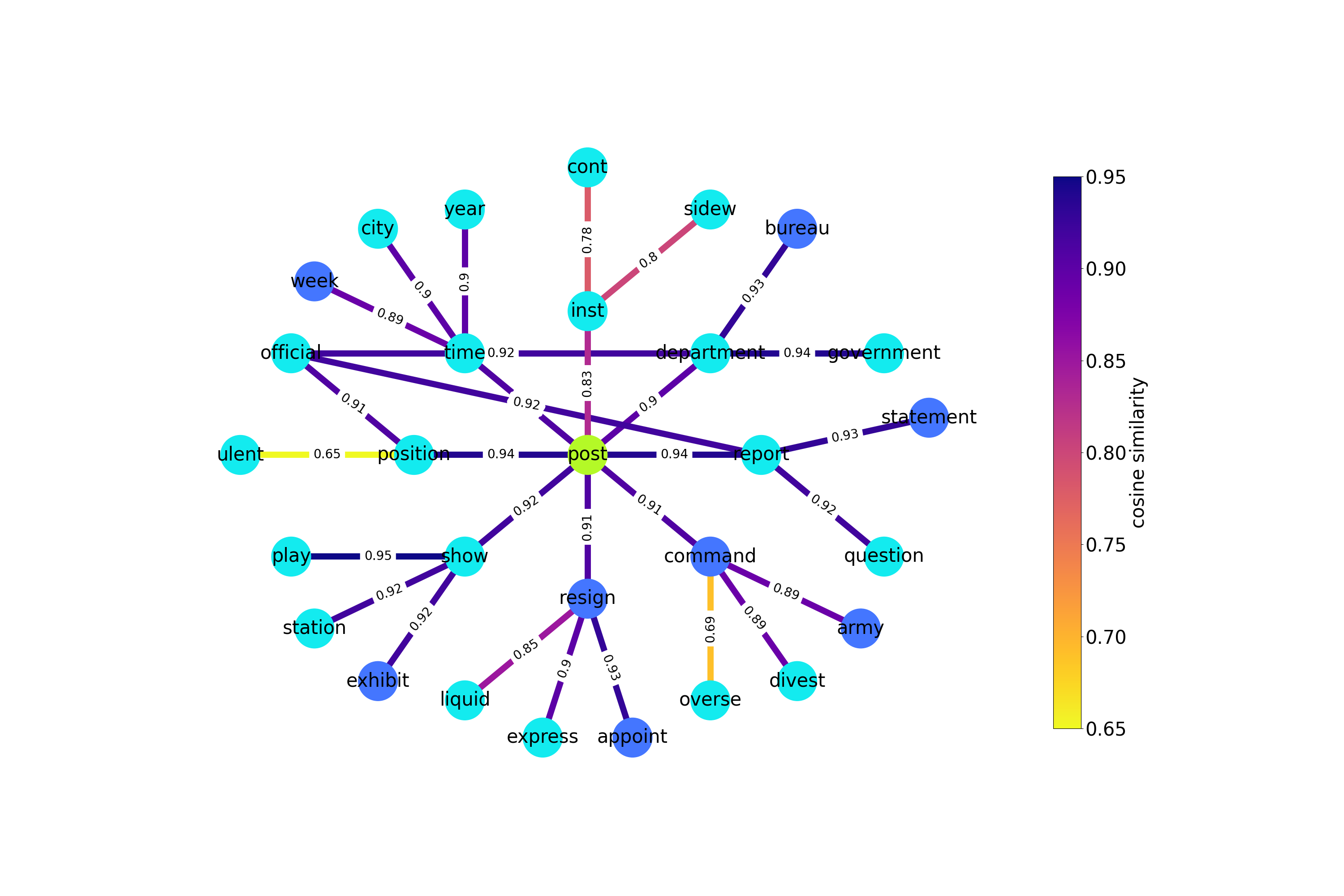}
		\caption{Year 1985}
		\label{fig:post_net_1985}
	\end{subfigure}
	\hfill
	\begin{subfigure}[b]{0.48\textwidth}
	   \centering
	   \includegraphics[width=\textwidth]{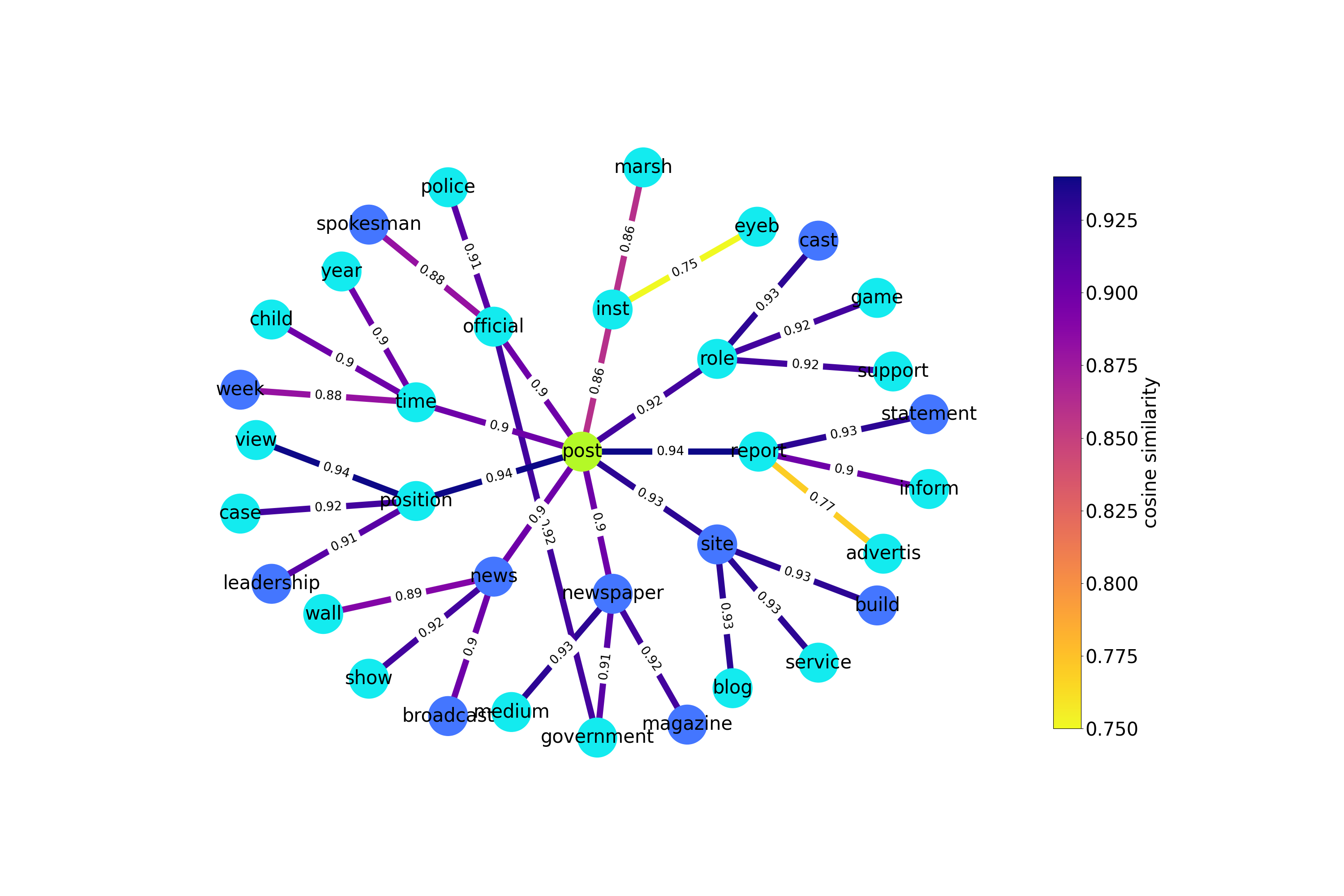}
	   \caption{Year 2005}
	   \label{fig:post_net_2005}
    \end{subfigure}
    \vfill
	\begin{subfigure}[b]{0.48\textwidth}
	   \centering
	   \includegraphics[width=\textwidth]{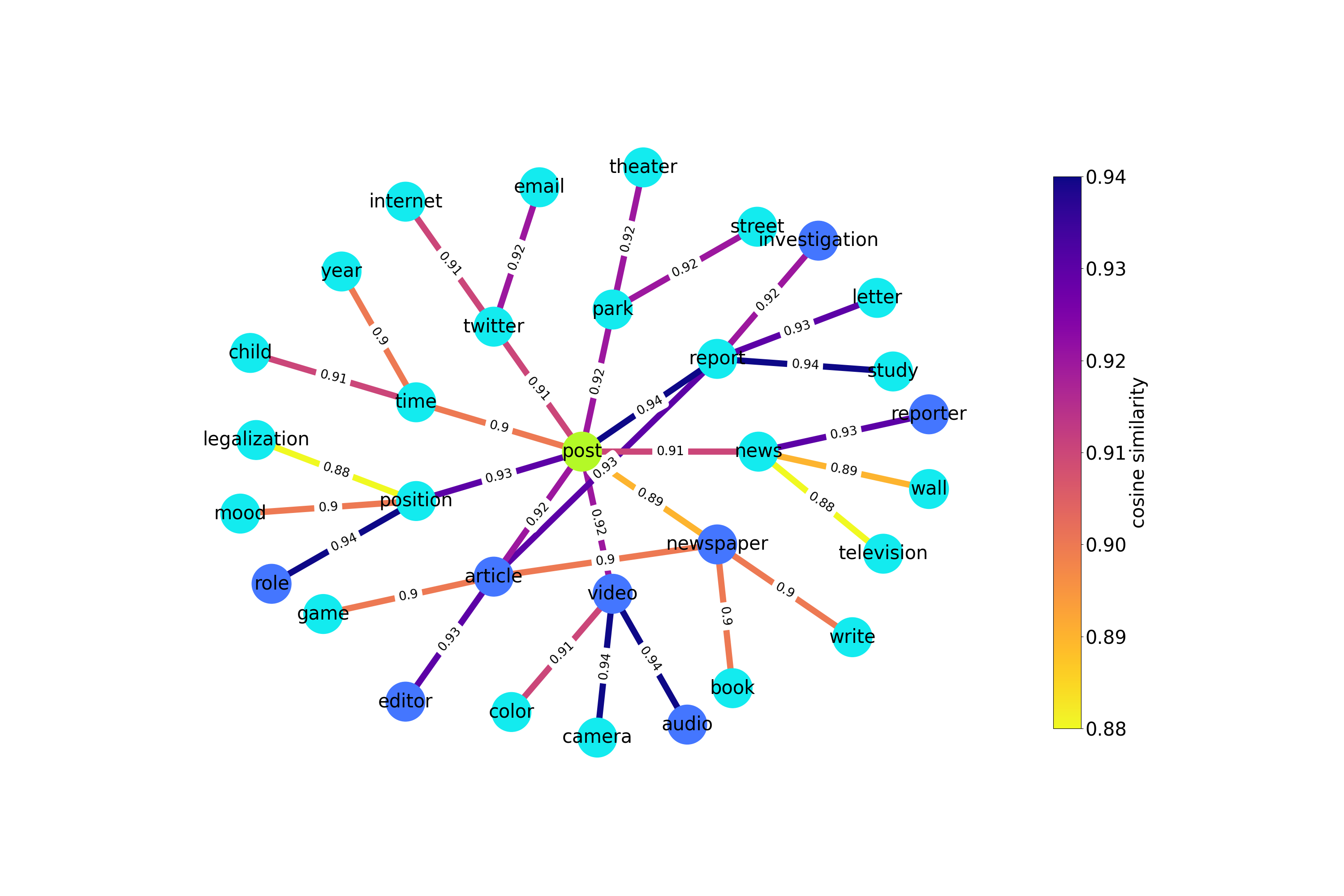}
	   \caption{Year 2010}
	   \label{fig:post_net_2010}
    \end{subfigure}
	\hfill
	\begin{subfigure}[b]{0.48\textwidth}
	   \centering
	   \includegraphics[width=\textwidth]{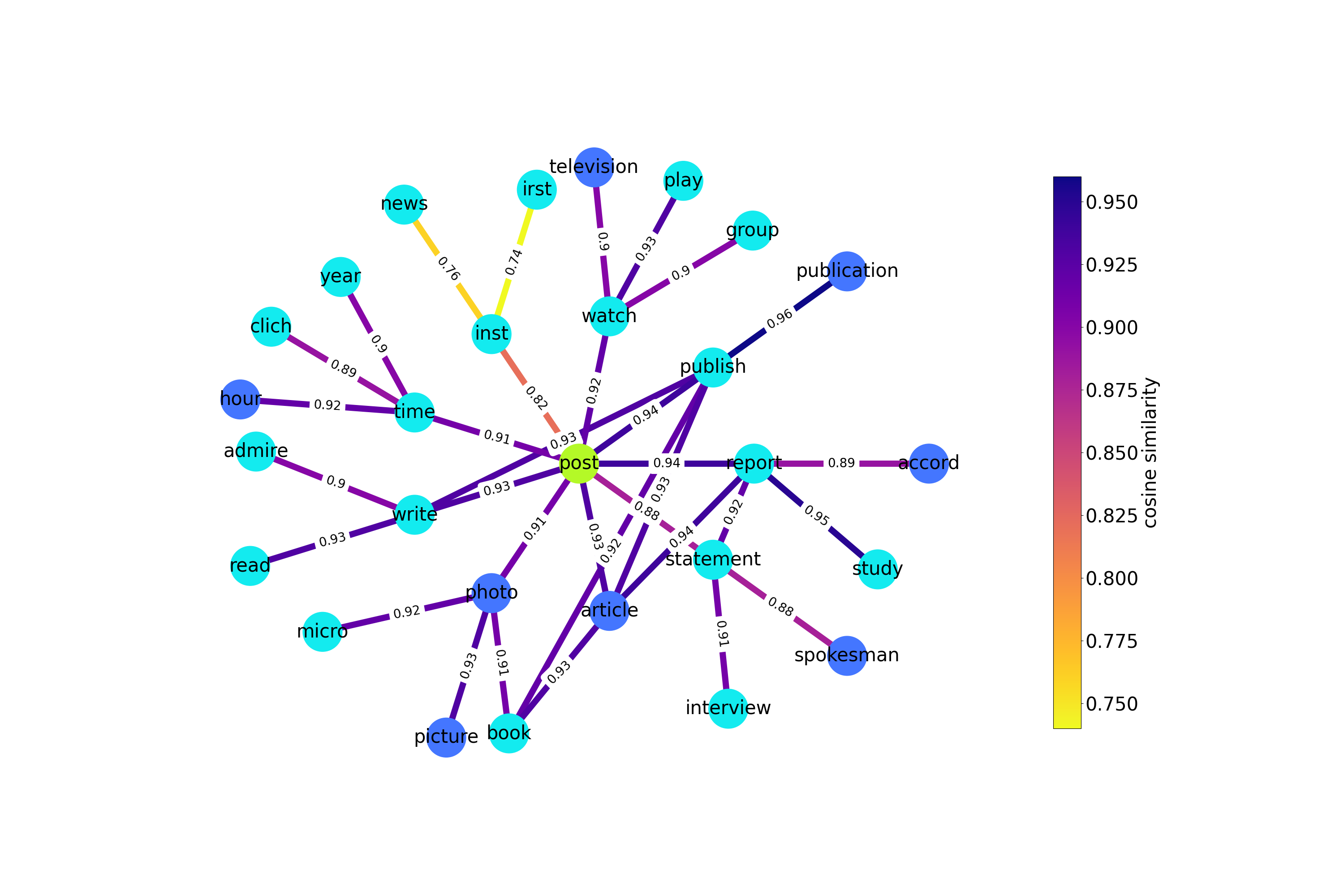}
	   \caption{Year 2015}
	   \label{fig:post_net_2015}
    \end{subfigure}
    \caption{Word-centered semantic networks of \textit{post} across multiple time periods. Later networks introduce technology-oriented neighbors, reflecting gradual associative broadening linked to digital communication.}
    \label{fig:post_other_nets}
\end{figure*}

\begin{figure*}
	\centering
	\begin{subfigure}[b]{0.48\textwidth}
		\centering
		\includegraphics[width=\textwidth]{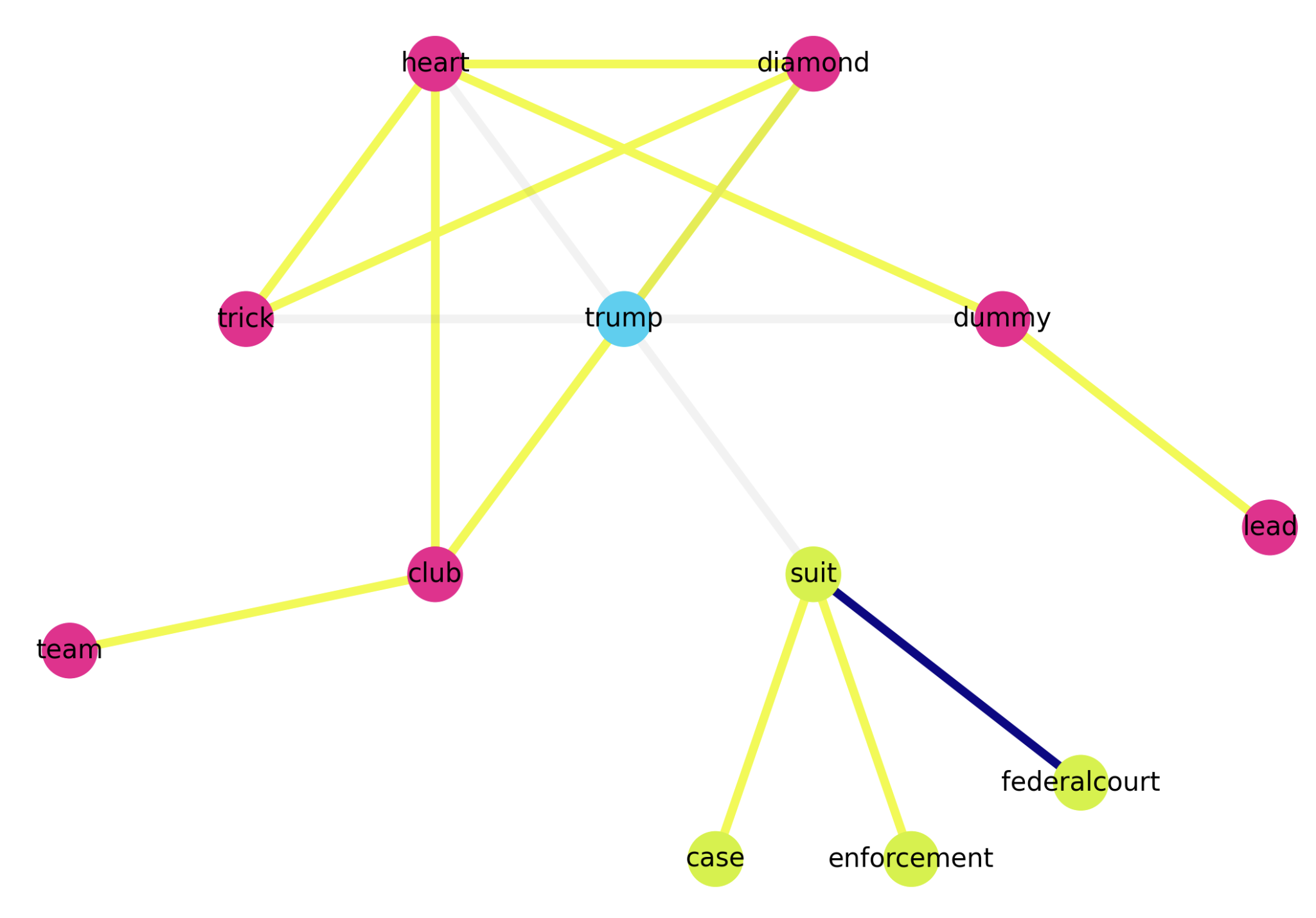}
		\caption{Year 1980}
		\label{fig:trump_cluster_1980}
	\end{subfigure}
	\hfill
	\begin{subfigure}[b]{0.48\textwidth}
	   \centering
	   \includegraphics[width=\textwidth]{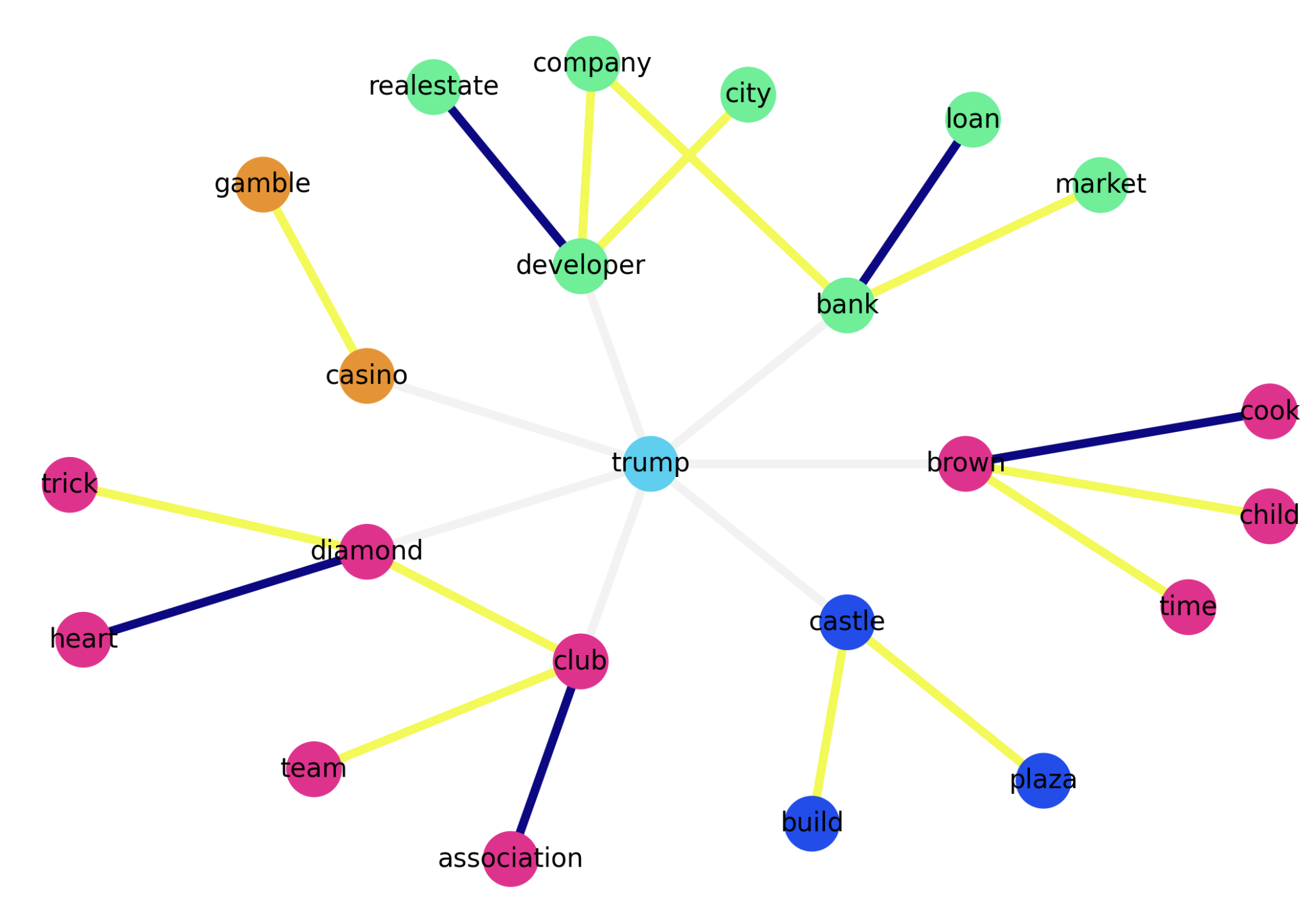}
	   \caption{Year 1990}
	   \label{fig:trump_cluster_1990}
    \end{subfigure}
	\vfill
	\begin{subfigure}[b]{0.48\textwidth}
	   \centering
	   \includegraphics[width=\textwidth]{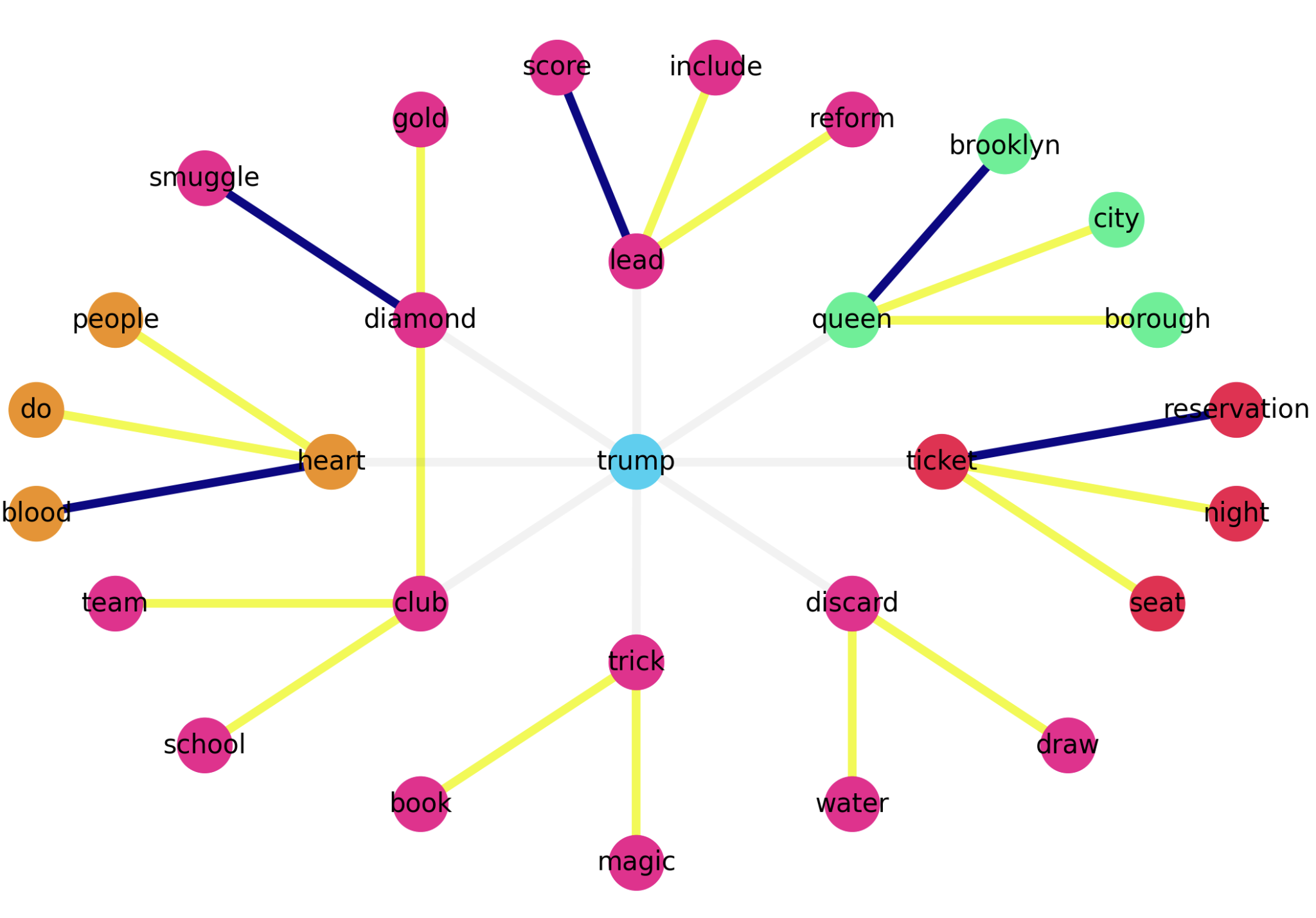}
	   \caption{Year 2000}
	   \label{fig:trump_cluster_2000}
    \end{subfigure}
	\hfill
	\begin{subfigure}[b]{0.48\textwidth}
	   \centering
	   \includegraphics[width=\textwidth]{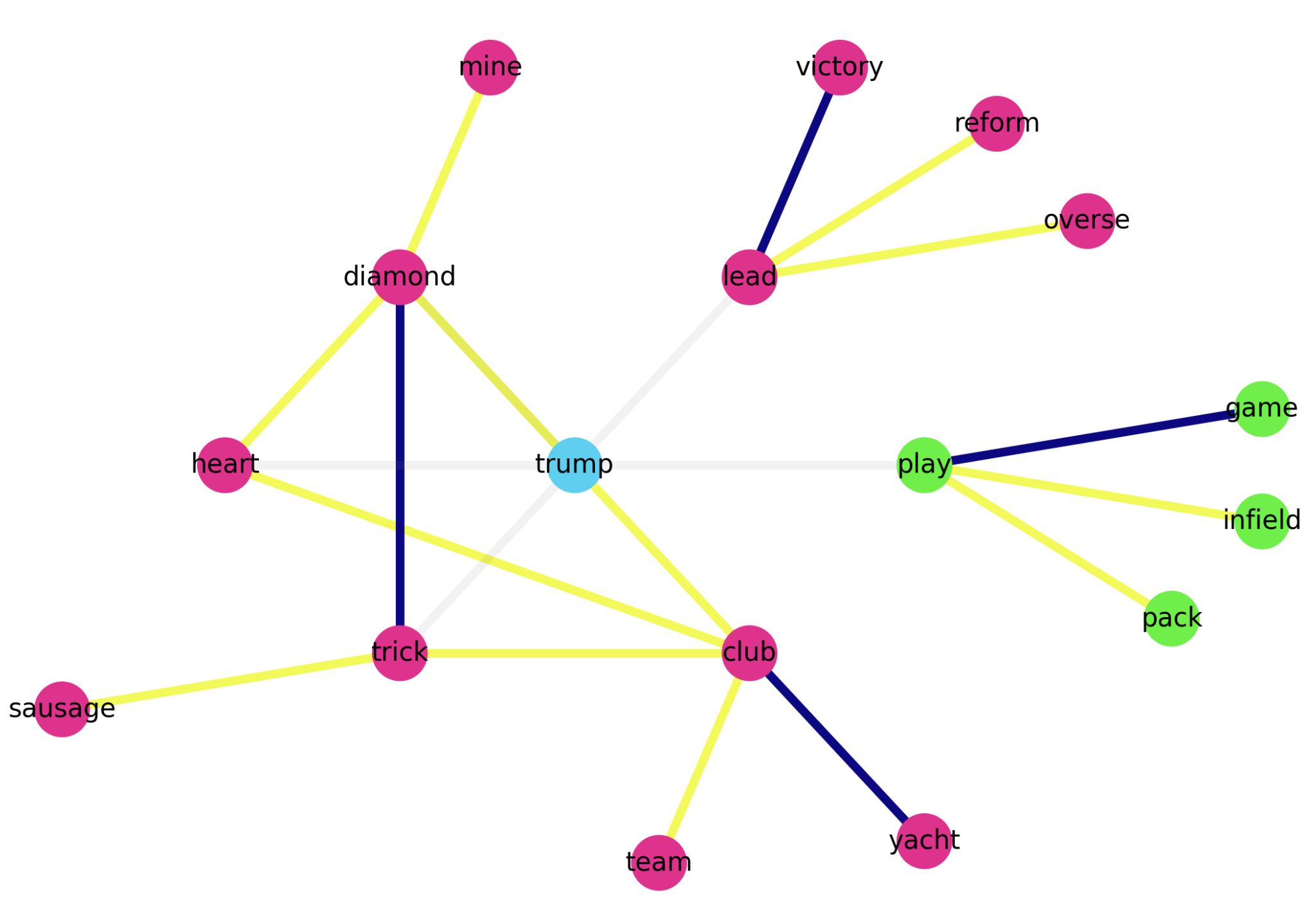}
	   \caption{Year 2010}
	   \label{fig:trump_cluster_2010}
    \end{subfigure}
	\vfill
	\begin{subfigure}[b]{0.48\textwidth}
	   \centering
	   \includegraphics[width=\textwidth]{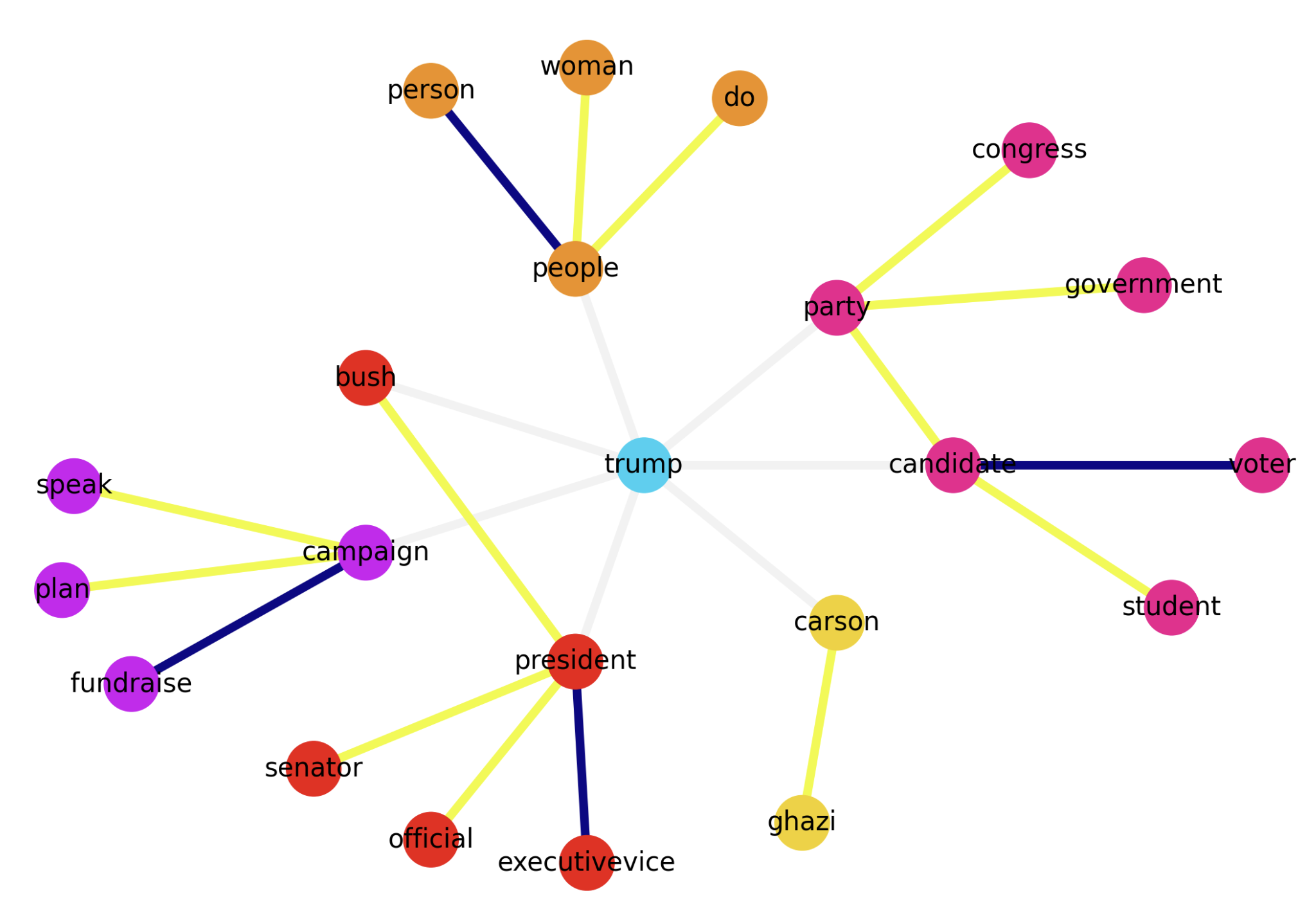}
	   \caption{Year 2015}
	   \label{fig:trump_cluster_2015}
    \end{subfigure}
	\hfill
	\begin{subfigure}[b]{0.48\textwidth}
	   \centering
	   \includegraphics[width=\textwidth]{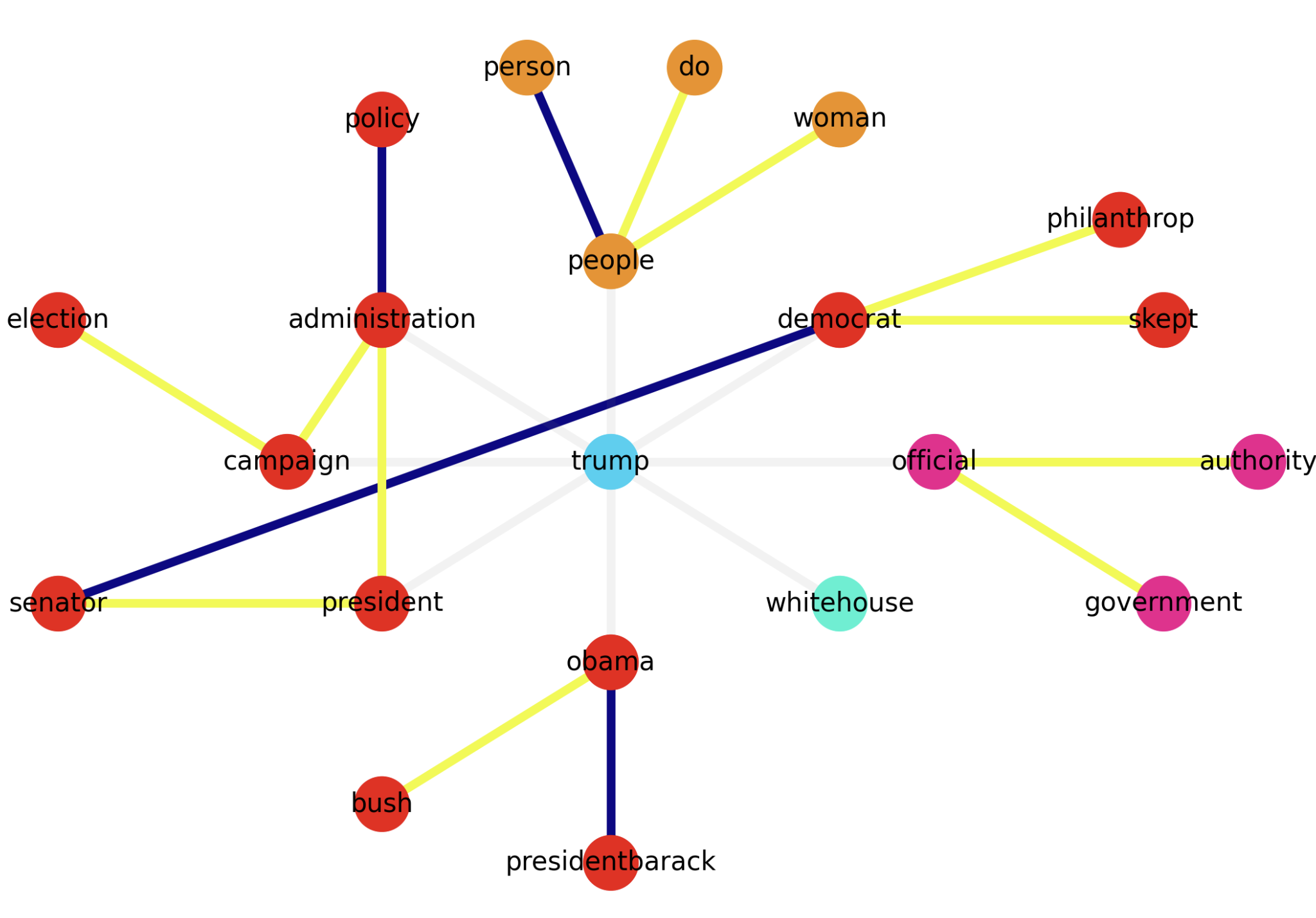}
	   \caption{Year 2017}
	   \label{fig:trump_cluster_2017}
	\end{subfigure}
	\caption{Peripheral connectivity clustering of the \textit{trump} neighborhood across time. Node colors denote distinct clusters (11 in total); gray edges correspond to removed edges incident to the target word. The emergence and disappearance of disconnected clusters reflects periods of heightened polysemy and sense competition.}
    \label{fig:trump_clusters}
\end{figure*}

\begin{figure*}
	\centering
	\begin{subfigure}[b]{0.48\textwidth}
		\centering
		\includegraphics[width=\textwidth]{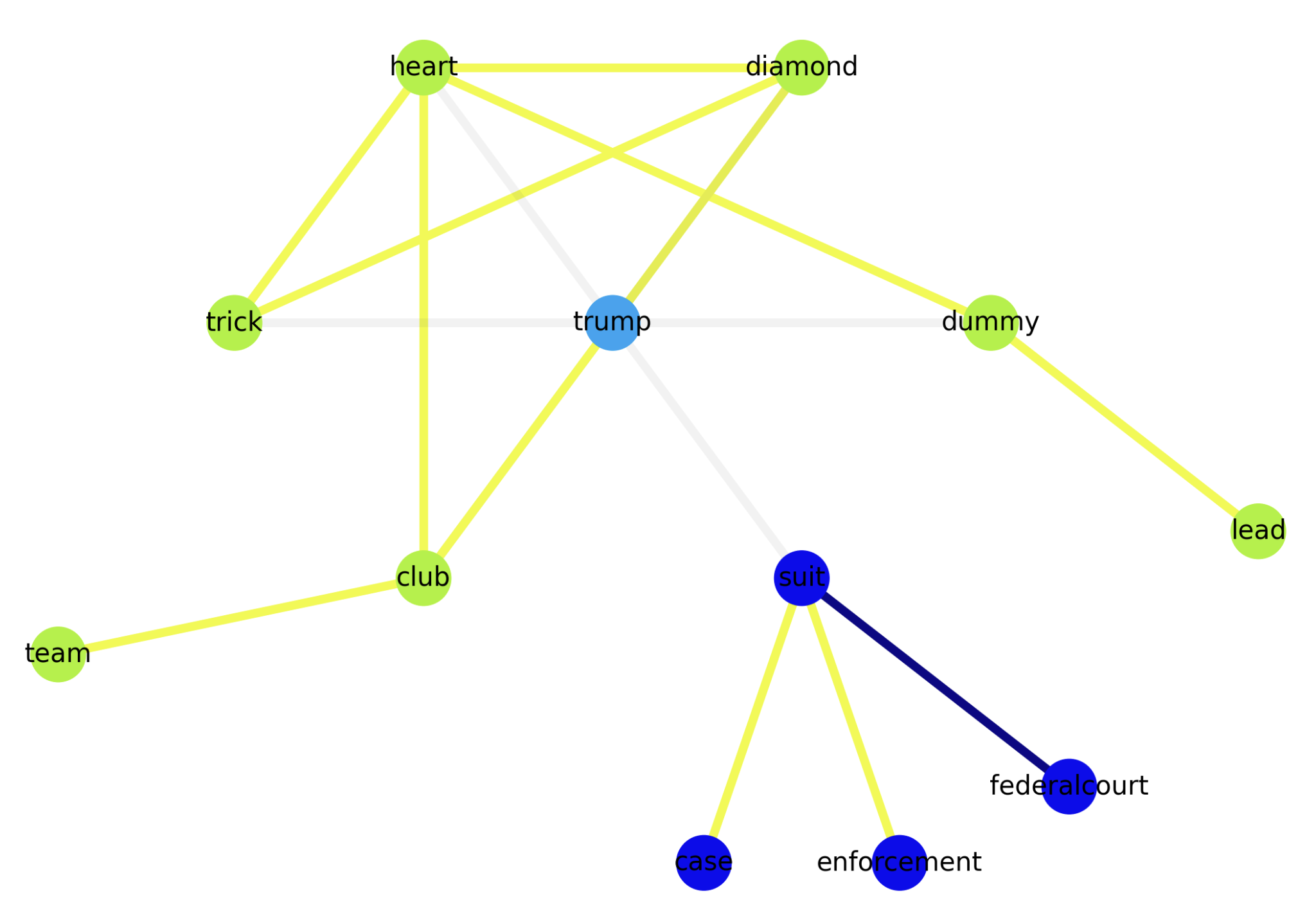}
		\caption{Year 1980}
		\label{fig:trump_cluster_refined_1980}
	\end{subfigure}
	\hfill
	\begin{subfigure}[b]{0.48\textwidth}
	   \centering
	   \includegraphics[width=\textwidth]{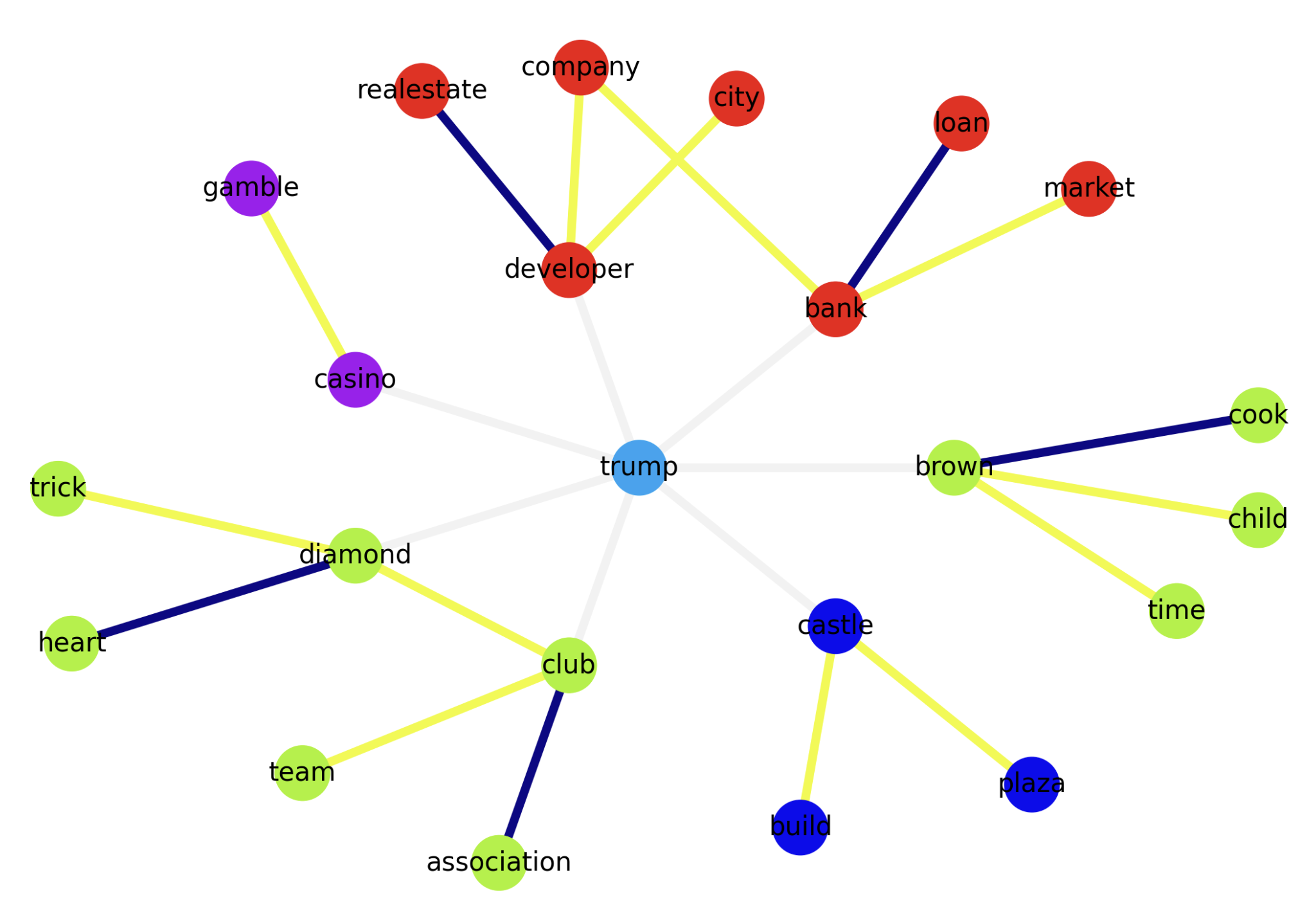}
	   \caption{Year 1990}
	   \label{fig:trump_cluster_refined_1990}
    \end{subfigure}
	\vfill
	\begin{subfigure}[b]{0.48\textwidth}
	   \centering
	   \includegraphics[width=\textwidth]{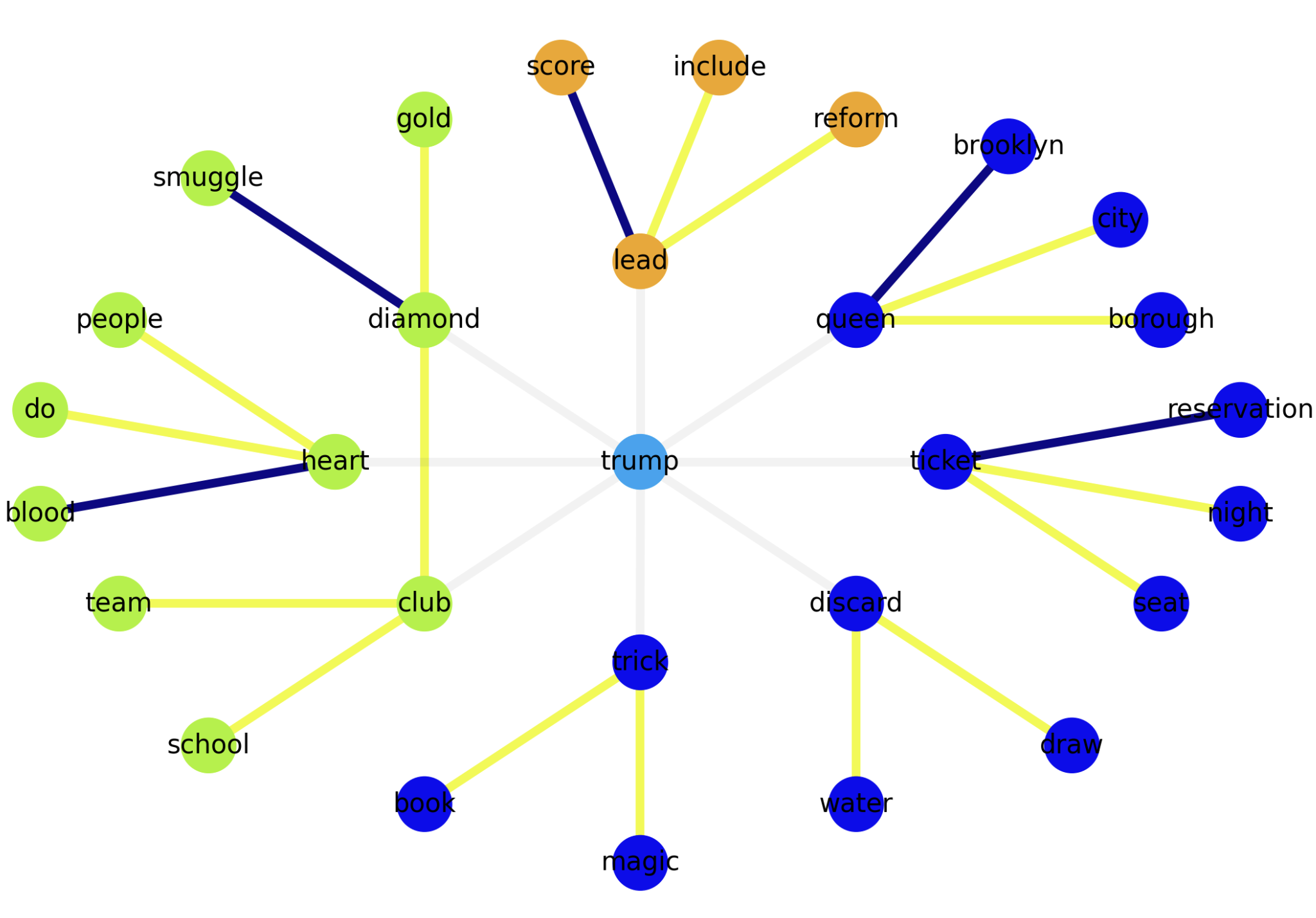}
	   \caption{Year 2000}
	   \label{fig:trump_cluster_refined_2000}
    \end{subfigure}
	\hfill
	\begin{subfigure}[b]{0.48\textwidth}
	   \centering
	   \includegraphics[width=\textwidth]{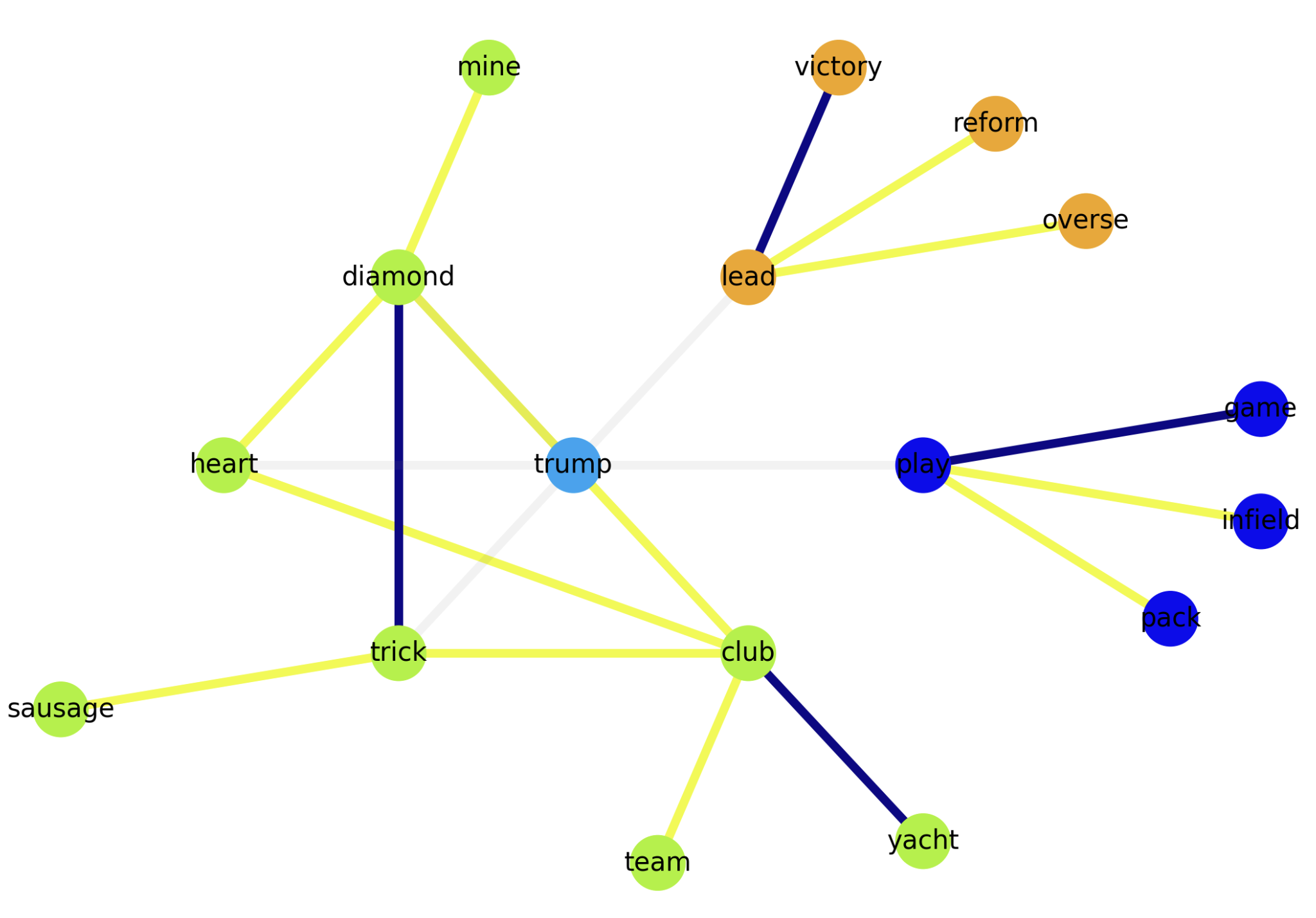}
	   \caption{Year 2010}
	   \label{fig:trump_cluster_refined_2010}
    \end{subfigure}
	\vfill
	\begin{subfigure}[b]{0.48\textwidth}
	   \centering
	   \includegraphics[width=\textwidth]{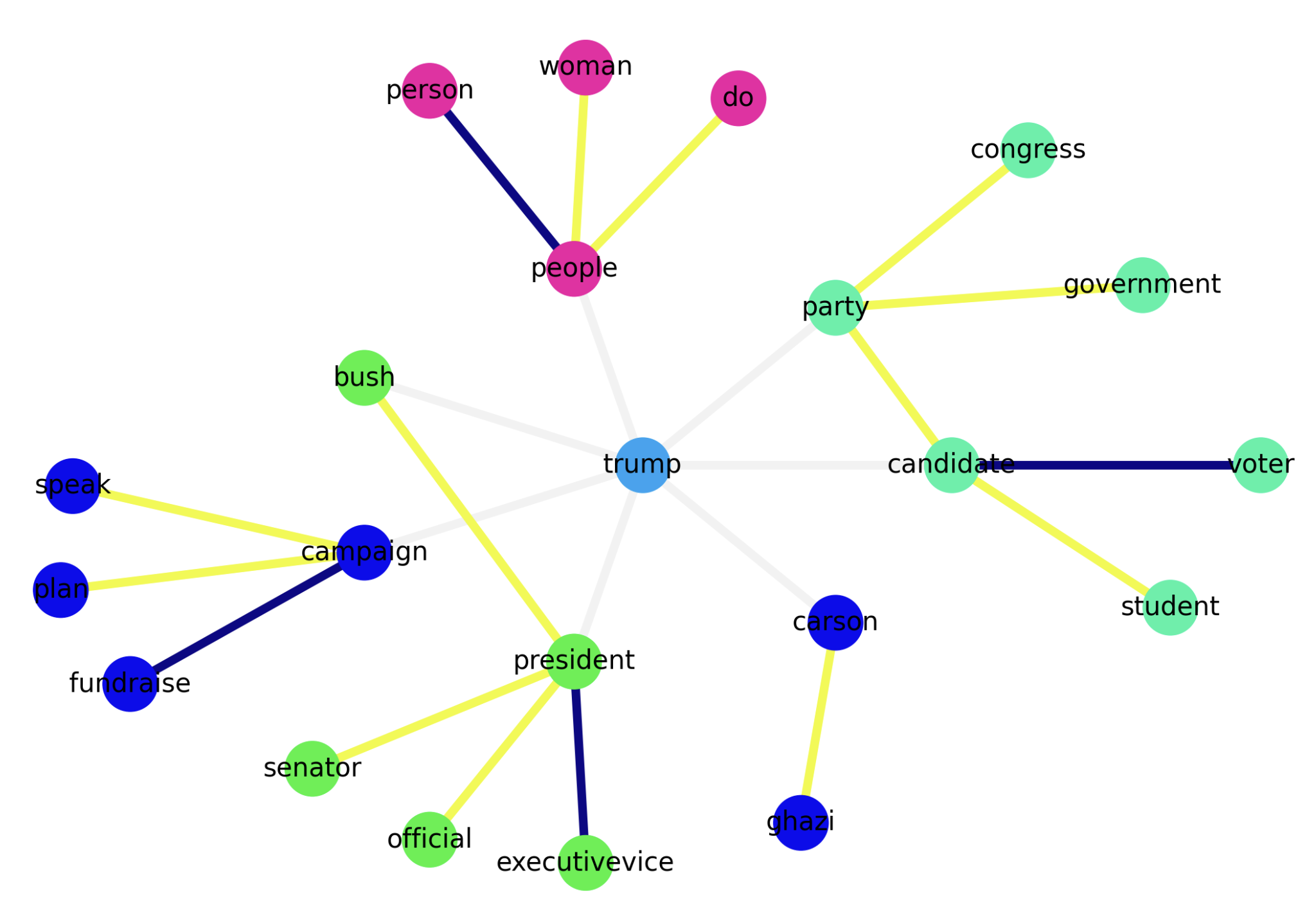}
	   \caption{Year 2015}
	   \label{fig:trump_cluster_refined_2015}
    \end{subfigure}
	\hfill
	\begin{subfigure}[b]{0.48\textwidth}
	   \centering
	   \includegraphics[width=\textwidth]{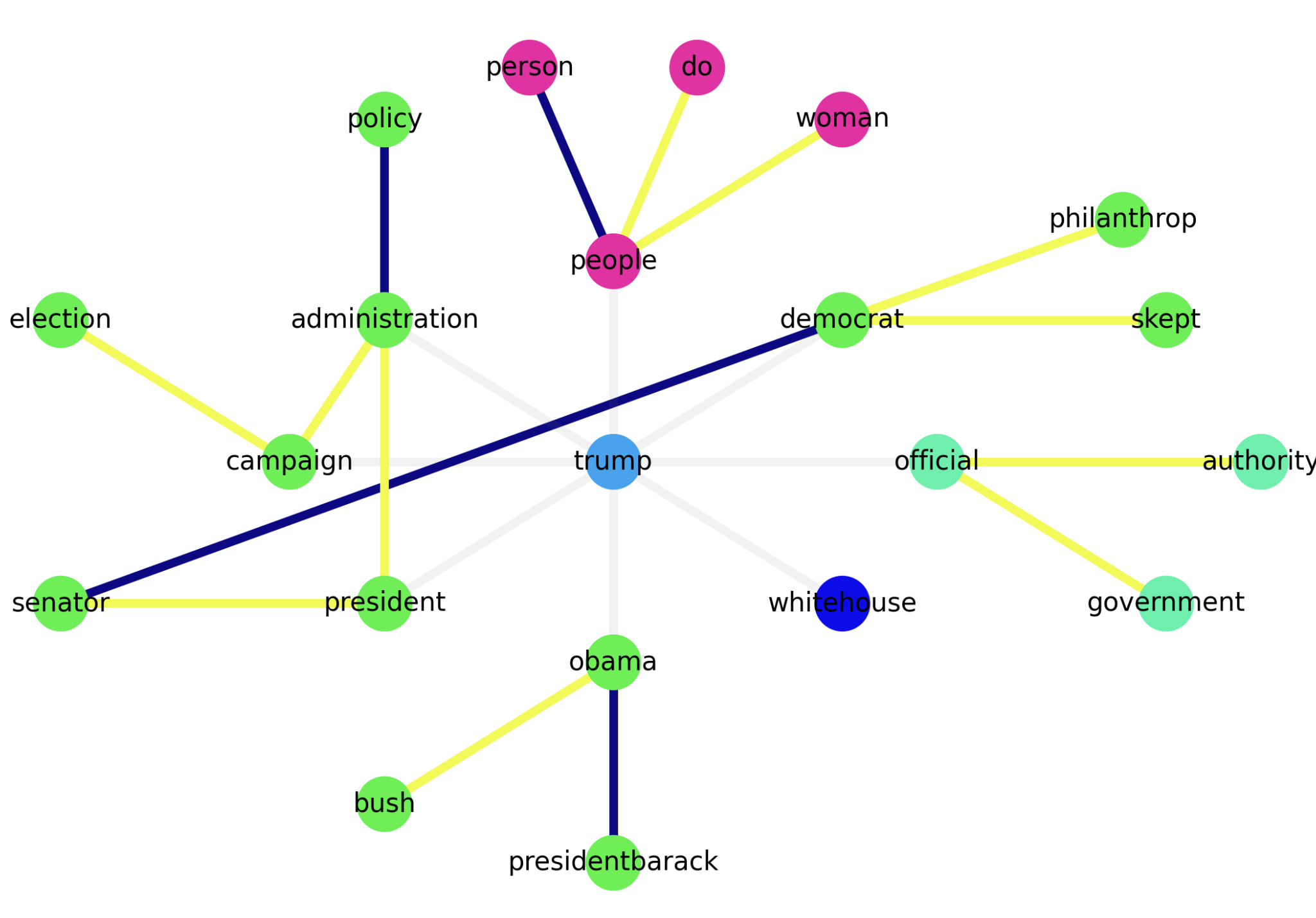}
	   \caption{Year 2017}
	   \label{fig:trump_cluster_refined_2017}
	\end{subfigure}
    \caption{Peripheral connectivity clusters of \textit{trump} aligned across consecutive time periods. Dark Blue: Residual cluster. Clusters (6 in total) are matched based on node overlap with the immediately preceding period, preserving fine-grained and temporally localized senses while increasing sensitivity to transient semantic change.}

    \label{fig:trump_clusters_refined}
\end{figure*}

\begin{figure*}
	\centering
	\begin{subfigure}[b]{0.48\textwidth}
		\centering
		\includegraphics[width=\textwidth]{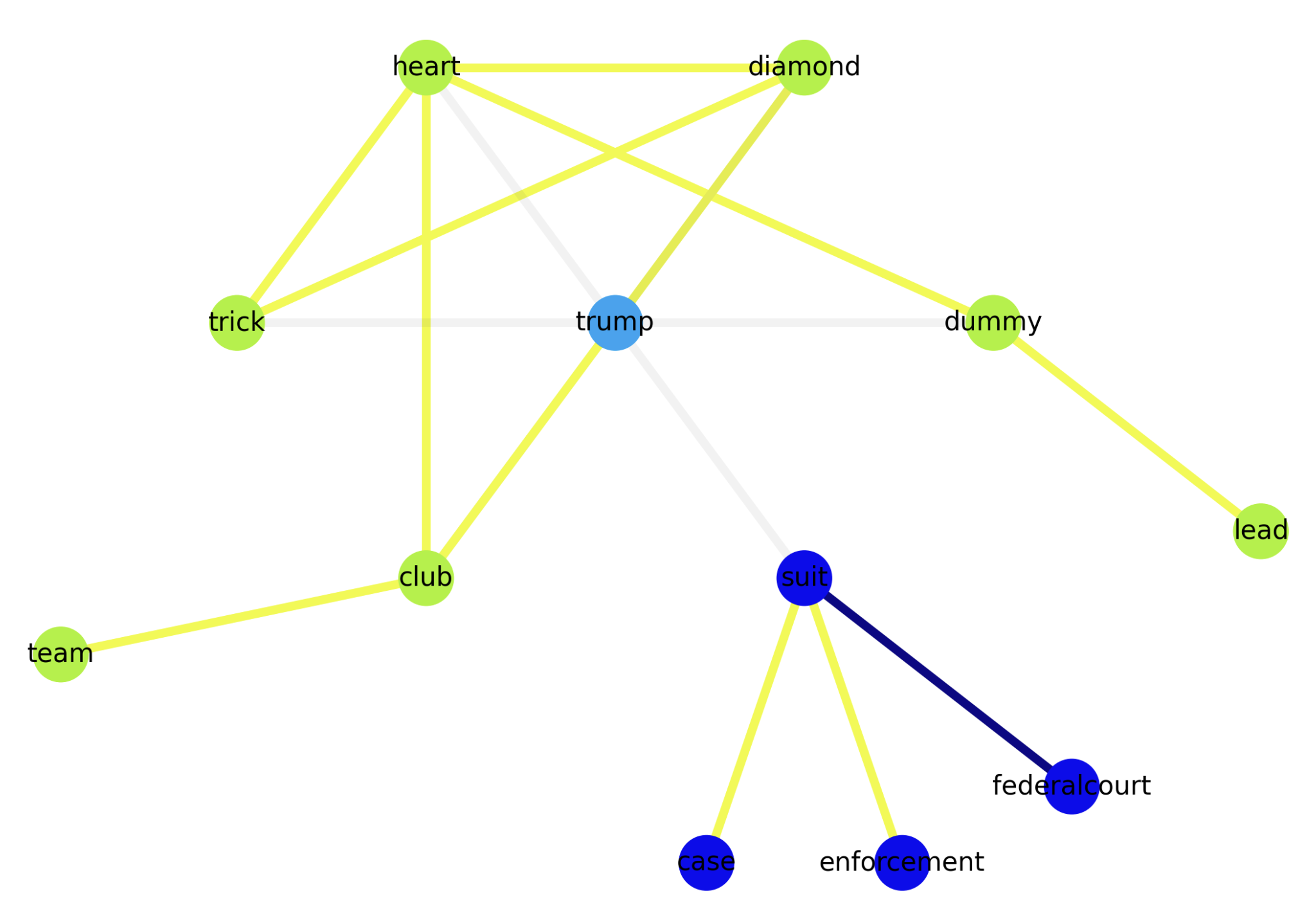}
		\caption{Year 1980}
		\label{fig:trump_cluster_refined2_1980}
	\end{subfigure}
	\hfill
	\begin{subfigure}[b]{0.48\textwidth}
	   \centering
	   \includegraphics[width=\textwidth]{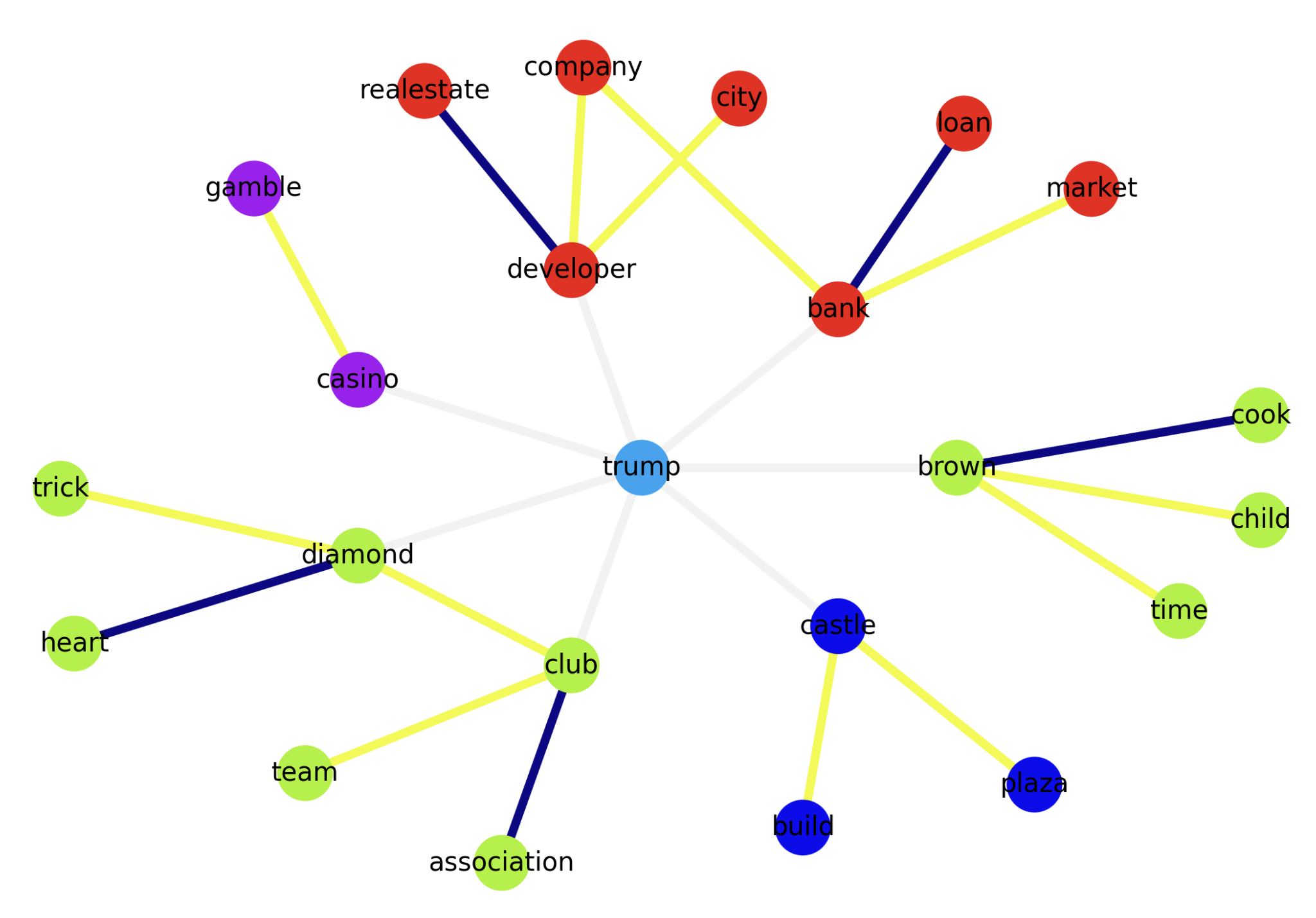}
	   \caption{Year 1990}
	   \label{fig:trump_cluster_refined2_1990}
    \end{subfigure}
	\vfill
	\begin{subfigure}[b]{0.48\textwidth}
	   \centering
	   \includegraphics[width=\textwidth]{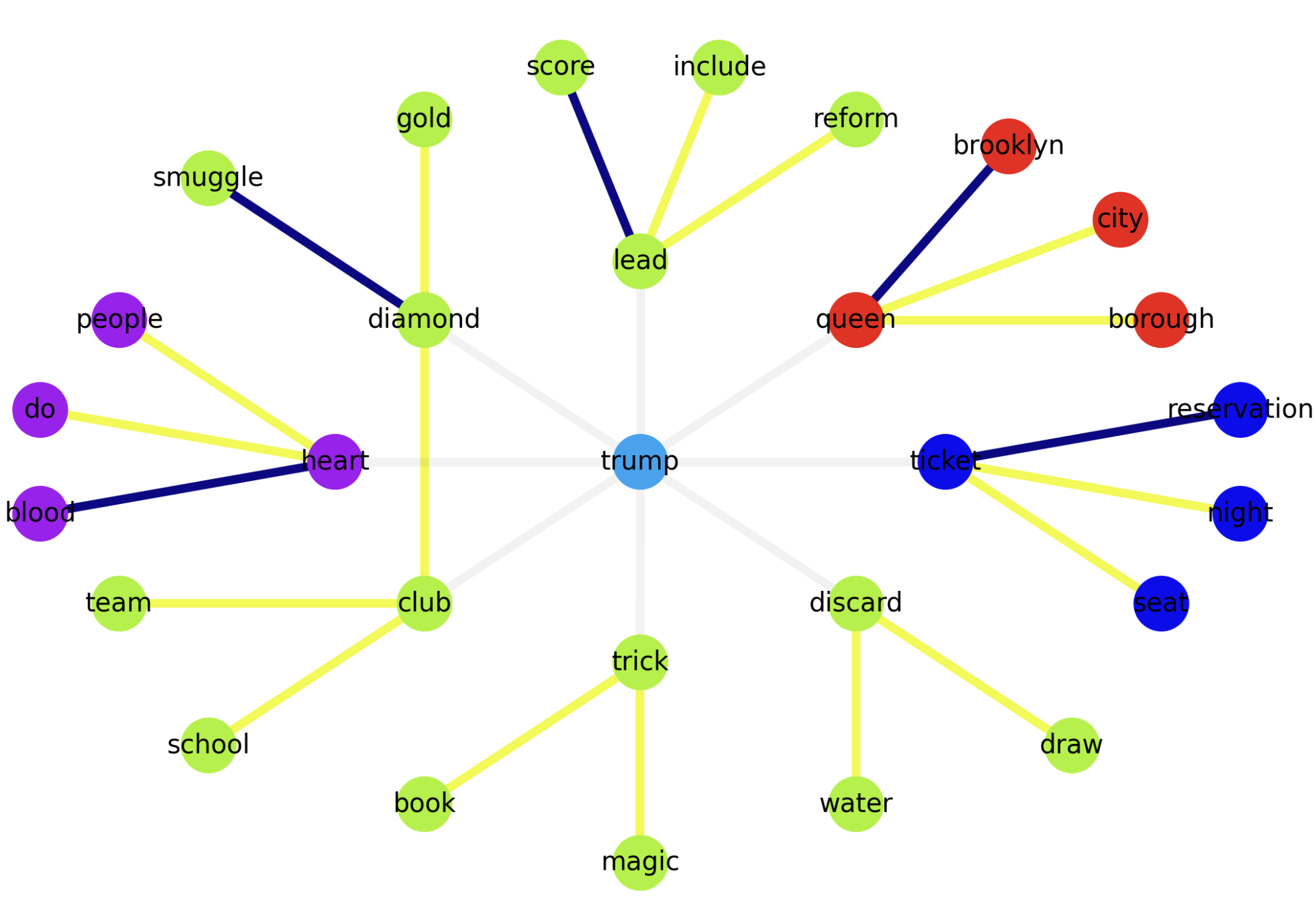}
	   \caption{Year 2000}
	   \label{fig:trump_cluster_refined2_2000}
    \end{subfigure}
	\hfill
	\begin{subfigure}[b]{0.48\textwidth}
	   \centering
	   \includegraphics[width=\textwidth]{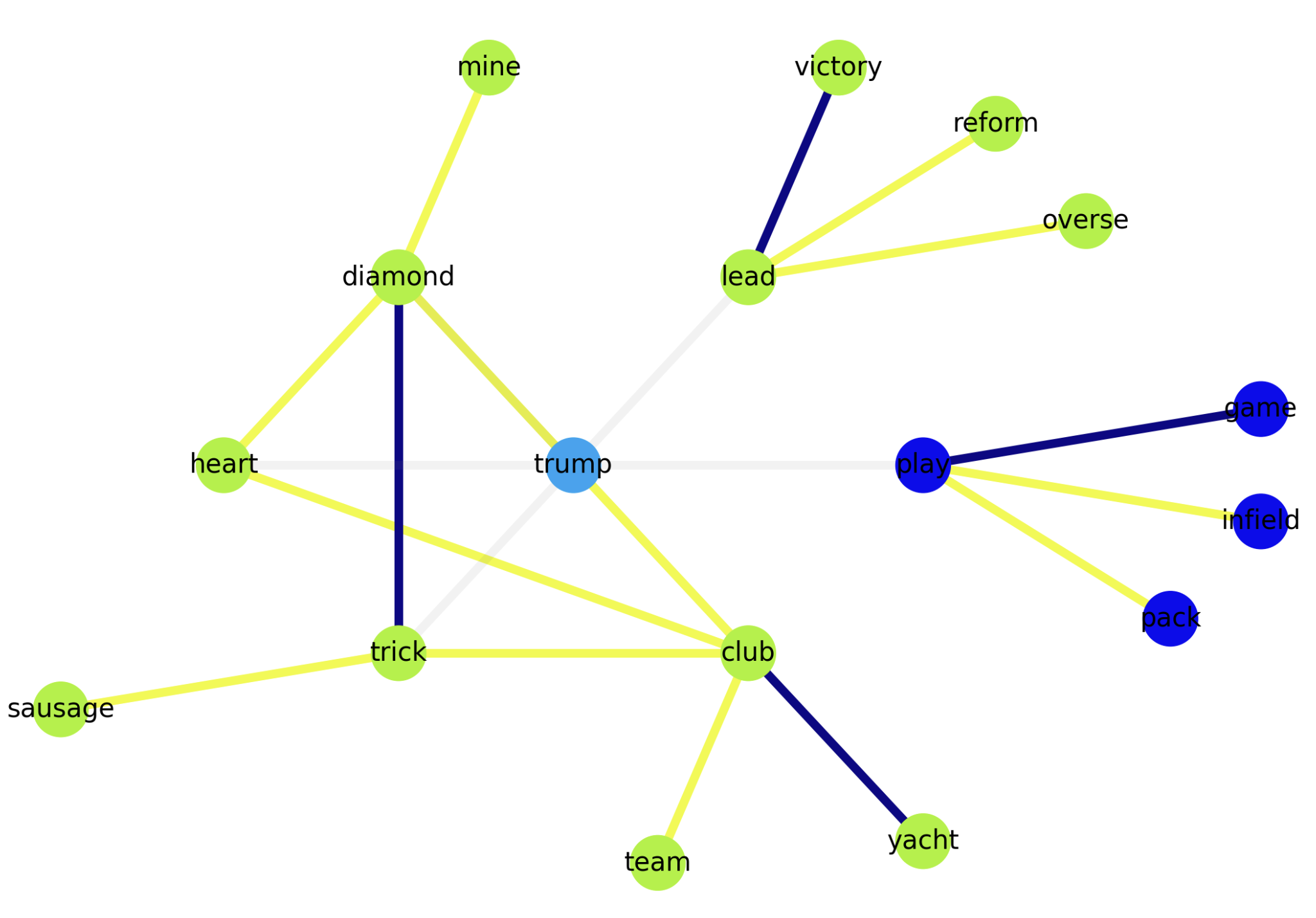}
	   \caption{Year 2010}
	   \label{fig:trump_cluster_refined2_2010}
    \end{subfigure}
	\vfill
	\begin{subfigure}[b]{0.48\textwidth}
	   \centering
	   \includegraphics[width=\textwidth]{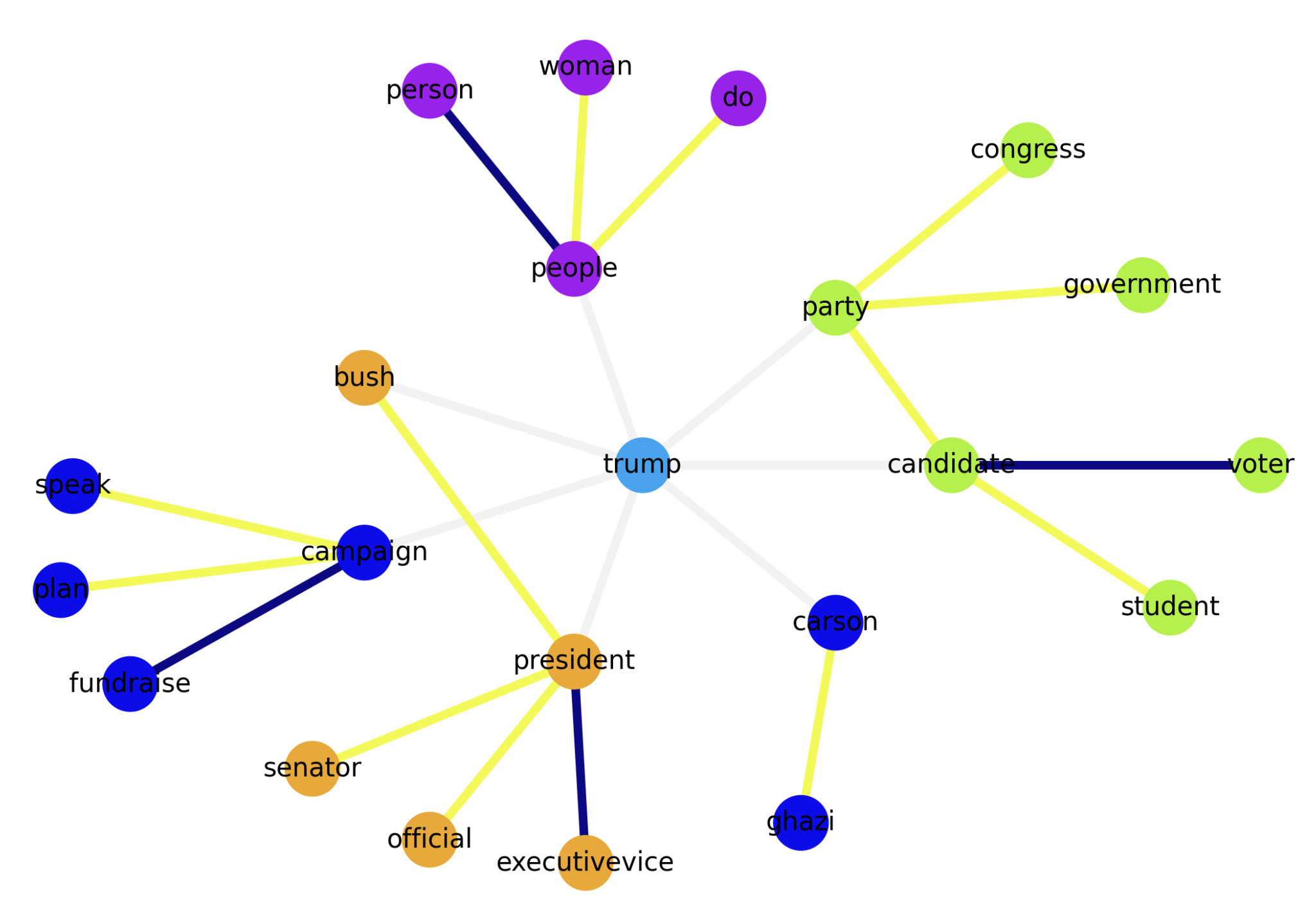}
	   \caption{Year 2015}
	   \label{fig:trump_cluster_refined2_2015}
    \end{subfigure}
	\hfill
	\begin{subfigure}[b]{0.48\textwidth}
	   \centering
	   \includegraphics[width=\textwidth]{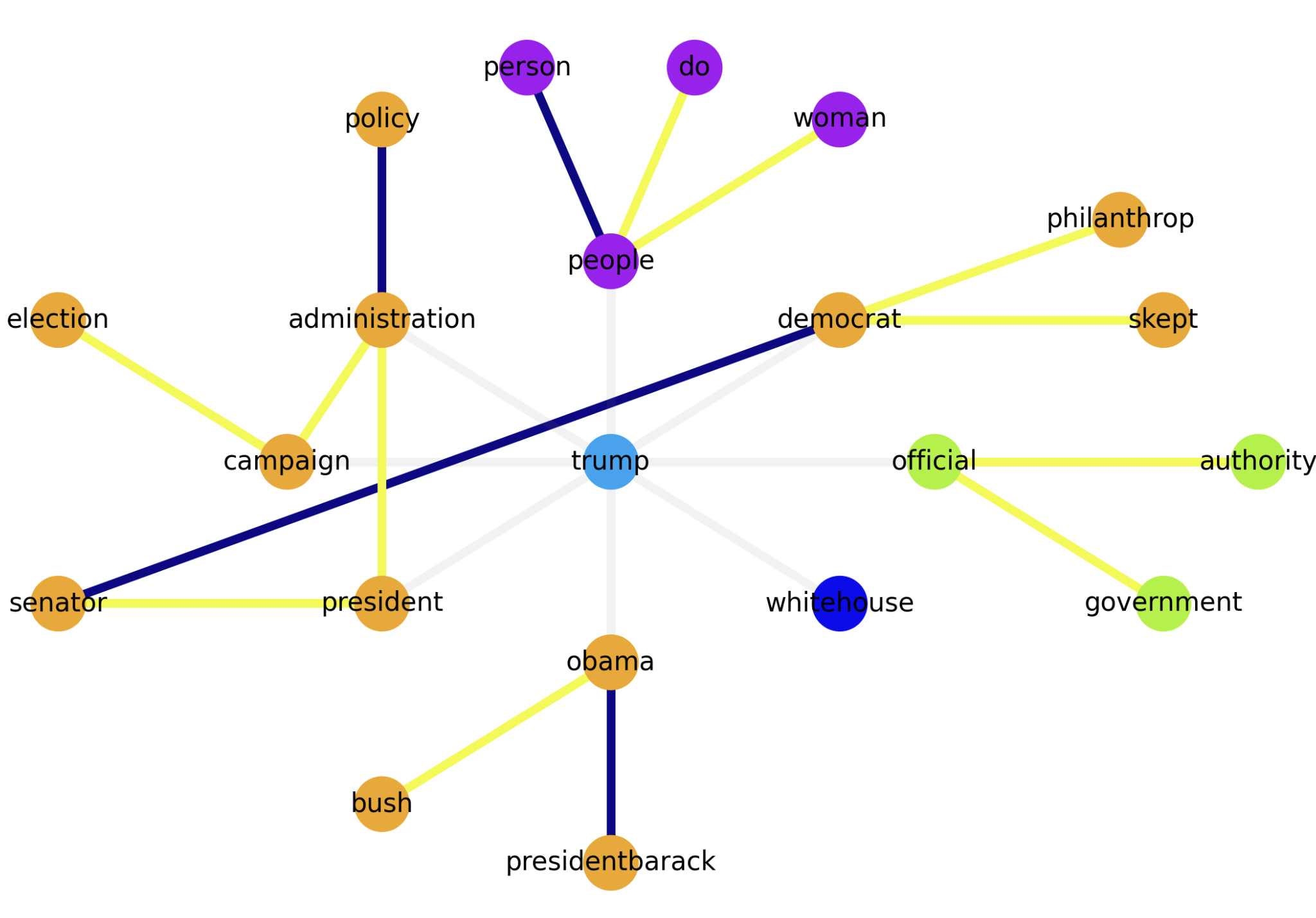}
	   \caption{Year 2017}
	   \label{fig:trump_cluster_refined2_2017}
	\end{subfigure}
    \caption{Peripheral connectivity clusters of \textit{trump} aligned across all historical periods. Dark Blue: Residual cluster. This alignment produces fewer (4 in total), more stable clusters but may merge emerging or transient senses into earlier dominant clusters.}
    \label{fig:trump_clusters_refined2}
\end{figure*}

\begin{figure*}
	\centering
	\begin{subfigure}[b]{0.48\textwidth}
		\centering
		\includegraphics[width=\textwidth]{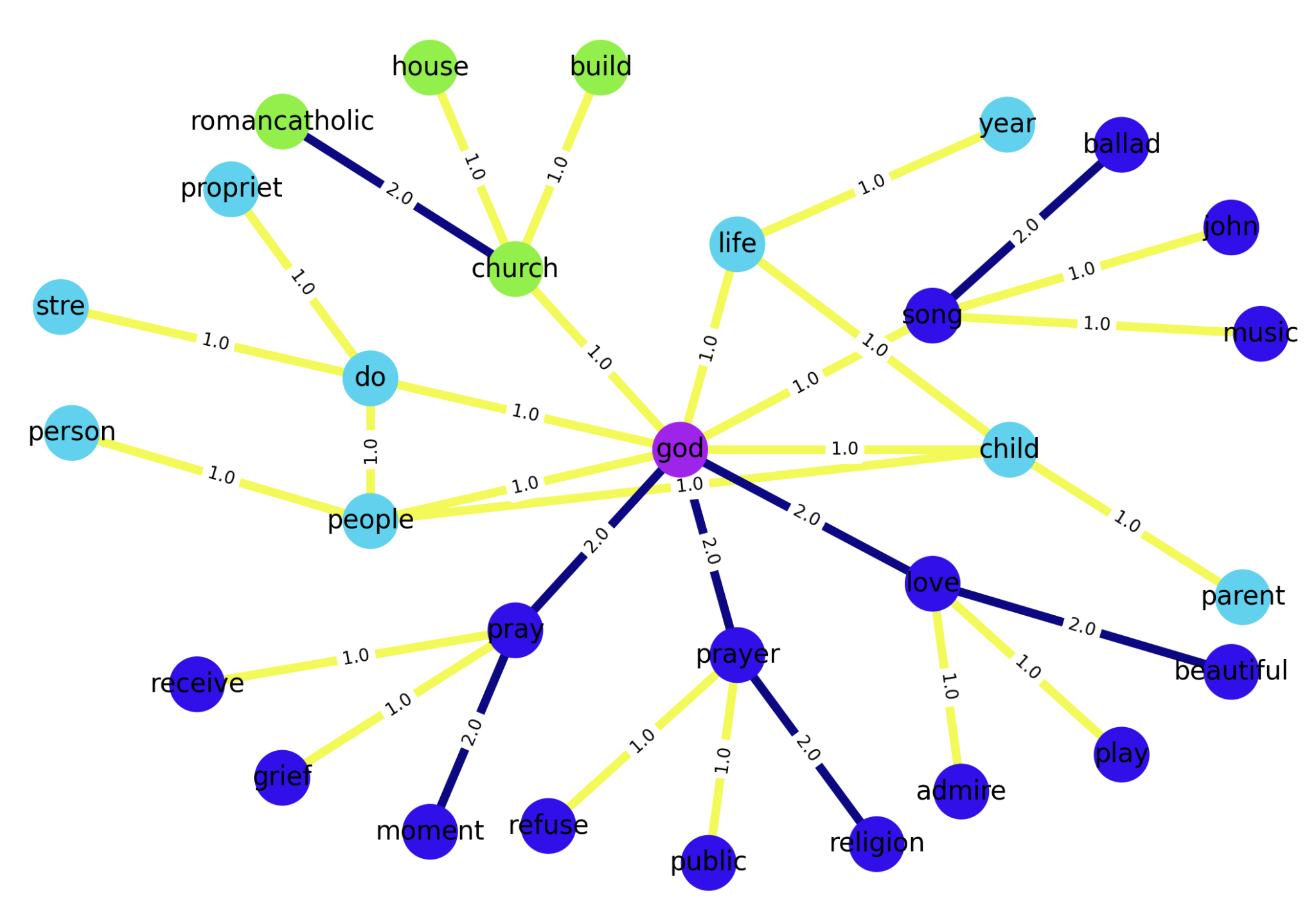}
		\caption{Year 1980}
		\label{fig:god_cluster_refined2_1980}
	\end{subfigure}
	\hfill
	\begin{subfigure}[b]{0.48\textwidth}
	   \centering
	   \includegraphics[width=\textwidth]{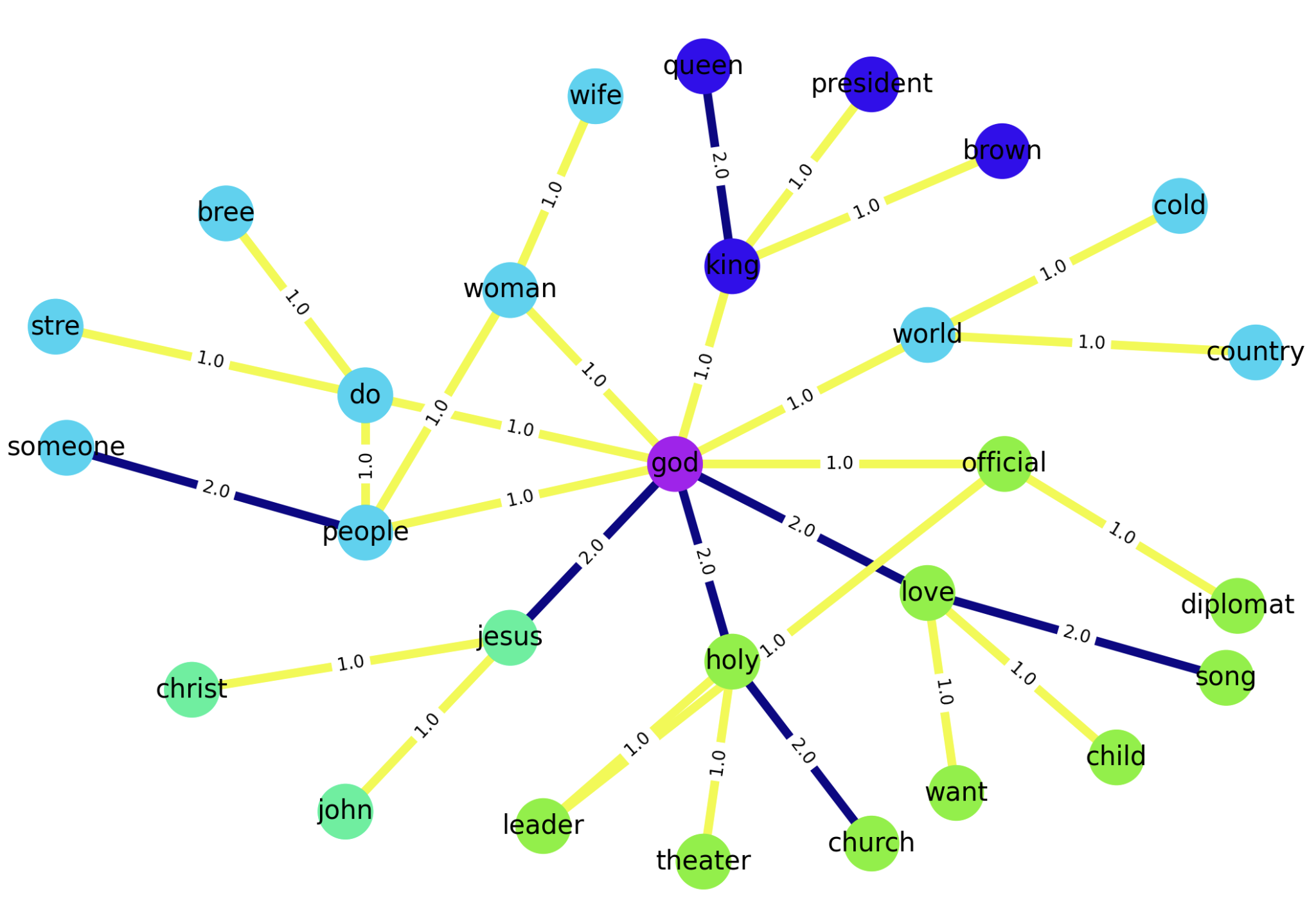}
	   \caption{Year 1990}
	   \label{fig:god_cluster_refined2_1990}
    \end{subfigure}
	\vfill
	\begin{subfigure}[b]{0.48\textwidth}
	   \centering
	   \includegraphics[width=\textwidth]{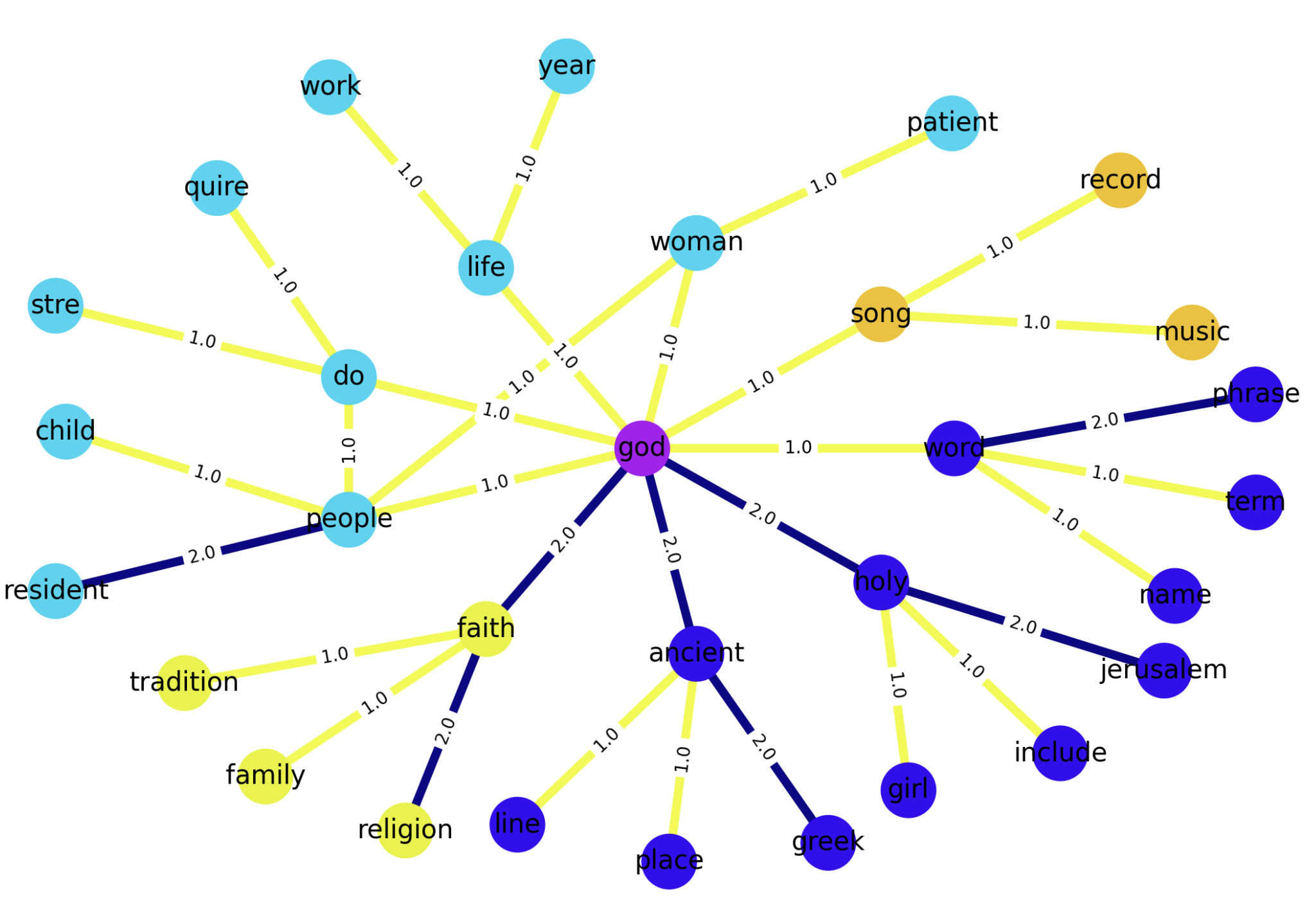}
	   \caption{Year 2000}
	   \label{fig:god_cluster_refined2_2000}
    \end{subfigure}
	\hfill
	\begin{subfigure}[b]{0.48\textwidth}
	   \centering
	   \includegraphics[width=\textwidth]{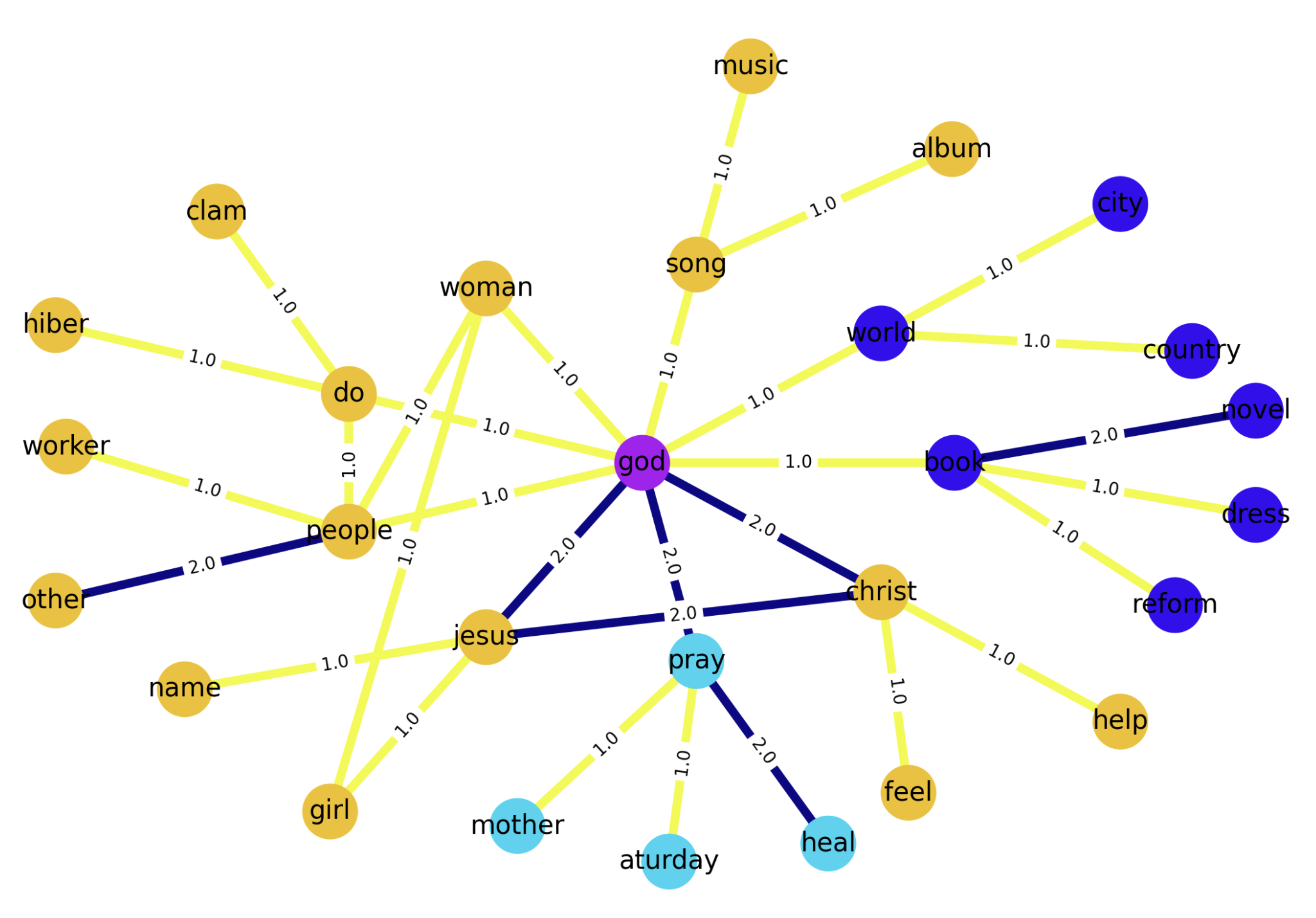}
	   \caption{Year 2010}
	   \label{fig:god_cluster_refined2_2010}
    \end{subfigure}
	\vfill
	\begin{subfigure}[b]{0.48\textwidth}
	   \centering
	   \includegraphics[width=\textwidth]{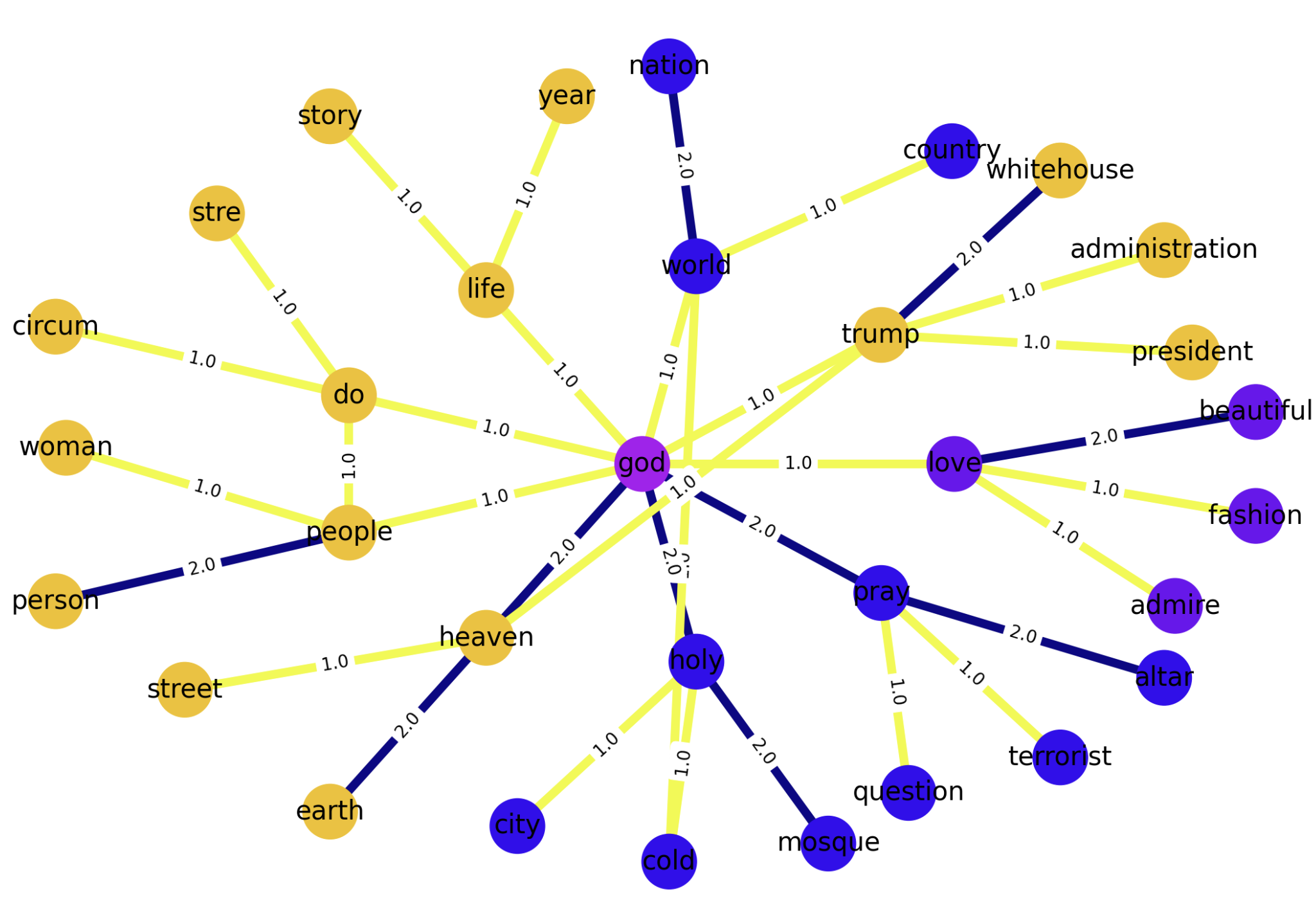}
	   \caption{Year 2017}
	   \label{fig:god_cluster_refined2_2017}
    \end{subfigure}
	\hfill
	\begin{subfigure}[b]{0.48\textwidth}
	   \centering
	   \includegraphics[width=\textwidth]{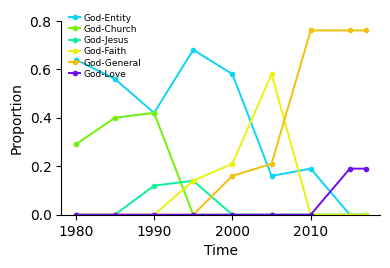}
	   \caption{Sense distribution}
	   \label{fig:god_sd}
	\end{subfigure}
    \caption{Peripheral connectivity clustering and sense distribution of \textit{god} aligned across all historical periods. Dark Blue: Residual cluster. Clusters exhibit strong semantic overlap and frequent node transitions, indicating facets of a single underlying sense rather than distinct competing meanings.}

    \label{fig:god_clusters_sd}
\end{figure*}

\begin{figure*}
	\centering
	\begin{subfigure}[b]{0.48\textwidth}
		\centering
		\includegraphics[width=\textwidth]{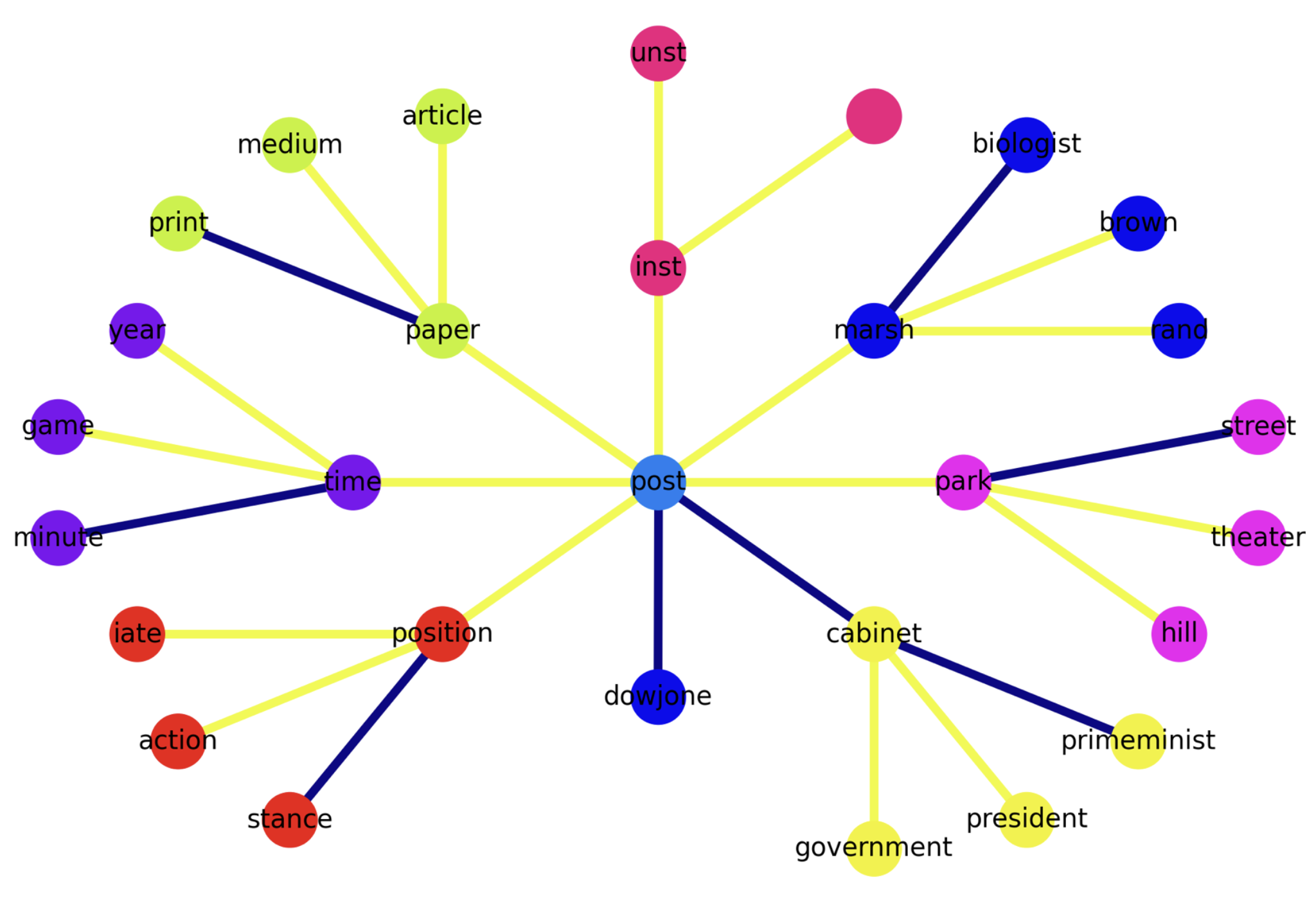}
		\caption{Year 1980}
		\label{fig:post_cluster_refined2_1980}
	\end{subfigure}
	\hfill
	\begin{subfigure}[b]{0.48\textwidth}
	   \centering
	   \includegraphics[width=\textwidth]{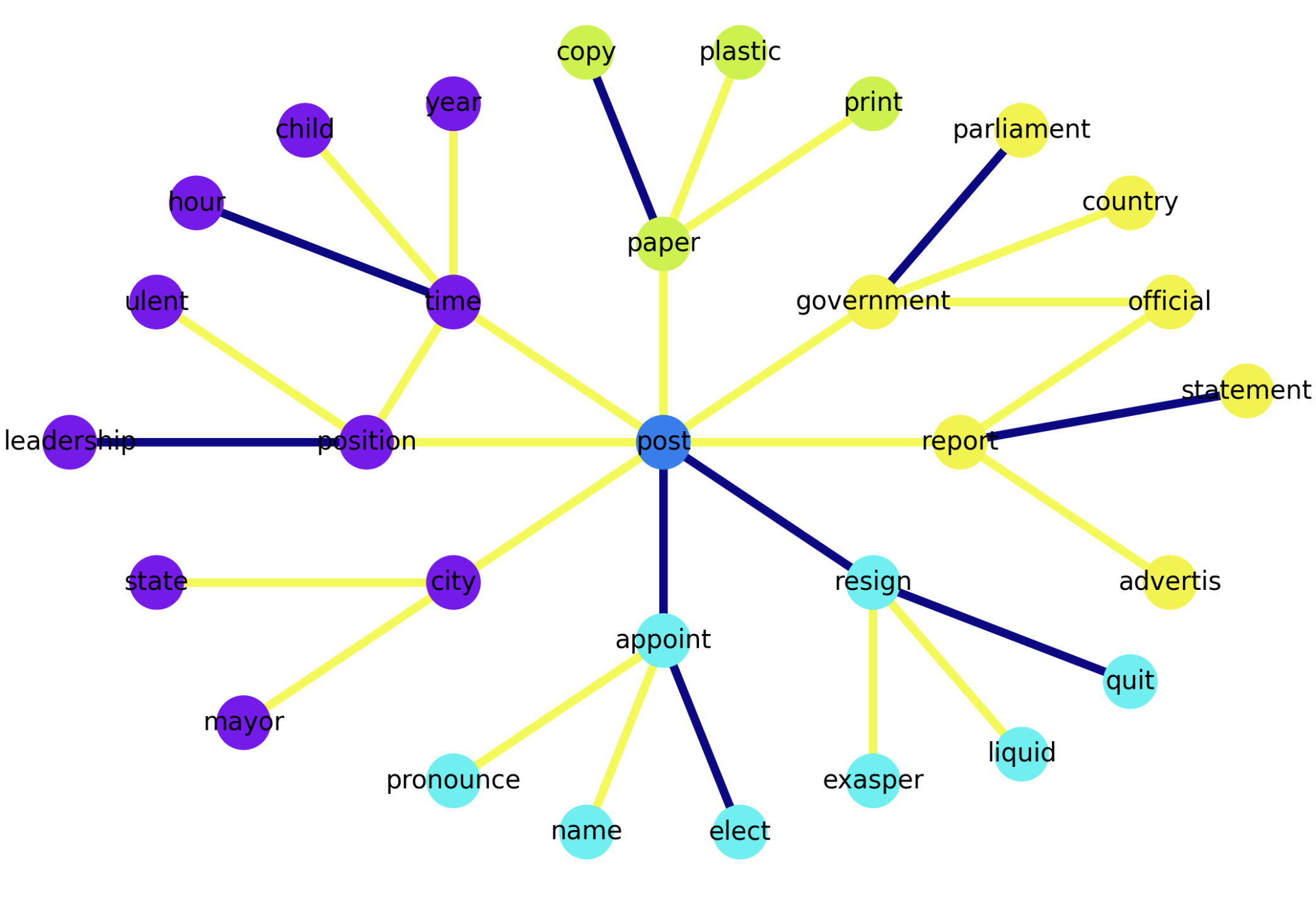}
	   \caption{Year 1990}
	   \label{fig:post_cluster_refined2_1990}
    \end{subfigure}
	\vfill
	\begin{subfigure}[b]{0.48\textwidth}
	   \centering
	   \includegraphics[width=\textwidth]{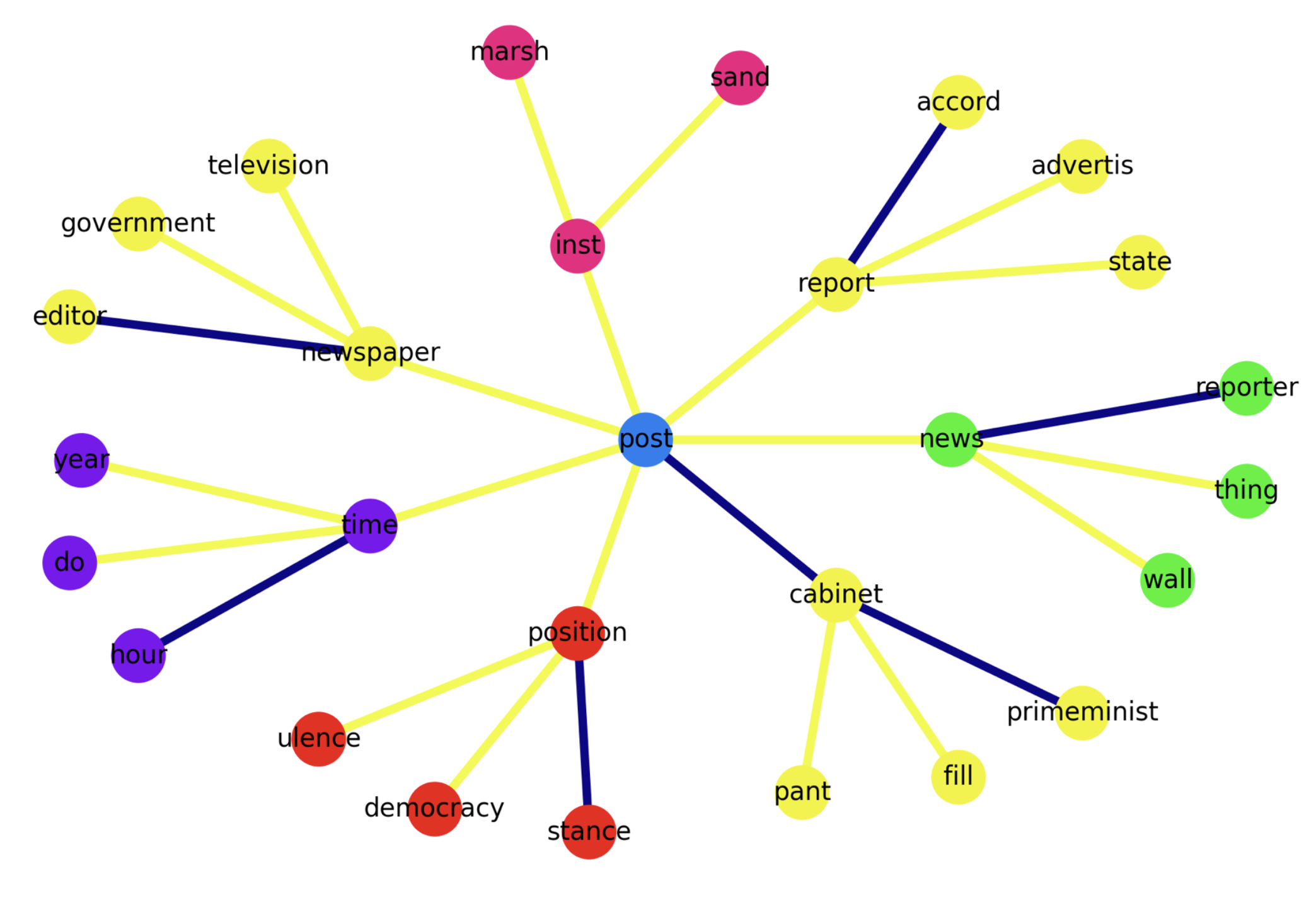}
	   \caption{Year 2000}
	   \label{fig:post_cluster_refined2_2000}
    \end{subfigure}
	\hfill
	\begin{subfigure}[b]{0.48\textwidth}
	   \centering
	   \includegraphics[width=\textwidth]{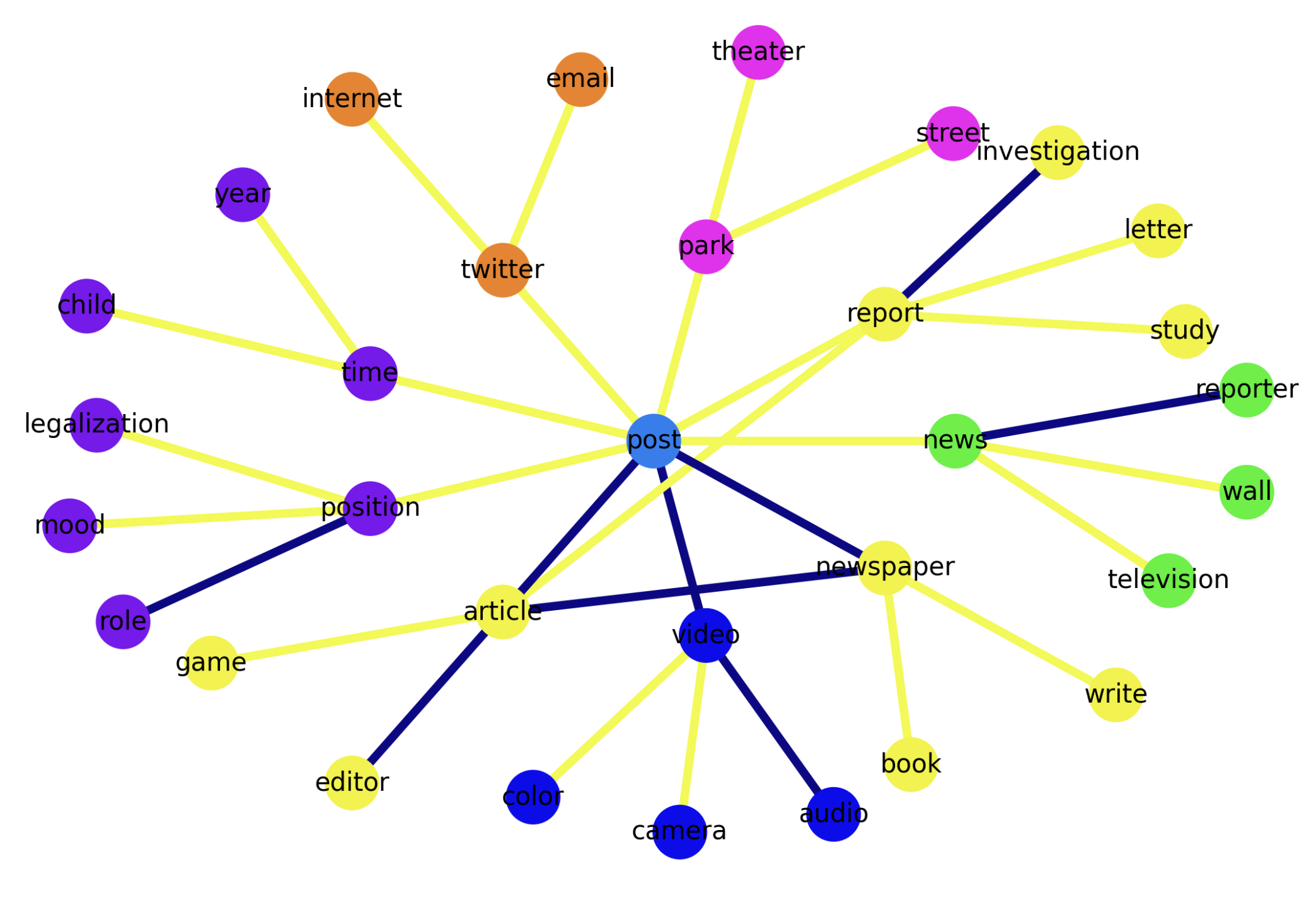}
	   \caption{Year 2010}
	   \label{fig:post_cluster_refined2_2010}
    \end{subfigure}
	\vfill
	\begin{subfigure}[b]{0.48\textwidth}
	   \centering
	   \includegraphics[width=\textwidth]{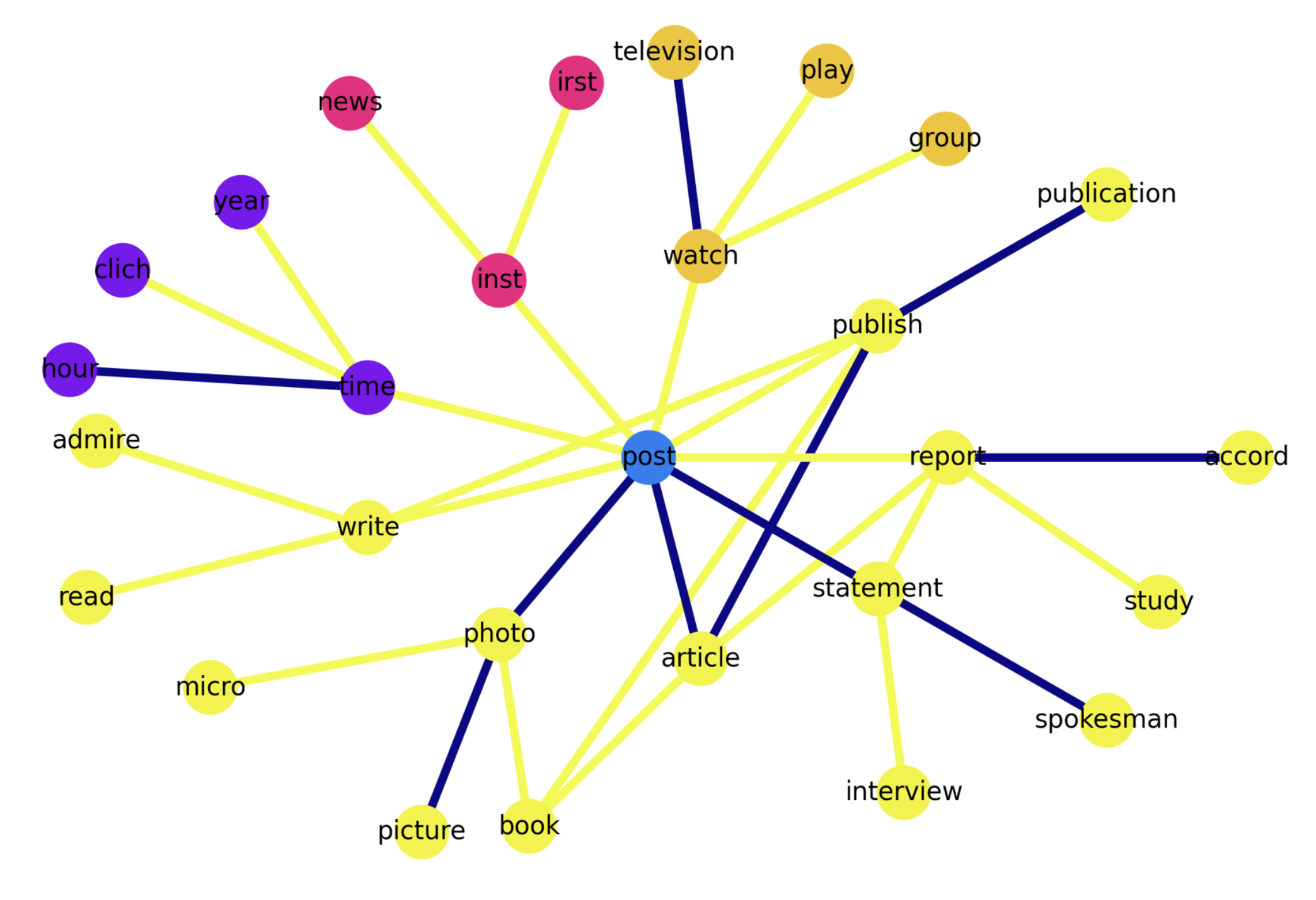}
	   \caption{Year 2015}
	   \label{fig:post_cluster_refined2_2015}
    \end{subfigure}
	\hfill
	\begin{subfigure}[b]{0.48\textwidth}
	   \centering
	   \includegraphics[width=\textwidth]{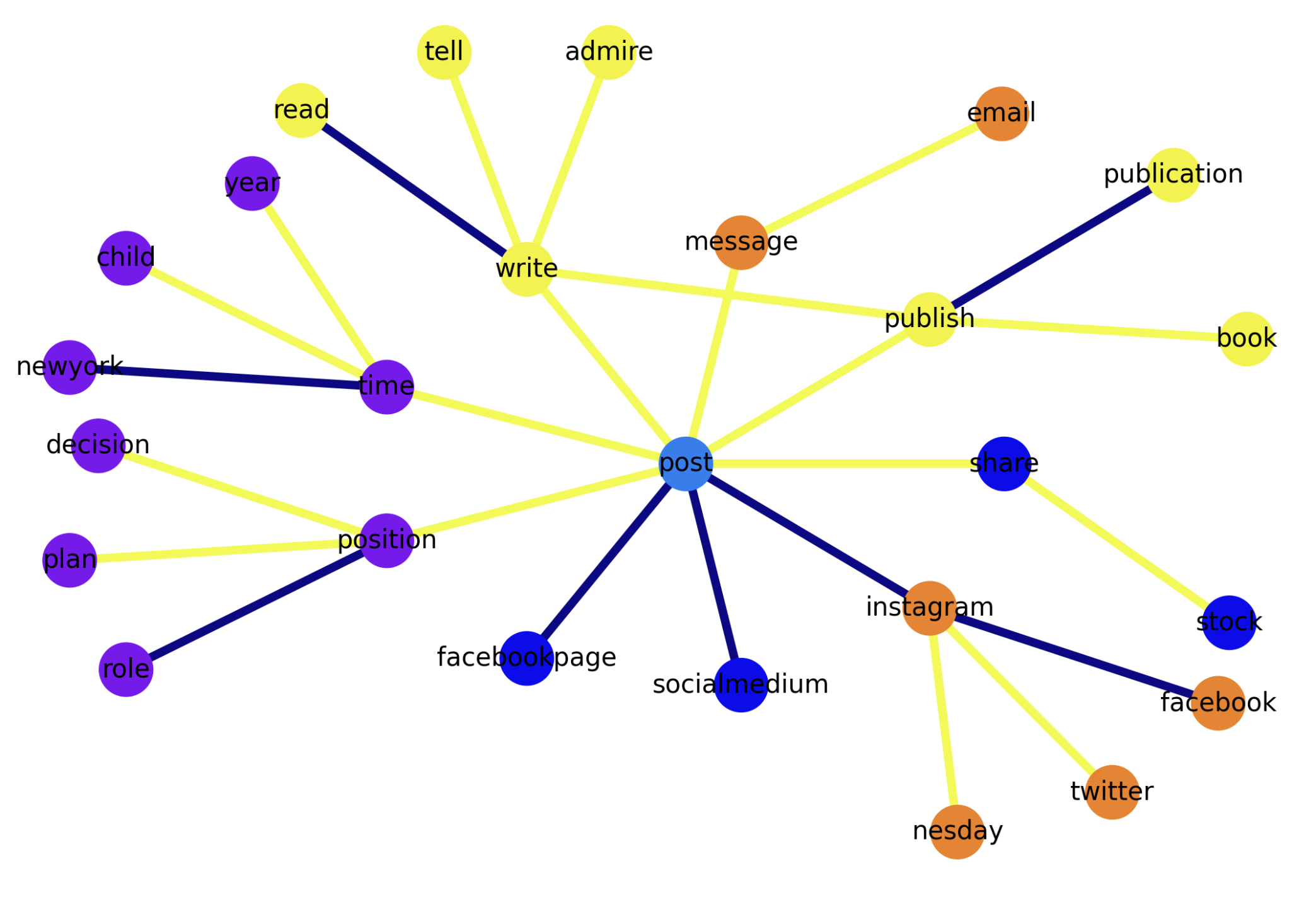}
	   \caption{Year 2017}
	   \label{fig:post_cluster_refined2_2017}
	\end{subfigure}
    \caption{Peripheral connectivity clustering of the \textit{post} neighborhood across time. Dark Blue: Residual cluster. Clusters associated with digital communication grow in prominence over later periods, reflecting gradual sense broadening rather than abrupt sense replacement.}
    \label{fig:post_clusters_refined2}
\end{figure*}

\end{document}